\def\arxivversion{1}
\documentclass[10pt,twocolumn,letterpaper]{article}

\ifdefined\arxivversion
\else
\documentclass[10pt,twocolumn,letterpaper]{article}
\fi
\ifdefined\arxivversion
\fi

\ifdefined\arxivversion
\usepackage[pagenumbers]{wacv} %
\else
\usepackage{wacv}              %
\fi

\usepackage{graphicx}
\usepackage{stfloats}
\usepackage{xr-hyper}
\ifdefined\arxivversion
\else
\IfFileExists{supp.aux}{%
  \externaldocument[S-]{supp}%
}{}
\fi
\newcommand{\suppref}[1]{%
  \ifdefined\arxivversion
    \cref{#1}%
  \else
    \cref{S-#1}%
  \fi
}
\usepackage{tikz}
\usetikzlibrary{calc,positioning}
\usepackage{pgfplots}
\pgfplotsset{compat=1.18}
\usepgfplotslibrary{groupplots}
\usepackage{booktabs}
\usepackage{multirow}
\usepackage{makecell}
\usepackage[normalem]{ulem}
\usepackage{subcaption}
\usepackage{siunitx}
\usepackage[accsupp]{axessibility}
\usepackage{xcolor}
\usepackage{colortbl}
\definecolor{tabgold}{RGB}{255,223,100}
\definecolor{tabsilver}{RGB}{210,210,210}
\definecolor{tabbrz}{RGB}{210,170,120}

\newcommand{\psiPD}{\ensuremath{\psi\text{-PD}}}

\ifdefined\arxivversion
\usepackage{algorithm}
\usepackage{algpseudocode}
\usepackage[export]{adjustbox}
\fi

\definecolor{wacvblue}{rgb}{0.21,0.49,0.74}
\usepackage[pagebackref,breaklinks,colorlinks,allcolors=wacvblue]{hyperref}

\def\wacvPaperID{1800} %
\def\confName{WACV}
\def\confYear{2027}

\title{Wavelet Phase Diffusion for Structurally and Semantically Consistent Sim-to-Real Translation}

\author{
Kaiwen Wang$^{1}$ \quad Frank Bieder$^{2}$ \quad Yinzhe Shen$^{1}$\\
Carlos Fernandez$^{1}$ \quad Jan-Hendrik Pauls$^{1}$ \quad Omer Sahin Tas$^{2}$\\
{$^{1}$Karlsruhe Institute of Technology, Germany}\\
{$^{2}$FZI Research Center for Information Technology, Germany}\\
{\tt\small kaiwen.wang@kit.edu}
}

\begin{document}

\makeatletter
\let\orig@@maketitle\@maketitle
\renewcommand{\@maketitle}{%
  \orig@@maketitle
  \vspace{0pt}
  \noindent\begin{minipage}{\linewidth}
    \centering
    \input{figures/teaser.tex}
    \captionsetup{hypcap=false}%
    \captionof{figure}{\textbf{$\psiPD$ achieves structurally and semantically consistent sim-to-real translation without paired data or heavy inference conditioning, and generalizes zero-shot beyond the simulation domain.}}
    \label{fig:teaser}
  \end{minipage}
  \vspace{6pt}
}
\makeatother

\let\origmaketitle\maketitle
\maketitle
\begin{abstract}
Simulation-to-reality translation must bridge the appearance gap between synthetic and real domains while preserving structural and semantic consistency.
Conditioning-based methods achieve spatial alignment but introduce computationally expensive control modules. Meanwhile, paired-data methods achieve realism but rely on complex synthesis pipelines, often altering scene geometry and semantics.
Training-free editing methods avoid both constraints but lack a learned appearance prior, limiting their perceptual quality.
Recently proposed phase-preserving diffusion presents a promising alternative, but Fourier-domain formulations are constrained by global spectral coupling. This coupling induces spatial artifacts such as ringing and boundary leakage, thereby degrading structural and semantic consistency.
We introduce Wavelet Phase Diffusion ($\psiPD$), which addresses this through two components.
First, we operate in the Dual-Tree Complex Wavelet Packet Transform (DT-$\mathbb{C}$WPT) domain, whose localized wavelet packets enable spatially adaptive phase injection without global spectral interference.
Second, Low-Frequency Randomization (LFR) replaces the low-frequency packet, decoupling the model from the synthetic illumination prior and enabling in-distribution real-world appearance.
Both components train on unpaired open-domain data, integrate into arbitrary diffusion backbones without architectural modification, and introduce negligible inference overhead.
The spatial locality further enables instance-level translation, where individual objects or regions are translated to photorealistic appearance independently while the surrounding scene remains untranslated.
On vKITTI $\to$ KITTI image translation, $\psiPD$ outperforms prior methods in realism and semantic consistency while maintaining competitive structural alignment.
For CARLA video translation, $\psiPD$ approaches the realism of paired-data methods while reducing VLM planner ADE and FDE by $5.4\%$ and $5.1\%$, respectively. 
It is the only evaluated method to jointly improve both appearance and downstream utility.
Code and models are available on our \href{https://kit-mrt.github.io/Wavelet-Phase-Diffusion/}{project page}.
\end{abstract}

\section{Introduction}
\label{sec:intro}

Simulation-to-reality (sim-to-real) translation seeks to bridge the visual domain gap between synthetic environments and real-world imagery while preserving the structural and semantic cues that downstream perception and planning pipelines depend on.
Unlike general image editing~\cite{brooks2023instructpix2pix}, realism here is necessary but not sufficient: a model that improves realism by hallucinating lane geometry or altering traffic sign appearance makes the translation \emph{worse} for its intended use, regardless of its photographic quality.

Traditional methods typically approach this challenge from three distinct angles, each carrying a characteristic cost.
Conditioning-based methods~\cite{controlnet,cosmos,jiang2025vace} achieve structural alignment by injecting dense control signals (\eg depth maps, edge maps, or semantic segmentations) into the diffusion process, but require dedicated control modules, introduce substantial inference-time overhead, and struggle to fully suppress synthetic artifacts when the signals themselves are derived from simulation.
Training on paired data~\cite{ditto} yields powerful domain mappings, but relies on complex, multi-stage pipelines to synthesize paired training samples, which is computationally expensive, introduces generator-specific biases, and struggles with domain generalization.
Training-free editing methods~\cite{kulikov2025flowedit,dnaedit} avoid these training and conditioning requirements, but without a learned appearance prior, inference-time manipulation alone cannot achieve the perceptual quality required to bridge the sim-to-real gap.

Phase-preserving diffusion has recently emerged as a principled alternative.
Prior work in signal processing~\cite{oppenheim2005importance} establishes that phase predominantly encodes geometric structure whereas magnitude governs texture statistics.
By constraining phase on the domain-invariant frequency band while randomizing magnitude, diffusion models can synthesize realistic texture without disrupting structural layout.
Recent methods~\cite{zeng2025neuralremaster} implement this in the Fourier domain, but Fourier basis functions have global support: frequency-domain phase constraints in one region implicitly affect the entire image, introducing global spectral coupling that leads to ringing and boundary leakage, particularly in scenes with heterogeneous structural requirements.

We propose $\psiPD$, a phase-preserving diffusion framework that requires no inference-time conditioning, no paired data, and overcomes the global spectral coupling of Fourier-domain methods.
$\psiPD$ comprises two components trained on unpaired, open-domain data.
First, we replace globally supported Fourier bases with the Dual-Tree Complex Wavelet Packet Transform (DT-$\mathbb{C}$WPT), which yields localized complex wavelet packets, enabling spatially adaptive phase injection without global interference.
Second, we introduce \emph{Low-Frequency Randomization} (LFR), which randomizes the low-frequency packet of the source latent during noise construction, decoupling the model from the synthetic global illumination prior encoded in simulated inputs and allowing it to produce in-distribution real-world lighting.
The spatially adaptive design further enables instance-level translation, where individual objects or regions are translated to photorealistic appearance independently while the surrounding scene remains untranslated.

We evaluate $\psiPD$ on sim-to-real image and video translation benchmarks~\cite{vkitti,kitti,dosovitskiy2017carla}.
Across both benchmarks, \textbf{$\psiPD$ is the only evaluated method to consistently improve photorealistic appearance while simultaneously enhancing downstream planning performance on videos}, requiring no paired data or conditioning signals.

The contributions of this work are as follows:
\begin{itemize}
    \item We introduce DT-$\mathbb{C}$WPT-based phase injection, which overcomes the global spectral coupling of Fourier-domain methods and enables spatially adaptive structure-preserving diffusion without architectural modification.
    \item We propose Low-Frequency Randomization (LFR), which decouples the generative process from the synthetic global illumination prior, enabling realistic in-distribution appearance without paired data.
    \item We adopt VLM-based trajectory planning error as a downstream metric for sim-to-real video translation, establishing a practical evaluation paradigm to assess planning-utility preservation.
    \item We show that $\psiPD$'s spatially adaptive cutoff map generalizes zero-shot to instance-level translation, even on non-simulated sources such as illustrations and comics.
\end{itemize}

\section{Related Work}
\label{sec:related}

\subsection{Diffusion Models}
Diffusion models achieve state-of-the-art generative performance across image~\cite{ldm,flux2024}, video~\cite{blattmann2023stable,hunyuanvideo2025,wan2025}, and 3D~\cite{zhang2024gaussiancube,voleti2024sv3d} synthesis, evolving from variational~\cite{ddpm,sohl2015deep} and score-based~\cite{song2019generative,song2020score} formulations to flow matching~\cite{lipman2022flow,liu2022flow,esser2024scaling} for more efficient sampling.
Despite their success, both diffusion and flow-based processes progressively transform data toward noise, degrading structure and semantics when strong appearance changes are required.

\subsection{Wavelet-Based Diffusion}
Wavelet representations~\cite{mallat2002theory} have been widely used for image compression~\cite{taubman2002jpeg2000}, restoration~\cite{chen2021all,huang2024wavedm}, and synthesis~\cite{guth2022wavelet,phung2023wavelet,friedrich2024wdm}, and have recently been integrated into diffusion models for reduced-resolution sampling or enhanced detail synthesis~\cite{friedrich2024cwdm,hiwave,latentwavelet}.
These works operate on real-valued coefficients, leaving open how localized spectral representations can impose semantics-aware structural constraints.
$\psiPD$ addresses this by operating in the DT-$\mathbb{C}$WPT domain, replacing global Fourier phase constraints with spatially localized wavelet phase injection.

\subsection{Sim-to-Real Translation}
Existing sim-to-real translation methods can be categorized by how they balance realism, structural fidelity, and data requirements.
Conditioning-based methods achieve structural alignment by injecting dense control signals (\eg depth maps, edge maps, or semantic segmentations) into the diffusion process at inference time, as in Cosmos-Transfer~2.5~\cite{cosmos} and VACE~\cite{jiang2025vace}.
These methods increase architectural complexity, incur significant computational overhead, and typically require task-specific supervision or annotations.
Paired-data methods such as Ditto~\cite{ditto} train large instruction-based editors on synthetically generated paired data, achieving high perceptual realism.
However, constructing such paired datasets is highly resource-intensive and often relies on synthetic generation pipelines, which can introduce domain bias and limit generalization to unseen environments. 
Furthermore, without explicit structural constraints during translation, they remain prone to altering scene geometry and semantics.
Training-free editing methods, \eg, FlowEdit~\cite{kulikov2025flowedit} and DNAEdit~\cite{dnaedit}, leverage the deterministic structure of rectified flow trajectories to steer generation without any training, but their edits are bounded by what inference-time trajectory perturbations can achieve.

Phase-preserving diffusion methods avoid conditioning by perturbing the input in frequency space: NeuralRemaster~\cite{zeng2025neuralremaster} preserves Fourier phase during diffusion training, achieving structure-aligned translation without conditioning modules but limited by global spectral coupling.
$\psiPD$ trains on unpaired open-domain data and requires no inference-time conditioning, approaching the realism of paired-data methods while preserving structure and enabling instance-level control that prior methods cannot.

\section{Preliminary: Phase-Preserving Diffusion}
\label{sec:preliminary}
As established by Oppenheim \etal~\cite{oppenheim2005importance}, the phase of a signal acts as the primary carrier of spatial structure, whereas its magnitude predominantly governs texture statistics. Specifically, we consider a complex-valued transform $\mathcal{T}$ that decomposes a latent $\mathbf{x}$ into a representation consisting of magnitude $|\mathcal{T}(\mathbf{x})|$ and phase $\angle\mathcal{T}(\mathbf{x})$ in polar form:
\begin{equation}
    \mathcal{T}(\mathbf{x}) = |\mathcal{T}(\mathbf{x})|\,e^{j\angle\mathcal{T}(\mathbf{x})}.
\end{equation}
Building on this principle, phase-preserving diffusion methods construct structured noise $\hat{\boldsymbol{\epsilon}}$ by combining the phase of the source latent $\mathbf{x}$ with the magnitude of a standard Gaussian noise sample $\boldsymbol{\epsilon}\sim\mathcal{N}(\mathbf{0},\mathbf{I})$.
Furthermore, to regulate the degree of structural preservation, a mask $\mathbf{M}$ defines the spatial or spectral regions where the source phase is injected and the noise phase is retained. The combined representation $\hat{\mathbf{y}}$ is formed as:
\begin{equation}
\label{eq:general_phase_preserve}
\hat{\mathbf{y}}
= \left|\mathcal{T}(\boldsymbol{\epsilon})\right|
    e^{j\cdot\big(\mathbf{M} \odot \angle\mathcal{T}(\mathbf{x})
    + (1-\mathbf{M}) \odot \angle\mathcal{T}(\boldsymbol{\epsilon})\big)},
\end{equation}
where $\odot$ denotes element-wise multiplication. The final structured noise $\hat{\boldsymbol{\epsilon}}$ is reconstructed via the inverse transform:
\begin{equation}
\label{eq:general_reconstruct}
\hat{\boldsymbol{\epsilon}} = \mathcal{T}^{-1}(\hat{\mathbf{y}}).
\end{equation}
For videos, the same per-frame construction is applied in the latent space of the backbone's 3D causal VAE.
Phase injection thus operates per meta-frame but, as each aggregates a short temporal window, constrains spatiotemporal structure rather than per-frame structure alone.
Inter-frame appearance consistency is enforced by a two-stage pipeline: an image stage first translates the initial frame into a photorealistic reference, which conditions the video stage so that all frames adopt a single, consistent real-domain appearance rather than independent per-frame realizations.

\section{Method}
\label{sec:method}

\subsection{Wavelet Analysis Fundamentals}
\label{ssec:wavelet_fundamentals}

\begin{figure}[h]
    \centering
    \begin{subfigure}[t]{0.19\linewidth}
        \includegraphics[width=\linewidth]{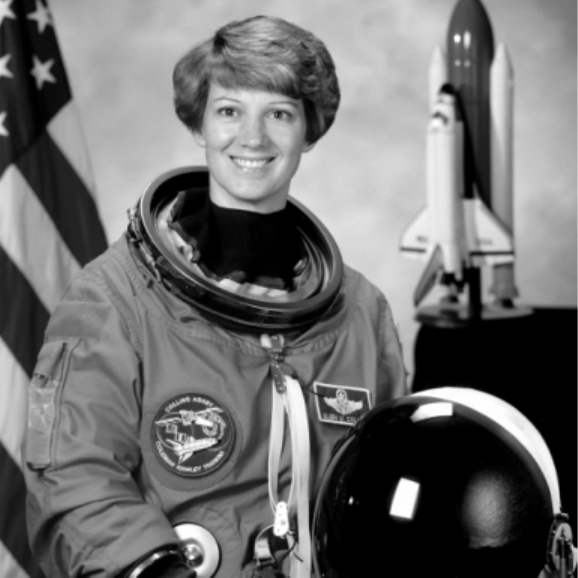}
        \caption*{\scriptsize Input}
    \end{subfigure}
    \hfill
    \begin{subfigure}[t]{0.19\linewidth}
       \includegraphics[width=\linewidth]{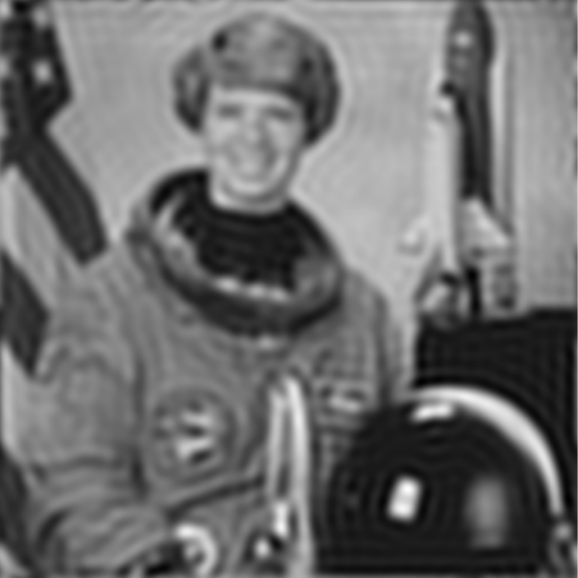}
        \caption*{\scriptsize Fourier $r{=}32$}
        \label{sfig:fft_32}
    \end{subfigure}
    \hfill
    \begin{subfigure}[t]{0.19\linewidth}
        \includegraphics[width=\linewidth]{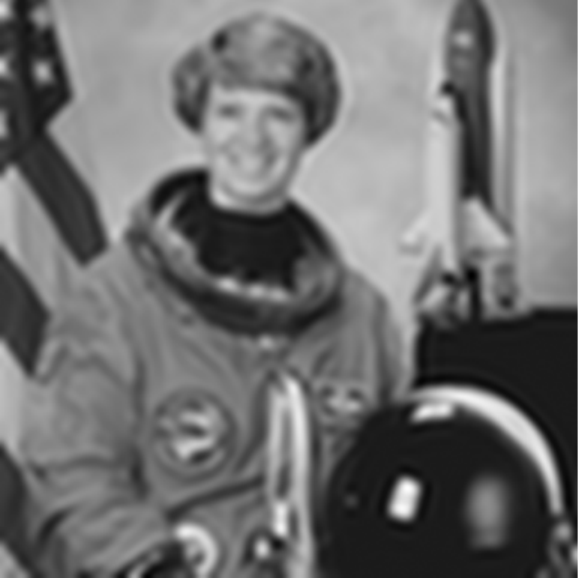}
        \caption*{\scriptsize Wavelet $r{=}32$}
        \label{sfig:dtcwpt_32}
    \end{subfigure}
    \hfill
    \begin{subfigure}[t]{0.19\linewidth}
        \includegraphics[width=\linewidth]{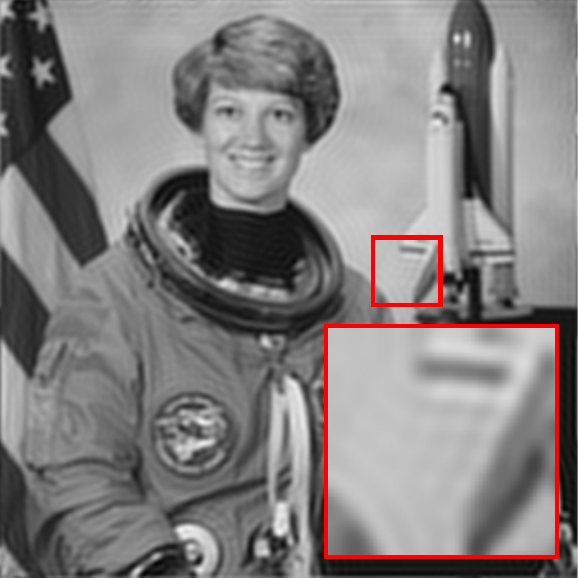}
        \caption*{\scriptsize Fourier $r{=}64$}
        \label{sfig:fft_64}
    \end{subfigure}
    \hfill
    \begin{subfigure}[t]{0.19\linewidth}
        \includegraphics[width=\linewidth]{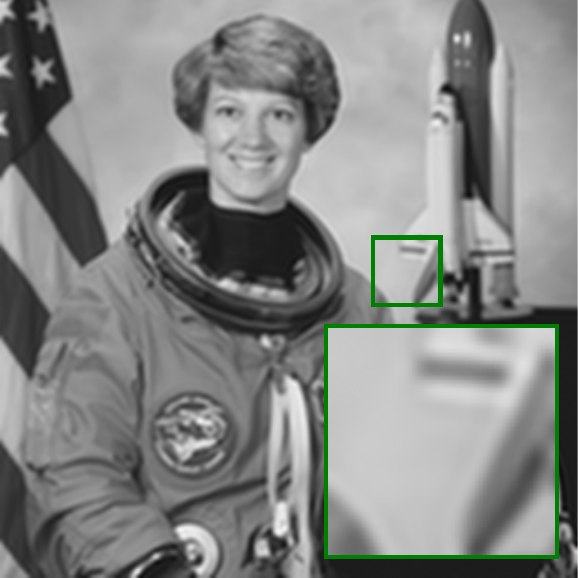}
        \caption*{\scriptsize Wavelet $r{=}64$}
        \label{sfig:dtcwpt_64}
    \end{subfigure}
    \caption{\textbf{Fourier- vs.\ Wavelet-domain low-pass filtering.} Fourier exhibits non-local ringing near edges, while DT-$\mathbb{C}$WPT preserves local geometry and edge structure more faithfully.}
    \label{fig:fft_vs_DTCWPT}
\end{figure}

Instantiating $\mathcal{T}$ as the Fourier Transform $\mathcal{F}$~\cite{zeng2025neuralremaster} is limited by global basis support, which propagates spectral modifications globally.
As a consequence, low-pass filtering induces non-local interference despite the use of smooth spectral attenuation, manifesting as Gibbs phenomenon~\cite{gottlieb1997gibbs} and boundary leakage~\cite{harris1978use} around high-contrast edges as shown in~\cref{fig:fft_vs_DTCWPT}.
During phase injection, these artifacts act as spurious structural constraints. The powerful diffusion prior then hallucinates realistic but incorrect geometry to accommodate these artifacts, degrading structural consistency.

\begin{figure*}[!tb]
    \centering
    \includegraphics[width=0.85\linewidth]{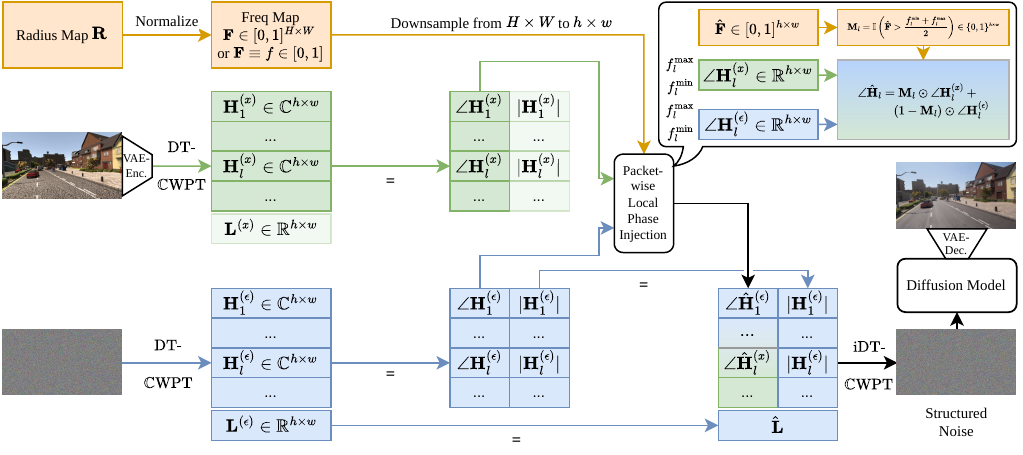}
    \caption{\textbf{Overview of $\psiPD$.} Source and noise latents are decomposed by DT-$\mathbb{C}$WPT into a low-frequency packet and multiple high-frequency packets. Source phase (green) is injected per packet under the cutoff map $\mathbf{F}$ while the low-frequency packet is randomized (LFR), and the inverse transform yields the structured noise $\hat{\boldsymbol{\epsilon}}$.}
    \label{fig:flow_chart}
\end{figure*}

To overcome these limitations, we instantiate $\mathcal{T}$ as DT-$\mathbb{C}$WPT~\cite{selesnick2008dual} with decomposition depth $J$ to decompose the latent $\mathbf{x}$ into a single real-valued low-frequency packet $\mathbf{L}$ and $L = 2^J - 1$ complex-valued high-frequency packets $\{\mathbf{H}_{l}\}_{l=1}^{L}$, indexed from high to low frequency.
This produces spatially localized coefficients and substantially reduces non-local interference compared with Fourier bases.

\subsection{Wavelet Phase Diffusion (\texorpdfstring{$\psi$-PD}{psiPD})}
\label{sec:wpd}

We construct structured noise by injecting source phase into a Gaussian noise sample in the DT-$\mathbb{C}$WPT domain.
An overview is depicted in~\cref{fig:flow_chart}.

We control structure preservation via a cutoff radius in radial frequency coordinates: a single global scalar $r$ in the simplest setting, or a radius map $\mathbf{R}$ for spatially adaptive control.
Concretely, we convert radii to a unitless cutoff map by normalizing with the Nyquist frequency $f_{\text{Nyq}}$:
\begin{equation}
\label{eq:cutoff_map_def}
\mathbf{F} = \operatorname{clip}\!\left(\tfrac{\mathbf{R}}{f_{\text{Nyq}}},0,1\right), \quad
f = \operatorname{clip}\!\left(\tfrac{r}{f_{\text{Nyq}}},0,1\right),
\end{equation}
where $\operatorname{clip}(x,a,b)=\min(\max(x,a),b)$.
Here $\mathbf{F}\in[0,1]^{H\times W}$ (and $f\in[0,1]$) indicates the highest normalized frequency up to which the latent phase should be preserved at each location.

\paragraph{Phase Extraction.}
We apply DT-$\mathbb{C}$WPT to both the source (green) and a Gaussian noise sample (blue), yielding a single real-valued low-frequency packet $\mathbf{L}$ and complex-valued high-frequency packets $\{\mathbf{H}_{l}\}_{l=1}^L$ that admit a direct magnitude--phase decomposition $\mathbf{H}_{l} = |\mathbf{H}_{l}|\,e^{j\angle \mathbf{H}_{l}}$.

\paragraph{Low-Frequency Randomization (LFR).}
\label{ssec:coarse_rand}
The $\mathbf{L}^{(x)}$ packet encodes the global illumination of the synthetic source latent, biasing the generative model toward the synthetic domain.
To decouple the model from this prior, we replace it with the noise low-frequency packet,
\begin{equation}
\hat{\mathbf{L}} = \mathbf{L}^{(\epsilon)},
\end{equation}
allowing the model to freely generate real-world coarse-scale appearance.

\paragraph{Packet-wise Local Phase Injection.}
For each high-frequency packet with support $[f_l^{\min},\allowbreak f_l^{\max}]$, we choose to inject the source phase based on the cutoff mask $\hat{\mathbf{F}}$.
This allows us to control the phase injection on a local level, enabling applications such as instance-level translation (\cf~\cref{ssec:instance}).
Using the midpoint frequency of each packet as a threshold, the phase is injected as:
\begin{equation}
\label{eq:phase_injection}
\begin{split}
\angle\hat{\mathbf{H}}_l &= \mathbf{M}_l \odot \angle{\mathbf{H}}^{(x)}_l + (1-\mathbf{M}_l) \odot \angle{\mathbf{H}}^{(\epsilon)}_l, \\
&\text{where } \mathbf{M}_l = \mathbb{I}\!\left(\hat{\mathbf{F}} > \tfrac{f_l^{\min} + f_l^{\max}}{2}\right),
\end{split}
\end{equation}
and $\mathbb{I}(\cdot)$ is the element-wise indicator function.
Typically, very high-frequency packets are dominated by noise phase, very low-frequency packets (except $\mathbf{L}$, handled by LFR) are dominated by source phase, and intermediate packets are mixed spatially depending on $\hat{\mathbf{F}}$.
After phase injection, we reconstruct the final noise $\hat{\boldsymbol{\epsilon}}$ using the inverse DT-$\mathbb{C}$WPT:
\begin{equation}
\hat{\boldsymbol{\epsilon}}
= \text{DT-}\mathbb{C}\text{WPT}^{-1}\!\Big(
\hat{\mathbf{L}},\,
\{\hat{\mathbf{H}}_{l}\}_{l=1}^{L}
\Big),
\end{equation}
where $\hat{\mathbf{H}}_l=\left|\mathbf{H}_l^{(\epsilon)}\right|e^{j\angle{\hat{\mathbf{H}}_l}}$.
In practice, DT-$\mathbb{C}$WPT is realized via a recursive DT-$\mathbb{C}$WT construction~\cite{cotter_2019} applied to all subbands, yielding the full wavelet packet decomposition.
A detailed comparison with NeuralRemaster~\cite{zeng2025neuralremaster} and implementation details are provided in~\suppref{supp:neuralremaster} and~\suppref{supp:algorithm}, respectively.

\section{Experiments}
\label{sec:experiments}
\subsection{Implementation Details}
\label{ssec:impl}
FLUX.1-dev~\cite{flux2024} serves as the image backbone and Wan~2.2-14B~\cite{wan2025} as the video backbone.
Both models are fine-tuned on unpaired open-domain datasets.
During training, the cutoff radius $r$ is sampled dynamically from a shifted exponential distribution following~\cite{zeng2025neuralremaster}.
Low-Frequency Randomization is applied with probability $p{=}0.8$; when applied, the decomposition depth $J$ is sampled uniformly in $[\max(J_{\mathrm{auto}}, 3),\, J_{\mathrm{auto}}+4]$, where $J_{\mathrm{auto}} = \lceil -\log_2(r / f_{\mathrm{Nyq}}) \rceil$.
Unless otherwise specified, at inference we use a \emph{single global scalar} cutoff radius $r$ (equivalently, a constant cutoff $\mathbf{F}$) set to half the Nyquist frequency of the latent space, with decomposition depth $J=4$.
We only use a \emph{spatially varying cutoff tensor} (radius map $\mathbf{R}$ and its normalized cutoff map $\mathbf{F}\in[0,1]^{H\times W}$) in~\cref{ssec:instance}.
Ablations on $r$ and $J$ are presented in~\cref{ssec:ablation} and~\suppref{supp:ablation_j}, respectively.

\subsection{Datasets and Metrics}
\label{ssec:datasets}

\paragraph{vKITTI $\to$ KITTI.}
Virtual KITTI (vKITTI)~\cite{vkitti} provides synthetic renders of five outdoor driving scenes with ground-truth depth and semantic labels.
We use 2{,}126 clone frames matched to real KITTI~\cite{kitti} tracking sequences.
We report KID and FID for realism, CLIP-IQA~\cite{clipiqa} for perceptual quality, DepSSIM (SSIM between the translated image's estimated depth and ground-truth) and AbsRel (absolute relative depth error)~\cite{zeng2025neuralremaster} for structural alignment, and mIoU via SegFormer~\cite{xie2021segformer} for semantic consistency.

\paragraph{CARLA.}
We follow~\cite{nucarla} and collect 60 driving sequences from the CARLA simulator~\cite{dosovitskiy2017carla} (109 frames each, across 3 towns), covering diverse road layouts and traffic scenarios.
We report semantically-matched patch KID (sKID) and FID (sFID)~\cite{Richter_2021}, which compute distribution distances on VGG-feature nearest-neighbor matched $128{\times}128$ patches against nuScenes~\cite{nuscenes2019}, providing a more discriminative realism signal than global KID/FID.
Beyond perceptual quality (CLIP-IQA) and temporal coherence via Motion Smoothness (MS)~\cite{huang2024vbench}, we evaluate downstream planning utility using ADE and FDE from the LightEMMA~\cite{lightemma} VLM planner (configured with a Gemini-2.5-Flash backbone) run on translated frames against simulator ground-truth trajectories. 
Specifically, the planner predicts a 3-second future trajectory conditioned on the current frame and the preceding 3-second trajectory. 
We restrict evaluation to the central 49 frames of each sequence at 5\,Hz (25 frames per sequence) to ensure complete availability of both past and future trajectories.
\subsection{Quantitative Results}
\paragraph{Image Translation.}
\cref{tab:img_results} summarizes quantitative performance on vKITTI $\to$ KITTI.
Training-free FlowEdit and DNAEdit achieve only modest realism gains and the lowest perceptual quality among translation methods, confirming that inference-time manipulation without a learned appearance prior cannot bridge the sim-to-real gap.
Cosmos Transfer~2.5 achieves the best DepSSIM ($0.870$) and AbsRel ($0.202$), but yields a lower CLIP-IQA ($0.469$) than other training-based methods and the second-worst mIoU ($39.36$), indicating that heavy conditioning preserves geometry at the cost of perceptual quality and semantic layout. 
We provide more results evaluating various Cosmos conditioning configurations in~\suppref{supp:cosmos_sensitivity}.
NeuralRemaster achieves a competitive CLIP-IQA ($0.527$) but the worst AbsRel ($0.344$) and lowest mIoU ($38.32$), consistent with global Fourier phase injection coupling semantically unrelated frequency bands.
Excluding the untranslated Input, which trivially upper-bounds mIoU and structural scores since it is unaltered, $\psiPD$ is the only method to rank first on three complementary axes simultaneously: realism (KID $4.41$), perceptual quality (CLIP-IQA $0.561$), and semantic consistency (mIoU $43.50$), while remaining competitive on FID and structural metrics.
This shows that localized wavelet phase injection achieves a superior realism--consistency operating point without any inference-time conditioning signal.

\begin{table}[ht]
    \caption{\textbf{Quantitative evaluations for image translation} (vKITTI $\to$ KITTI).
    \textbf{Bold} indicates best, \uline{underlined} second best.
    VACE and Ditto are video-native and thus not evaluated here. KID is reported $\times 10^2$.}
    \label{tab:img_results}
    \centering
    \footnotesize
    \setlength{\tabcolsep}{2pt}
    \begin{tabular}{l cccccc}
    \toprule
    Method & KID$\downarrow$ & FID$\downarrow$ & \makecell{CLIP-\\IQA$\uparrow$} & mIoU$\uparrow$ & \makecell{Dep-\\SSIM$\uparrow$} & AbsRel$\downarrow$ \\
    \midrule
    Input              & 6.06 & 97.29 & 0.281 & 50.39 & 0.900 & 0.157 \\
    \midrule
    FlowEdit~\cite{kulikov2025flowedit}   & 4.85 & 82.41 & 0.401 & \uline{42.72} & 0.812 & 0.260 \\
    DNAEdit~\cite{dnaedit}     & 4.78 & 85.47 & 0.322 & 41.22 & 0.827 & 0.254 \\
    NeuralRemaster~\cite{zeng2025neuralremaster} & 4.87 & 78.95 & \uline{0.527} & 38.32 & 0.811 & 0.344 \\
    Cosmos Transfer~2.5~\cite{cosmos}        & \uline{4.52} & \textbf{73.52} & 0.469 & 39.36 & \textbf{0.870} & \textbf{0.202} \\
    \midrule
    \psiPD\ (\textbf{Ours})      & \textbf{4.41} & \uline{73.84} & \textbf{0.561} & \textbf{43.50} & \uline{0.839} & \uline{0.229} \\
    \bottomrule
    \end{tabular}
\end{table}

\definecolor{myred}{RGB}{180,40,40}
\definecolor{mygreen}{RGB}{0,120,70}
\begin{table*}[tb]
    \centering
    \footnotesize
    \setlength{\tabcolsep}{3pt}
    \caption{\textbf{Quantitative evaluation on video translation} (CARLA).
    Relative changes w.r.t.\ the input are reported in \%.
    \textbf{Bold} indicates best, \uline{underlined} second best.
    FlowEdit is an image-only method and is not applicable to video. sKID is reported $\times 10^2$.}
    \label{tab:video_results}
    \begin{tabular}{@{}lccccccccccc@{}}
    \toprule
    \multirow{2}{*}{Method} & sKID & sFID & CLIP- & MS & \multicolumn{4}{c}{ADE (m) $\downarrow$} & FDE \\
    & $\downarrow$ & $\downarrow$ & IQA$\uparrow$ & (\%)$\uparrow$ & 1s & 2s & 3s & Avg. & (m)$\downarrow$ \\
    \midrule
    Input  & 1.97 & 48.89 & 0.367 & 98.58 & 0.520 & 2.008 & 4.484 & 2.337 & 5.223 \\
    \midrule
    DNAEdit~\cite{dnaedit}
        & 2.24 & 50.22 & 0.315 & 98.20
        & 0.581 \textcolor{myred}{(+11.8\%)}
        & 2.260 \textcolor{myred}{(+12.5\%)}
        & 5.039 \textcolor{myred}{(+12.4\%)}
        & 2.627 \textcolor{myred}{(+12.4\%)}
        & 5.872 \textcolor{myred}{(+12.4\%)} \\
    VACE~\cite{jiang2025vace}
        & 2.81 & 59.91 & 0.405 & \uline{98.47}
        & 0.503 \textcolor{mygreen}{(-3.3\%)}
        & \uline{1.952} \textcolor{mygreen}{(-2.8\%)}
        & \uline{4.385} \textcolor{mygreen}{(-2.2\%)}
        & \uline{2.280} \textcolor{mygreen}{(-2.5\%)}
        & \uline{5.116} \textcolor{mygreen}{(-2.1\%)} \\
    Cosmos Transfer~2.5~\cite{cosmos}
        & 2.42 & 51.98 & 0.429 & 97.84
        & 0.516 \textcolor{mygreen}{(-0.9\%)}
        & 1.999 \textcolor{mygreen}{(-0.5\%)}
        & 4.444 \textcolor{mygreen}{(-0.9\%)}
        & 2.319 \textcolor{mygreen}{(-0.8\%)}
        & 5.173 \textcolor{mygreen}{(-1.0\%)} \\
    NeuralRemaster~\cite{zeng2025neuralremaster}
        & 1.45 & 37.65 & 0.450 & 98.30
        & 0.532 \textcolor{myred}{(+2.2\%)}
        & 2.079 \textcolor{myred}{(+3.5\%)}
        & 4.630 \textcolor{myred}{(+3.2\%)}
        & 2.413 \textcolor{myred}{(+3.3\%)}
        & 5.389 \textcolor{myred}{(+3.2\%)} \\
    Ditto~\cite{ditto}
        & \uline{1.43} & \textbf{33.99} & \textbf{0.496} & 98.01
        & \uline{0.495} \textcolor{mygreen}{(-4.8\%)}
        & 2.022 \textcolor{myred}{(+0.7\%)}
        & 4.635 \textcolor{myred}{(+3.4\%)}
        & 2.384 \textcolor{myred}{(+2.0\%)}
        & 5.424 \textcolor{myred}{(+3.8\%)} \\
    \midrule
    \psiPD\ (\textbf{Ours})
        & \textbf{1.35} & \uline{36.58} & \uline{0.495} & \textbf{98.58}
        & \textbf{0.485} \textcolor{mygreen}{(-6.7\%)}
        & \textbf{1.898} \textcolor{mygreen}{(-5.5\%)}
        & \textbf{4.251} \textcolor{mygreen}{(-5.2\%)}
        & \textbf{2.211} \textcolor{mygreen}{(-5.4\%)}
        & \textbf{4.959} \textcolor{mygreen}{(-5.1\%)} \\
    \bottomrule
    \end{tabular}
\end{table*}

\begin{figure*}[tb]
    \centering
    \input{figures/qualitative.tex}
    \caption{\textbf{Qualitative comparison on vKITTI $\to$ KITTI.} Competing methods either hallucinate structure or distort lane/vehicle semantics; $\psiPD$ preserves both while achieving realistic appearance. See \cref{ssec:qualitative} for per-method analysis.}
    \label{fig:qualitative}
\end{figure*}

\begin{figure*}[tb]
    \centering
    \input{figures/qualitative_video.tex}
    \caption{\textbf{Qualitative comparison on CARLA video translation.} Baselines either retain CARLA's synthetic look or trade realism for altered semantics (weather, traffic lights, lanes); $\psiPD$ matches Ditto's realism while keeping semantics intact. See \cref{ssec:qualitative}.}
    \label{fig:qualitative_video}
\end{figure*}

\paragraph{Video Translation.}
\cref{tab:video_results} presents quantitative results on CARLA video translation.
$\psiPD$ achieves the best sKID ($1.35$) and second-best sFID ($36.58$), which is comparable to the paired-data-based baseline Ditto.
Among other baselines, DNAEdit decreases both perceptual quality and realism, which worsens planning metrics substantially ($+12.4\%$ average ADE).
Conditioning-based methods VACE and Cosmos Transfer~2.5 increase perceptual quality modestly and thus lead to downstream planning improvements ($-2.5\%$ and $-0.8\%$ average ADE, respectively).
While NeuralRemaster and Ditto translate videos closer to the realism domain, they result in worse planning results ($+3.3\%$ and $+2.0\%$ average ADE, respectively) due to structural and semantic inconsistency as discussed in~\cref{ssec:qualitative}.
$\psiPD$ is the only evaluated method that simultaneously improves realism, perceptual quality, and downstream planning (reducing average ADE by $5.4\%$ and FDE by $5.1\%$) while preserving the original temporal coherence of the simulator.

\subsection{Qualitative Results}
\label{ssec:qualitative}

\cref{fig:qualitative} shows qualitative comparisons on vKITTI $\to$ KITTI.
FlowEdit and NeuralRemaster both hallucinate an ego-vehicle dashboard in Example 1.
NeuralRemaster additionally reverses the orientation of oncoming vehicles, making them face the wrong way in Example 2.
Cosmos Transfer~2.5 avoids these hallucinations but introduces incorrect lane semantics in Example 3 and yields a desaturated appearance.
$\psiPD$ produces output without such artifacts.

\begin{figure}[ht]
    \centering
    \input{figures/vlm.tex}
    \caption{\textbf{VLM planning on translated CARLA frames.}
    Predicted waypoints ({\color{red}red}) vs.\ ground-truth ({\color{green!60!black}green}).}
    \label{fig:vlm}
\end{figure}

\cref{fig:qualitative_video} shows CARLA video comparisons across two scenes.
DNAEdit fails to escape CARLA's synthetic appearance.
Conditioning-based methods change the appearance more aggressively but introduce their own artifacts: VACE produces a strong lens-flare and severe colorful road-surface artifacts, while Cosmos Transfer 2.5 yields a cleaner but still synthetic-looking result.
NeuralRemaster improves local texture yet retains CARLA's warm sunset illumination.
Ditto achieves photorealistic texture but disregards scene semantics: it alters weather conditions and introduces incorrect traffic light states.
$\psiPD$ produces output comparable in realism to Ditto.
\cref{fig:vlm} shows that $\psiPD$'s translated frames yield VLM waypoints closely aligned with ground truth, while Ditto's translation changes the traffic light state to red.
The VLM planner responds by producing a large deviation from the ground-truth trajectory. 
Detailed chain-of-thought logs are provided in~\suppref{supp:vlm}.

\begin{figure*}[t]
    \centering
    \input{figures/application_qualitative.tex}
    \caption{\textbf{Zero-shot instance-level video translation.}
    The task requires only the character's mirror reflection (highlighted in green in the first input frame) to appear photorealistic while the surrounding comic-style scene remains untranslated.
    Ditto translates the entire frame indiscriminately and cannot isolate the edit to the reflection region; $\psiPD$ confines the translation to the target region via a per-pixel map.}
    \label{fig:application}
\end{figure*}

\begin{figure}[t]
    \centering
    \resizebox{\linewidth}{!}{\usetikzlibrary{calc}

\definecolor{cPPD}{HTML}{f59e0b}    %
\definecolor{cWPD}{HTML}{9333ea}    %
\definecolor{cOurs}{HTML}{60a5fa}   %
\definecolor{cInput}{HTML}{ef4444}  %

\pgfplotsset{
  abpanel/.style={
    width=5.5cm, height=4.0cm,
    grid=both,
    grid style={line width=.1pt, draw=gray!15},
    major grid style={line width=.2pt, draw=gray!50},
    tick align=outside, tick pos=left,
    label style={font=\footnotesize},
    tick label style={font=\scriptsize},
    clip=false,
  },
  abkid/.style={
    xmin=0.035, xmax=0.065, x dir=reverse,
    xtick={0.040,0.050,0.060},
    scaled x ticks=false,
  },
  abclip/.style={
    xmin=0.25, xmax=0.58,
    xtick={0.40,0.50},
    xticklabel style={/pgf/number format/fixed,
                      /pgf/number format/precision=2},
    x coord trafo/.code={
      \pgfmathparse{##1 < 0.4 ? (##1 < 0.3 ? ##1 : 0.3 + (##1-0.3)*0.2) : ##1 - 0.08}
    },
    x coord inv trafo/.code={
      \pgfmathparse{##1 < 0.32 ? (##1 < 0.3 ? ##1 : 0.3 + (##1-0.3)/0.2) : ##1 + 0.08}
    },
    after end axis/.append code={
      \pgfmathsetmacro{\ybottom}{\pgfkeysvalueof{/pgfplots/ymin}}
      \pgfmathsetmacro{\ytop}{\pgfkeysvalueof{/pgfplots/ymax}}
      \fill[white] ($(axis cs:0.32,\ybottom) + (0,-4pt)$) rectangle ($(axis cs:0.38,\ytop) + (0,4pt)$);
      \draw[gray!50, line width=.2pt, densely dashed] (axis cs:0.32,\ybottom) -- (axis cs:0.32,\ytop);
      \draw[gray!50, line width=.2pt, densely dashed] (axis cs:0.38,\ybottom) -- (axis cs:0.38,\ytop);
      \draw[black, line width=0.6pt] ($(axis cs:0.32,\ybottom) + (-2pt,-4pt)$) -- ($(axis cs:0.32,\ybottom) + (2pt,4pt)$);
      \draw[black, line width=0.6pt] ($(axis cs:0.38,\ybottom) + (-2pt,-4pt)$) -- ($(axis cs:0.38,\ybottom) + (2pt,4pt)$);
      \draw[black, line width=0.6pt] ($(axis cs:0.32,\ytop) + (-2pt,-4pt)$) -- ($(axis cs:0.32,\ytop) + (2pt,4pt)$);
      \draw[black, line width=0.6pt] ($(axis cs:0.38,\ytop) + (-2pt,-4pt)$) -- ($(axis cs:0.38,\ytop) + (2pt,4pt)$);
    }
  },
  abmiou/.style={
    ymin=31, ymax=52, ytick={35,39,43,47,51},
  },
  abdep/.style={
    ymin=0.760, ymax=0.910, ytick={0.78,0.82,0.86,0.90},
    scaled y ticks=false,
    yticklabel style={/pgf/number format/fixed,
                      /pgf/number format/precision=2},
    yticklabels={.78,.82,.86,.90},
  },
  }

  \begin{tikzpicture}
\def\INPUTkid{0.0606}   \def\INPUTclip{0.2808}
\def\INPUTmiou{50.39}   \def\INPUTdep{0.9002}

\begin{groupplot}[
  group style={
    group size=2 by 2,
    horizontal sep=0.3cm,
    vertical sep=0.4cm,
    x descriptions at=edge bottom,
    y descriptions at=edge left,
  },
  abpanel,
]

\nextgroupplot[abkid, abdep,
  ylabel={Dep-SSIM $\uparrow$},
  legend to name=grouplegend,
  legend style={legend columns=4, font=\scriptsize, draw=none, fill=none,
                cells={anchor=west},
                /tikz/every even column/.append style={column sep=6pt}},
]
  \addplot[cInput, only marks, mark=asterisk, thick, mark size=3.5pt]
    coordinates {(\INPUTkid,\INPUTdep)};
  \addlegendentry{Input}
  \addplot[cPPD, thick, mark=square, mark size=2pt,
    mark options={fill=white,draw=cPPD,line width=0.8pt}]
    coordinates {(0.0527,0.7809)(0.0487,0.8107)(0.0474,0.8378)(0.0498,0.8470)};
  \addlegendentry{NeuralRemaster}
  \addplot[cWPD, thick, mark=*, mark size=2.2pt,
    mark options={fill=white,draw=cWPD,line width=0.8pt}]
    coordinates {(0.0455,0.7725)(0.0450,0.8335)(0.0621,0.8688)};
  \addlegendentry{$\psi$-PD w/o LFR}
  \addplot[cOurs, line width=1.3pt, mark=triangle*, mark size=3pt,
    mark options={fill=cOurs,draw=cOurs,line width=0.6pt}]
    coordinates {(0.0361,0.7883)(0.0441,0.8394)(0.0485,0.8532)(0.0592,0.8763)};
  \addlegendentry{$\psi$-PD (Ours)}

\nextgroupplot[abclip, abdep, yticklabels=\empty]
  \addplot[cInput, only marks, mark=asterisk, thick, mark size=3.5pt]
    coordinates {(\INPUTclip,\INPUTdep)};
  \addplot[cPPD, thick, mark=square, mark size=2pt,
    mark options={fill=white,draw=cPPD,line width=0.8pt}]
    coordinates {(0.5347,0.7809)(0.5267,0.8107)(0.5179,0.8378)(0.5152,0.8470)};
  \addplot[cWPD, thick, mark=*, mark size=2.2pt,
    mark options={fill=white,draw=cWPD,line width=0.8pt}]
    coordinates {(0.5625,0.7725)(0.5532,0.8335)(0.5122,0.8688)(0.4060,0.8814)};
  \addplot[cOurs, line width=1.3pt, mark=triangle*, mark size=3pt,
    mark options={fill=cOurs,draw=cOurs,line width=0.6pt}]
    coordinates {(0.5670,0.7883)(0.5607,0.8394)(0.5212,0.8532)(0.4094,0.8763)};

\nextgroupplot[abkid, abmiou,
  xlabel={KID ($\times 10^2$) $\downarrow$},
  ylabel={mIoU (\%) $\uparrow$},
  xticklabels={4,5,6},
]
  \addplot[cInput, only marks, mark=asterisk, thick, mark size=3.5pt]
    coordinates {(\INPUTkid,\INPUTmiou)};
  \addplot[cPPD, thick, mark=square, mark size=2pt,
    mark options={fill=white,draw=cPPD,line width=0.8pt}]
    coordinates {(0.0527,32.65)(0.0487,38.32)(0.0474,43.75)(0.0498,44.87)};
  \addplot[cWPD, thick, mark=*, mark size=2.2pt,
    mark options={fill=white,draw=cWPD,line width=0.8pt}]
    coordinates {(0.0455,32.73)(0.0450,44.23)(0.0621,48.32)};
  \addplot[cOurs, line width=1.3pt, mark=triangle*, mark size=3pt,
    mark options={fill=cOurs,draw=cOurs,line width=0.6pt}]
    coordinates {(0.0361,38.11)(0.0441,43.50)(0.0485,45.04)(0.0592,44.98)};

\nextgroupplot[abclip, abmiou,
  xlabel={CLIP-IQA $\uparrow$},
]
  \addplot[cInput, only marks, mark=asterisk, thick, mark size=3.5pt]
    coordinates {(\INPUTclip,\INPUTmiou)};
  \addplot[cPPD, thick, mark=square, mark size=2pt,
    mark options={fill=white,draw=cPPD,line width=0.8pt}]
    coordinates {(0.5347,32.65)(0.5267,38.32)(0.5179,43.75)(0.5152,44.87)};
  \addplot[cWPD, thick, mark=*, mark size=2.2pt,
    mark options={fill=white,draw=cWPD,line width=0.8pt}]
    coordinates {(0.5625,32.73)(0.5532,44.23)(0.5122,48.32)(0.4060,48.46)};
  \addplot[cOurs, line width=1.3pt, mark=triangle*, mark size=3pt,
    mark options={fill=cOurs,draw=cOurs,line width=0.6pt}]
    coordinates {(0.5670,38.11)(0.5607,43.50)(0.5212,45.04)(0.4094,44.98)};

\end{groupplot}

\path let \p1=(current bounding box.west),
          \p2=(current bounding box.east),
          \p3=(group c1r2.south)
  in node[anchor=north] at ({0.5*(\x1+\x2)},{\y3-1.05cm})
       {\pgfplotslegendfromname{grouplegend}};

\end{tikzpicture}}
    \caption{\textbf{Ablation: Pareto frontiers (realism vs.\ consistency).}}
    \label{fig:ablation1}
\end{figure}

\subsection{Zero-Shot Instance-Level Translation}
\label{ssec:instance}

The spatial cutoff map $\mathbf{F}$ makes instance-level translation zero-shot, with no instance-level supervision.
Lowering the cutoff over a target region renders only that object photorealistic while the rest stays untranslated.
The construction only assumes a geometrically-reliable but non-photorealistic source, so it works beyond simulation: \cref{fig:application} translates a single instance in a comic-style scene.
Unlike Ditto, which alters the whole frame, $\psiPD$ stays per-pixel selective at no extra inference or architectural cost.
Additional results are in~\suppref{supp:instance}.

\subsection{Ablation Study}
\label{ssec:ablation}

We ablate two design choices using vKITTI $\to$ KITTI.
$\psiPD$ w/o LFR is a separately trained model in which the $\mathbf{L}$ packet retains source phase via Fourier phase injection rather than being randomized.
Comparing $\psiPD$ w/o LFR to NeuralRemaster therefore isolates the benefit of DT-$\mathbb{C}$WPT over Fourier phase injection; comparing full $\psiPD$ to $\psiPD$ w/o LFR isolates the benefit of LFR.

\paragraph{DT-$\mathbb{C}$WPT vs.\ Fourier Phase Injection.}
We sweep $r$ and plot the realism--consistency Pareto frontier.
NeuralRemaster's frontier is constrained: at comparable KID, it achieves strictly lower mIoU than $\psiPD$ w/o LFR, consistent with global spectral coupling.
This gap reflects the downstream impact of false structure injection, where the diffusion prior hallucinates realistic but incorrect geometry to accommodate spurious Fourier low-pass artifacts, degrading consistency.
$\psiPD$ w/o LFR shifts both axes outward via compact wavelet support.

\paragraph{Low-Frequency Randomization.}
Full $\psiPD$ extends the frontier to the right toward higher realism (lower KID and higher CLIP-IQA in~\cref{fig:ablation1}), with every radius step yielding a monotone realism gain.
Randomizing the $\mathbf{L}$ packet frees the model from locking to the synthetic illumination prior regardless of the high-frequency treatment as can be observed in \cref{fig:qualitative_video}.

\paragraph{Inference Overhead.}
On an RTX 6000 Ada at $704{\times}1280$, the $\psiPD$ noise construction takes $0.073$\,s for images (vs.\ $63.5$\,s for 50-step FLUX inference) and $0.096$\,s for video (vs.\ $204$\,s for 4-step, 49-frame Wan inference), i.e.\ under $0.15\%$ of total inference time, confirming negligible overhead.
In contrast, conditioning-based methods add dedicated control modules that incur substantial per-step inference cost on top of the backbone.

\section{Conclusion}
\label{sec:conclusion}

We introduced $\psiPD$, a phase-preserving diffusion framework for sim-to-real translation that achieves spatially adaptive structure preservation via DT-$\mathbb{C}$WPT-domain phase injection.
Low-Frequency Randomization of the $\mathbf{L}$ packet decouples the model from synthetic global illumination, enabling in-distribution real-world appearance.
On vKITTI $\to$ KITTI, $\psiPD$ outperforms prior methods in realism and semantic consistency while maintaining competitive structural alignment.
On CARLA video translation, it approaches paired-data perceptual quality while reducing VLM planner ADE by $5.4\%$, demonstrating direct downstream benefit for trajectory planning.
The spatially adaptive design further supports instance-level translation, rendering individual objects photorealistically while leaving the surrounding scene untranslated.
$\psiPD$ is thus broadly applicable to closed-loop evaluation and controllable scene editing pipelines.

\section{Acknowledgements}
\label{sec:acknowledgements}

The authors gratefully acknowledge the computing time provided
on the high-performance computer HoreKa by the National
High-Performance Computing Center at KIT (NHR@KIT).
This center is jointly supported by the Federal Ministry of
Research, Technology and Space and the Ministry of Science,
Research and the Arts of Baden-Württemberg, as part of the
National High-Performance Computing (NHR) joint funding program.
HoreKa is partly funded by the German Research Foundation (DFG).

The authors thank the Gauss Centre for Supercomputing e.V. (GCS)
for providing computing time on the Supercomputer JUWELS
\cite{jsc2021juwels} at the Jülich Supercomputing Centre.
JUWELS is supported by GCS through funding by the German
Federal Ministry of Research, Technology and Space and the
Ministry of Culture and Science of the State of North
Rhine-Westphalia.

{
    \small
    \bibliographystyle{ieeenat_fullname}
    \bibliography{main}
}

\ifdefined\arxivversion
\clearpage
\begingroup
\makeatletter
\def\@title{Supplementary Material:\\
Wavelet Phase Diffusion for Structurally and Semantically Consistent Sim-to-Real Translation}
\def\@author{
Kaiwen Wang$^{1}$ \quad Frank Bieder$^{2}$ \quad Yinzhe Shen$^{1}$\\
Carlos Fernandez$^{1}$ \quad Jan-Hendrik Pauls$^{1}$ \quad Omer Sahin Tas$^{2}$\\
{$^{1}$Karlsruhe Institute of Technology, Germany}\\
{$^{2}$FZI Research Center for Information Technology, Germany}\\
{\tt\small kaiwen.wang@kit.edu}
}
\let\@maketitle\orig@@maketitle
\makeatother
\origmaketitle
\endgroup
\setcounter{section}{0}
\setcounter{figure}{0}
\setcounter{table}{0}
\setcounter{equation}{0}
\setcounter{algorithm}{0}
\renewcommand{\thesection}{S\arabic{section}}
\renewcommand{\thefigure}{S\arabic{figure}}
\renewcommand{\thetable}{S\arabic{table}}
\renewcommand{\theequation}{S\arabic{equation}}
\renewcommand{\thealgorithm}{S\arabic{algorithm}}
\makeatletter
\renewcommand{\theHsection}{supp.\arabic{section}}
\renewcommand{\theHfigure}{supp.\arabic{figure}}
\renewcommand{\theHtable}{supp.\arabic{table}}
\renewcommand{\theHequation}{supp.\arabic{equation}}
\renewcommand{\theHalgorithm}{supp.\arabic{algorithm}}
\makeatother
\section{Pseudo-code for \texorpdfstring{$\psiPD$}{psiPD} Noise Construction}
\label{supp:algorithm}

Algorithm~\ref{alg:wavelet_phase_diffusion} details the $\psiPD$ noise construction.
It recursively injects localized source phase based on the given cutoff map $\mathbf{F}$, while decoupling synthetic illumination via Low-Frequency Randomization (LFR).

\begin{algorithm}[H]
\caption{$\psiPD$ Structured Noise Construction}
\label{alg:wavelet_phase_diffusion}
\begingroup\small
\begin{algorithmic}[1]
\Require Source latent $\mathbf{x}$, Gaussian noise $\boldsymbol{\epsilon}$, cutoff map $\mathbf{F}\in[0,1]^{H\times W}$, decomposition depth $J$ (default $4$)
\Ensure Structured noise $\hat{\boldsymbol{\epsilon}}$

\State $\mathbf{L}^{(x)},\{\mathbf{H}_l^{(x)}\}_{l=1}^J \gets \mathrm{DT\text{-}\mathbb{C}WT}(\mathbf{x},J)$
\State $\mathbf{L}^{(\epsilon)},\{\mathbf{H}_l^{(\epsilon)}\}_{l=1}^J \gets \mathrm{DT\text{-}\mathbb{C}WT}(\boldsymbol{\epsilon},J)$

\State $\hat{\mathbf{L}} \gets \mathbf{L}^{(\epsilon)}$ \Comment{Low-Frequency Randomization}

\For{$l \in \{1,\dots,J\}$}
  \State $\hat{\mathbf{H}}_l \gets \textsc{RecursiveInject}(\mathbf{H}_l^{(x)},\mathbf{H}_l^{(\epsilon)},\mathbf{F},[f_l^{\min},f_l^{\max}])$
\EndFor

\State $\hat{\boldsymbol{\epsilon}} \gets \mathrm{DT\text{-}\mathbb{C}WT}^{-1}\!\bigl(\hat{\mathbf{L}},\{\hat{\mathbf{H}}_l\}\bigr)$
\State \Return $\hat{\boldsymbol{\epsilon}}$

\Statex
\Function{RecursiveInject}{$\mathbf{H}^{(x)},\mathbf{H}^{(\epsilon)},\mathbf{F},[f^{\min},f^{\max}]$}
  \State $f^{\mathrm{mid}} \gets (f^{\min}+f^{\max})/2$
  \If{$f^{\max} \le \min(\mathbf{F})$} \Comment{inject source phase}
    \State \Return $|\mathbf{H}^{(\epsilon)}|\exp(j\angle\mathbf{H}^{(x)})$
  \ElsIf{$f^{\min} > \max(\mathbf{F})$} \Comment{pure noise}
    \State \Return $\mathbf{H}^{(\epsilon)}$
  \ElsIf{MaxDepthReached()} \Comment{Mixed spatially}
    \State $\mathbf{M} \gets \mathbf{F} > f^{\mathrm{mid}}$
    \State \Return $|\mathbf{H}^{(\epsilon)}|\exp\!\bigl(j(\mathbf{M}\odot\angle\mathbf{H}^{(x)}+(1{-}\mathbf{M})\odot\angle\mathbf{H}^{(\epsilon)})\bigr)$
  \Else \Comment{Recurse into sub-bands}
    \State $\mathbf{lo}^{(x)},\mathbf{hi}^{(x)} \gets \mathrm{DT\text{-}\mathbb{C}WT}(\mathbf{H}^{(x)},1)$
    \State $\mathbf{lo}^{(\epsilon)},\mathbf{hi}^{(\epsilon)} \gets \mathrm{DT\text{-}\mathbb{C}WT}(\mathbf{H}^{(\epsilon)},1)$
    \State $\hat{\mathbf{lo}} \gets \textsc{RecursiveInject}(\mathbf{lo}^{(x)},\mathbf{lo}^{(\epsilon)},\mathbf{F},[f^{\min},f^{\mathrm{mid}}])$
    \State $\hat{\mathbf{hi}} \gets \textsc{RecursiveInject}(\mathbf{hi}^{(x)},\mathbf{hi}^{(\epsilon)},\mathbf{F},[f^{\mathrm{mid}},f^{\max}])$
    \State \Return $\mathrm{Merge}(\hat{\mathbf{lo}},\hat{\mathbf{hi}})$
  \EndIf
\EndFunction
\end{algorithmic}
\endgroup
\end{algorithm}

\section{CLIP-IQA Prompt Configuration}
\label{supp:clipiqa}

We use a domain-specific ensemble of five sim-to-real antonym pairs in Table~\ref{tab:clipiqa_prompts} as CLIP anchors.
We refer to this configuration as \textbf{CLIP-IQA} throughout the paper.

\begin{table}[ht]
    \caption{\textbf{CLIP-IQA prompt pairs.} Each row is one antonym pair; scores are averaged over all five pairs.}
    \label{tab:clipiqa_prompts}
    \centering
    \footnotesize
    \setlength{\tabcolsep}{4pt}
    \begin{tabular}{ll}
    \toprule
    Positive (real) & Negative (synthetic) \\
    \midrule
    A photo of real cars on a road. & Computer graphics of cars on a road. \\
    Realistic materials. & Plastic-looking materials. \\
    A sharp real photo. & A blurry computer render. \\
    A photograph of a real place. & A CGI scene. \\
    A photo with natural colors. & An image with artificial colors. \\
    \bottomrule
    \end{tabular}
\end{table}

\begin{figure*}[t]
\centering
\noindent
\adjustbox{valign=t}{%
\begin{subfigure}[t]{0.30\linewidth}
  \centering
  \includegraphics[width=0.92\linewidth]{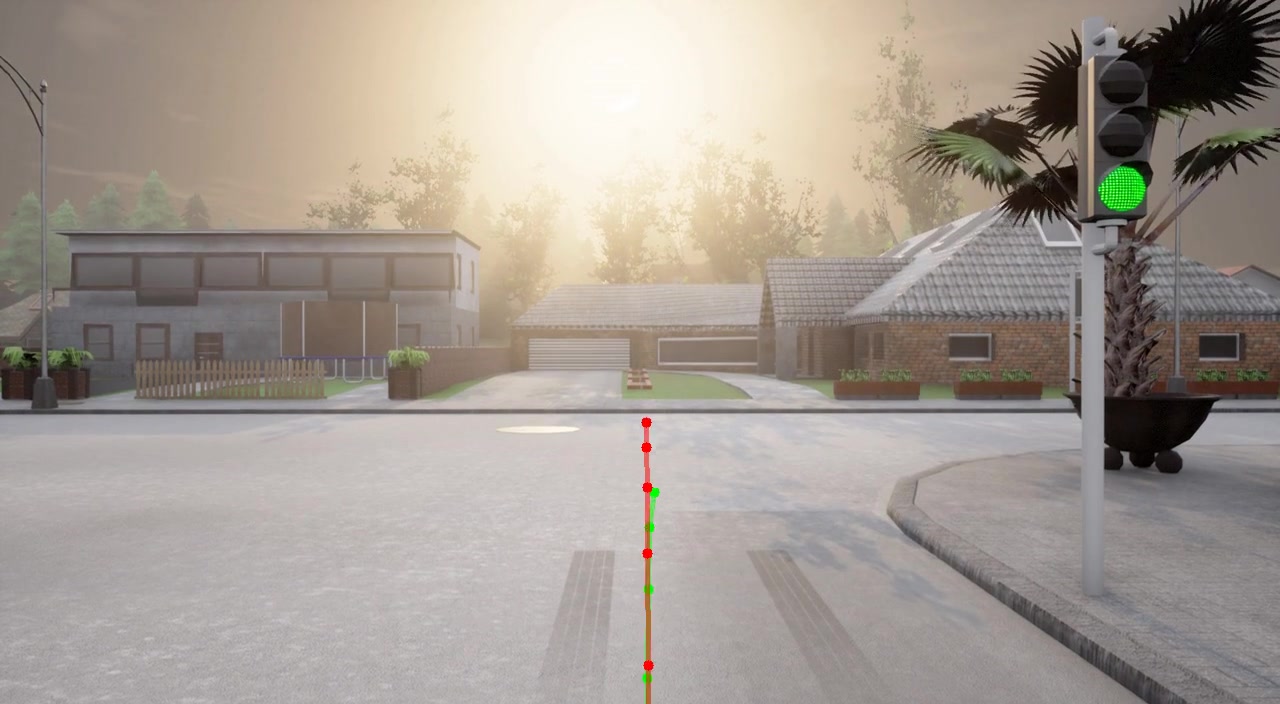}
  \caption*{\centering\footnotesize Input (raw sim)}
\end{subfigure}}\hfill
\adjustbox{valign=t}{%
\begin{minipage}[t]{0.68\linewidth}
  \raggedright\rmfamily\fontsize{7}{9}\selectfont
  \textit{Scene:} The front-view camera image displays a suburban street scene under hazy, bright conditions with no visible vehicles or pedestrians.
  The asphalt road surface shows two faint tire track markings in the foreground and a circular white road marking further ahead.
  On the right side of the road, a sidewalk with a curb is present, and a traffic light pole stands with its bottom light \textbf{illuminated green}.
  A large palm tree in a black pot is positioned next to the traffic light.
  In the background, there are several houses with brown brick facades and gray tiled gabled roofs.

  \smallskip
  \textit{Intent:} The ego's previous intent was to \textbf{decelerate significantly} from approximately 4.79\,m/s down to a near stop at 0.066\,m/s.
  Taking into account the ego's previous intent and the current scene, it should now \textbf{accelerate}, aiming to reach approximately \textbf{10\,m/s (3.3\,m/s² acceleration)} over the next 3 seconds, as the \textbf{traffic light is green} and there are no visible vehicles or pedestrians on the road.
\end{minipage}}

\vspace{3pt}

\noindent
\adjustbox{valign=t}{%
\begin{subfigure}[t]{0.30\linewidth}
  \centering
  \includegraphics[width=0.92\linewidth]{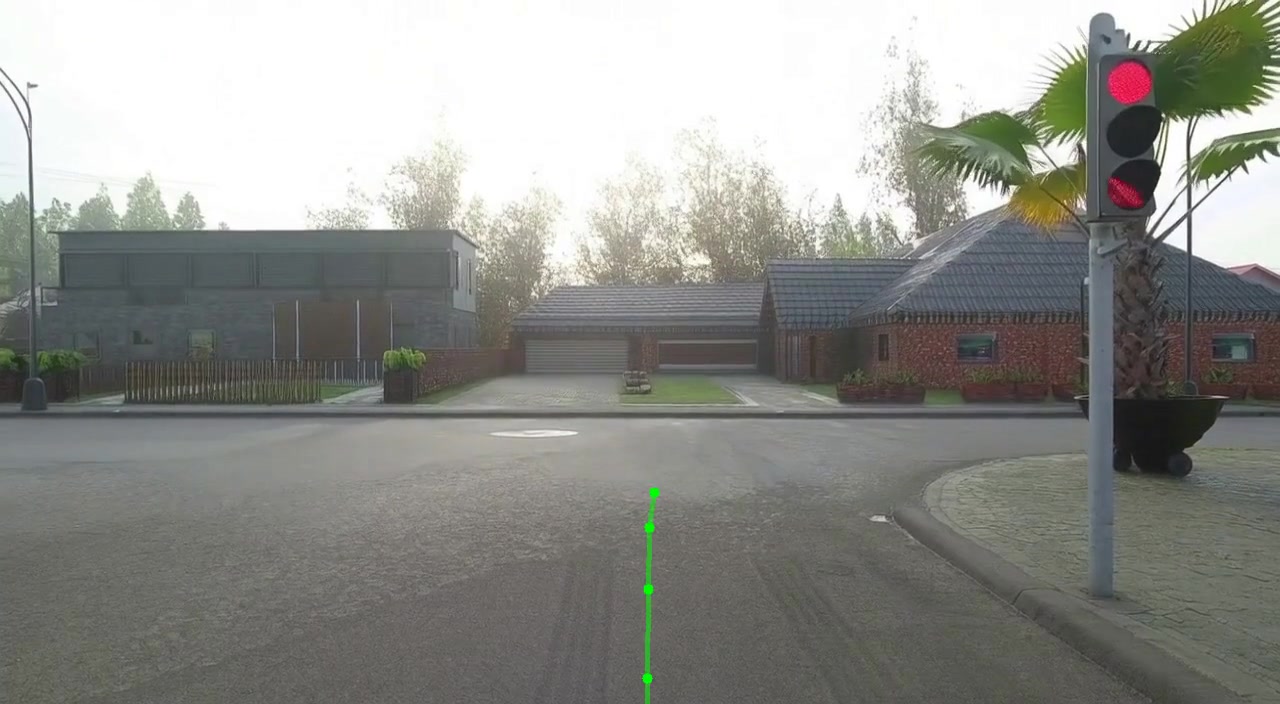}
  \caption*{\centering\footnotesize Ditto}
\end{subfigure}}\hfill
\adjustbox{valign=t}{%
\begin{minipage}[t]{0.68\linewidth}
  \raggedright\rmfamily\fontsize{7}{9}\selectfont
  \textit{Scene:} The camera view presents a residential street scene under bright, somewhat hazy daylight conditions.
  The asphalt road surface appears wet or reflective.
  On the far right, a vertical traffic light pole holds a traffic signal displaying a \textbf{solid red light at the top} and a \textbf{red downward arrow light illuminated below it}.
  No vehicles or pedestrians are present in the visible area.

  \smallskip
  \textit{Intent:} The ego's previous intent was to \textbf{decelerate significantly} from approximately 4.79\,m/s to a near stop.
  Taking into account the ego's previous intent and the current scene displaying a \textbf{solid red traffic light with a red downward arrow}, the ego should \textbf{decelerate} to come to a \textbf{complete stop at 0\,m/s} and then maintain that speed for the next 3 seconds.
\end{minipage}}

\vspace{3pt}

\noindent
\adjustbox{valign=t}{%
\begin{subfigure}[t]{0.30\linewidth}
  \centering
  \includegraphics[width=0.92\linewidth]{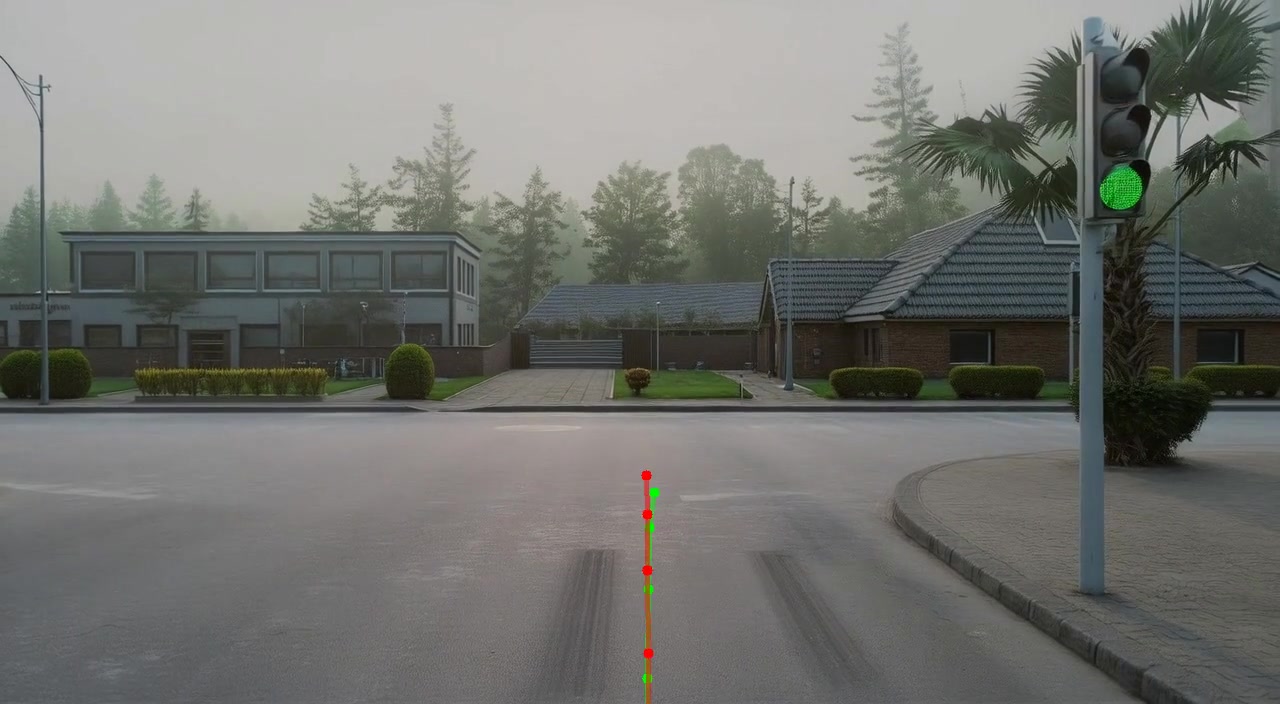}
  \caption*{\centering\footnotesize\textbf{$\psiPD$ (Ours)}}
\end{subfigure}}\hfill
\adjustbox{valign=t}{%
\begin{minipage}[t]{0.68\linewidth}
  \raggedright\rmfamily\fontsize{7}{9}\selectfont
  \textit{Scene:} The scene depicts a road on a foggy or hazy day, with visibility significantly reduced in the distant background.
  A traffic light pole is positioned on the right side of the road, and its vertical signal head shows a \textbf{green light illuminated at the bottom}.
  A large palm tree stands prominently next to the traffic light pole.
  No vehicles or pedestrians are visible in the immediate vicinity.

  \smallskip
  \textit{Intent:} The ego vehicle's previous intent was to \textbf{decelerate}, reducing its speed from 4.79\,m/s to a near stop at 0.066\,m/s.
  Taking into account the ego's previous intent and the current scene with a \textbf{green traffic light} and no visible obstacles, it should \textbf{accelerate}.
  A smooth acceleration of approximately \textbf{1.5 to 2\,m/s²} would be appropriate given the foggy conditions, which would increase its speed by 4.5 to 6\,m/s over the next 3 seconds.
\end{minipage}}
\caption{Qualitative results on downstream utility of translated videos.}
\label{fig:vlm_case1}
\end{figure*}

\begin{table}[ht]
    \caption{\textbf{Cosmos Transfer~2.5 sensitivity to control-signal choice} (vKITTI $\to$ KITTI).
    D=depth, E=edge, S=seg, V=vis.
    \textbf{Bold} indicates best, \uline{underlined} second best (excluding raw input). KID is reported $\times 10^2$.}
    \label{tab:cosmos_diverse}
    \centering
    \footnotesize
    \setlength{\tabcolsep}{2pt}
    \begin{tabular}{l cccccc}
    \toprule
    Controls & KID$\downarrow$ & FID$\downarrow$ & \makecell{CLIP-\\IQA$\uparrow$} & mIoU$\uparrow$ & \makecell{Dep-\\SSIM$\uparrow$} & AbsRel$\downarrow$ \\
    \midrule
    Input        & 6.06 & 97.29  & 0.281 & 50.39 & 0.900 & 0.157 \\
    \midrule
    D            & 5.41 & 77.18  & 0.466 & 38.37 & 0.852 & 0.216 \\
    D+E          & \uline{4.52} & \textbf{73.52} & 0.469 & 39.36 & 0.870 & 0.202 \\
    D+V          & 6.91 & 97.49  & 0.370 & 40.99 & 0.872 & \uline{0.163} \\
    D+S          & 6.94 & 92.34  & 0.427 & 36.11 & 0.870 & 0.202 \\
    S+E          & 5.52 & 79.78  & \uline{0.497} & 37.32 & 0.844 & 0.275 \\
    D+S+E        & 5.55 & 81.80  & 0.441 & 38.92 & \textbf{0.875} & 0.197 \\
    D+S+V        & 7.59 & 102.83 & 0.376 & 42.11 & \uline{0.874} & 0.165 \\
    D+S+V+E      & 7.05 & 98.59  & 0.376 & \textbf{44.22} & \textbf{0.875} & \textbf{0.161} \\
    \midrule
    \textbf{Ours} & \textbf{4.41} & \uline{73.84} & \textbf{0.561} & \uline{43.50} & 0.839 & 0.229 \\
    \bottomrule
    \end{tabular}
\end{table}

\section{Cosmos Transfer 2.5 Control Signals}
\label{supp:cosmos_sensitivity}

We evaluated eight combinations of Cosmos Transfer~2.5 control signals on vKITTI $\to$ KITTI in \cref{tab:cosmos_diverse}.
\textbf{D+E} is used in the main paper image translation results as it achieves the best KID (4.52) and FID (73.52) among Cosmos configurations; adding more signals improves mIoU but incurs sharply worse KID and FID, as denser conditioning anchors semantics at the cost of distributional realism.
For video translation, we use \textbf{D+S+V+E} to preserve traffic light colors and other semantic cues.

\section{Ablation: Effect of Decomposition Depth \texorpdfstring{$J$}{J}}
\label{supp:ablation_j}

\cref{tab:ablation_J_full} ablates decomposition depth $J$ at $r{=}12$ and $r{=}16$.
Larger $J$ improves structural metrics (DepSSIM, mIoU) at the cost of higher KID and lower CLIP-IQA.
At $r{=}12$, $J{=}5$ gains only $+$0.9\% DepSSIM and $+$3.0\% mIoU over $J{=}4$ while incurring $+$18.6\% KID and $-$4.4\% CLIP-IQA. On the other hand, while $J{=}3$ yields slightly better KID and CLIP-IQA, it leads to a severe degradation in semantic consistency, with mIoU dropping by 6.5\,pp (from 43.50\% to 37.00\%). Thus, $J{=}4$ is chosen as the default to best balance realistic textures and layout consistency.

\begin{table}[ht]
    \centering
    \caption{\textbf{Effect of decomposition depth $J$} at $r{=}12$ and $r{=}16$ on vKITTI $\to$ KITTI. $J{=}4$ (default) best balances realism (KID) and structural fidelity (DepSSIM, mIoU). KID is reported $\times 10^2$.}
    \label{tab:ablation_J_full}
    \begin{tabular}{cc cccc}
    \toprule
    $r$ & $J$ & KID$\downarrow$ & \makecell{CLIP-\\IQA$\uparrow$} & \makecell{Dep-\\SSIM$\uparrow$} & mIoU$\uparrow$ \\
    \midrule
    \multirow{3}{*}{12}
        & 3 & \textbf{4.19} & \textbf{0.593} & 0.818 & 37.00 \\
        & 4 & 4.41 & 0.561 & 0.839 & 43.50 \\
        & 5 & 5.23 & 0.536 & \textbf{0.847} & \textbf{44.81} \\
    \midrule
    \multirow{3}{*}{16}
        & 3 & \textbf{4.50} & \textbf{0.562} & 0.838 & 38.73 \\
        & 4 & 4.74 & 0.531 & 0.851 & 44.31 \\
        & 5 & 5.74 & 0.507 & \textbf{0.855} & \textbf{45.84} \\
    \bottomrule
    \end{tabular}
\end{table}

\section{VLM Planner Case Studies}
\label{supp:vlm}

\cref{fig:vlm_case1} shows full chain-of-thought logs for a representative example, illustrating how translation quality affects the planner's scene description and driving intent.
This comparison underscores that preserving fine-grained semantic states (such as traffic light signals) is crucial for sim-to-real translation.

\section{Additional Qualitative Results}
\label{supp:vkitti_qual}

\cref{fig:vkitti_teaser1}--\cref{fig:vkitti_s3} show qualitative comparisons on vKITTI $\to$ KITTI.
FlowEdit and DNAEdit are excluded due to worse realism and consistency. 
We refer the reader to the images and videos on our project page for further details and additional CARLA video translation results.

\newlength{\suppimgw}
\setlength{\suppimgw}{\textwidth}
\newcommand{\supprowlabel}[2]{%
  \node[anchor=east, inner sep=0pt, outer sep=0pt]
    at ($(#1.west)+(-0.5em,0)$)
    {\rotatebox{90}{\makebox[1.8cm][c]{\fontsize{6}{6}\selectfont #2}}};%
}

\begin{figure*}[p]
\centering
\begin{tikzpicture}[piccell/.style={inner sep=0pt,outer sep=0pt,anchor=north west}]
\node[inner sep=0pt,outer sep=0pt] at (0,0) {};
\node[piccell] (r1) at (0,0) {\includegraphics[width=\suppimgw]{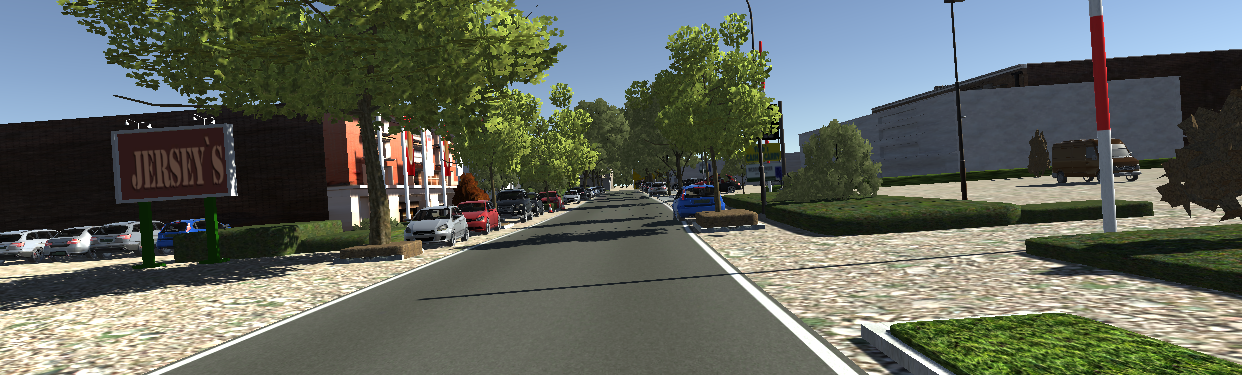}};
\supprowlabel{r1}{Input}
\node[piccell,below=0pt of r1] (r2) {\includegraphics[width=\suppimgw]{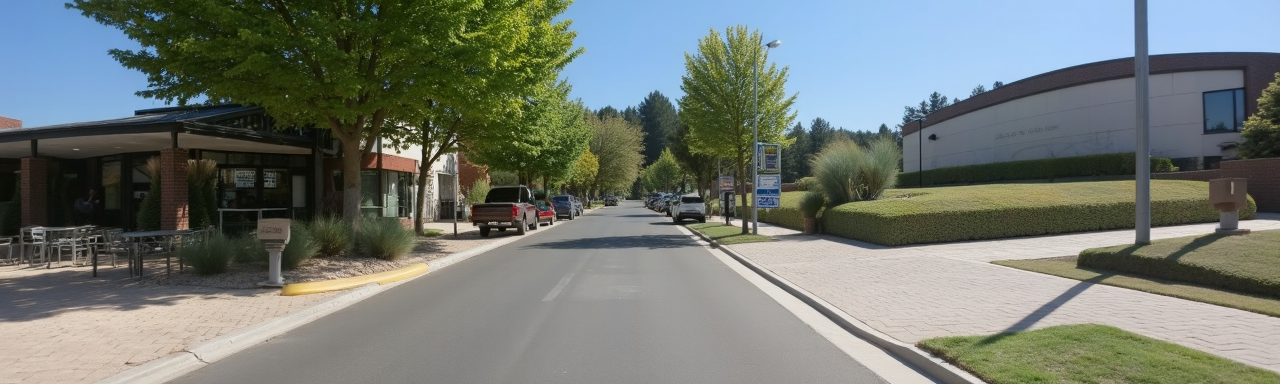}};
\supprowlabel{r2}{NeuralRemaster}
\node[piccell,below=0pt of r2] (r3) {\includegraphics[width=\suppimgw]{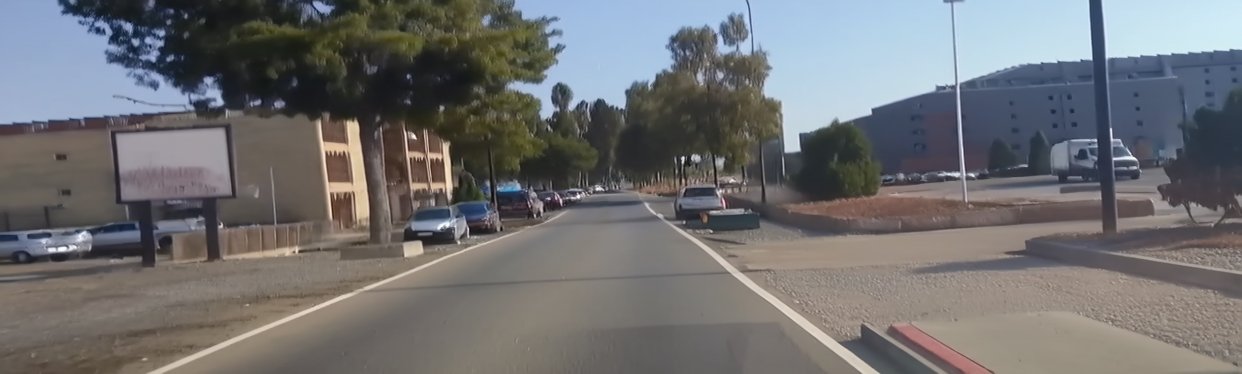}};
\supprowlabel{r3}{Cosmos Tr.~2.5}
\node[piccell,below=0pt of r3] (r4) {\includegraphics[width=\suppimgw]{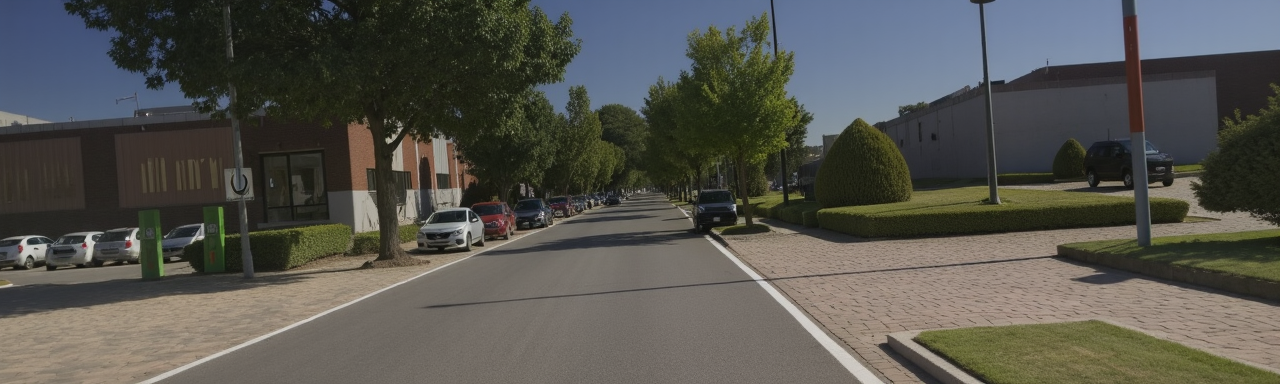}};
\supprowlabel{r4}{\textbf{$\psiPD$ (Ours)}}
\end{tikzpicture}
\caption{\textbf{vKITTI $\to$ KITTI qualitative comparison, Scene~1.}}
\label{fig:vkitti_teaser1}
\end{figure*}

\begin{figure*}[p]
\centering
\begin{tikzpicture}[piccell/.style={inner sep=0pt,outer sep=0pt,anchor=north west}]
\node[inner sep=0pt,outer sep=0pt] at (0,0) {};
\node[piccell] (r1) at (0,0) {\includegraphics[width=\suppimgw]{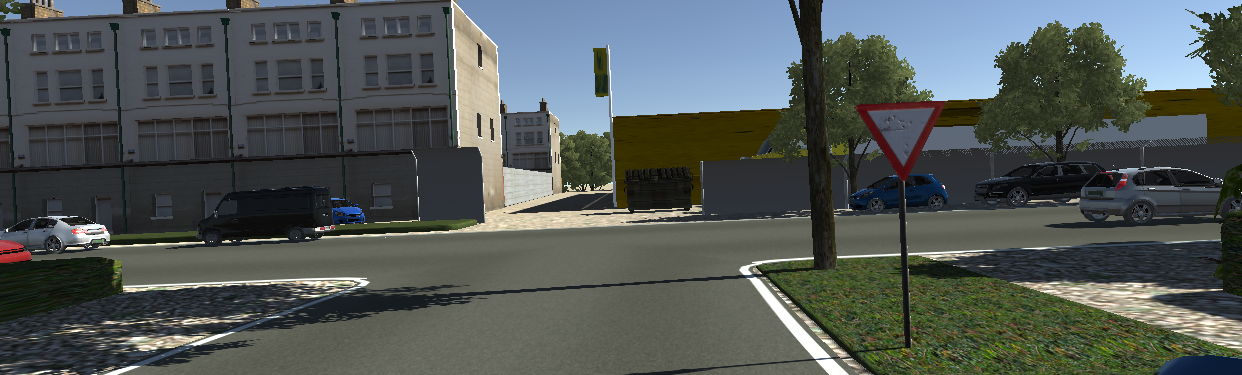}};
\supprowlabel{r1}{Input}
\node[piccell,below=0pt of r1] (r2) {\includegraphics[width=\suppimgw]{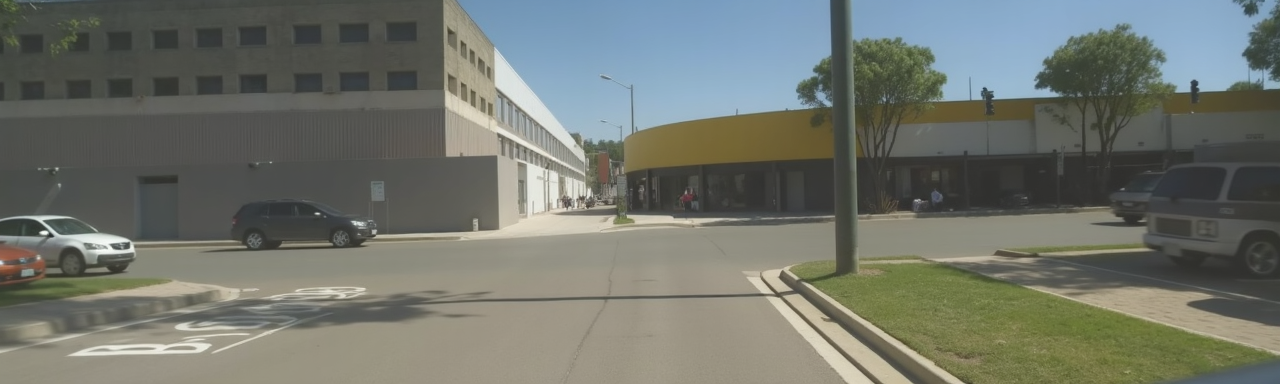}};
\supprowlabel{r2}{NeuralRemaster}
\node[piccell,below=0pt of r2] (r3) {\includegraphics[width=\suppimgw]{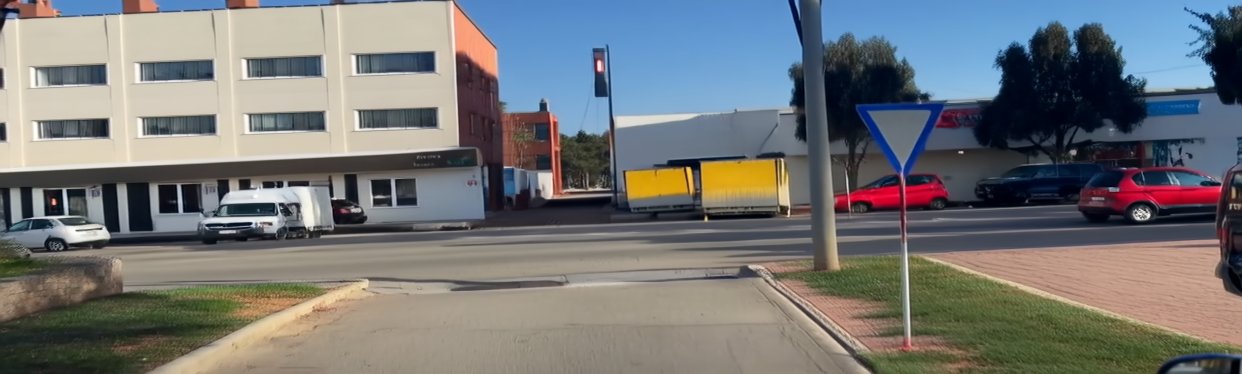}};
\supprowlabel{r3}{Cosmos Tr.~2.5}
\node[piccell,below=0pt of r3] (r4) {\includegraphics[width=\suppimgw]{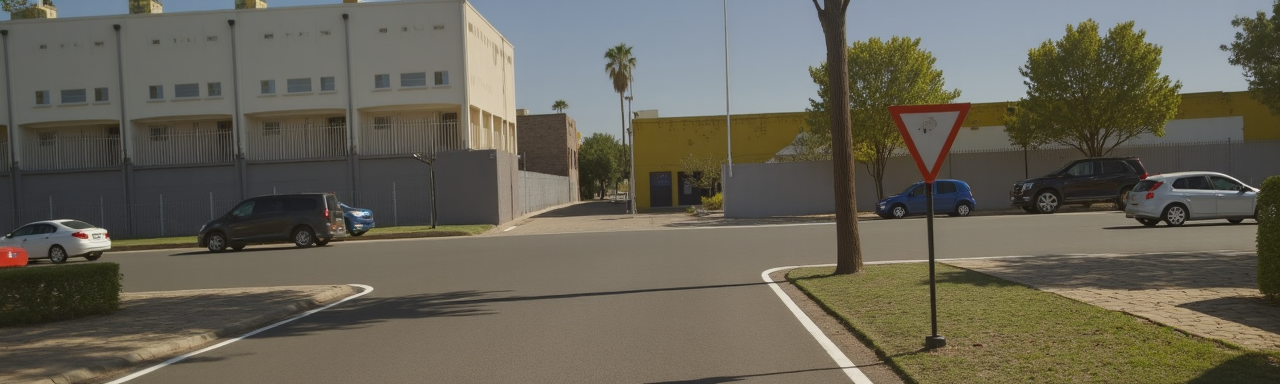}};
\supprowlabel{r4}{\textbf{$\psiPD$ (Ours)}}
\end{tikzpicture}
\caption{\textbf{vKITTI $\to$ KITTI qualitative comparison, Scene~2.}}
\label{fig:vkitti_teaser2}
\end{figure*}

\begin{figure*}[p]
\centering
\begin{tikzpicture}[piccell/.style={inner sep=0pt,outer sep=0pt,anchor=north west}]
\node[inner sep=0pt,outer sep=0pt] at (0,0) {};
\node[piccell] (r1) at (0,0) {\includegraphics[width=\suppimgw]{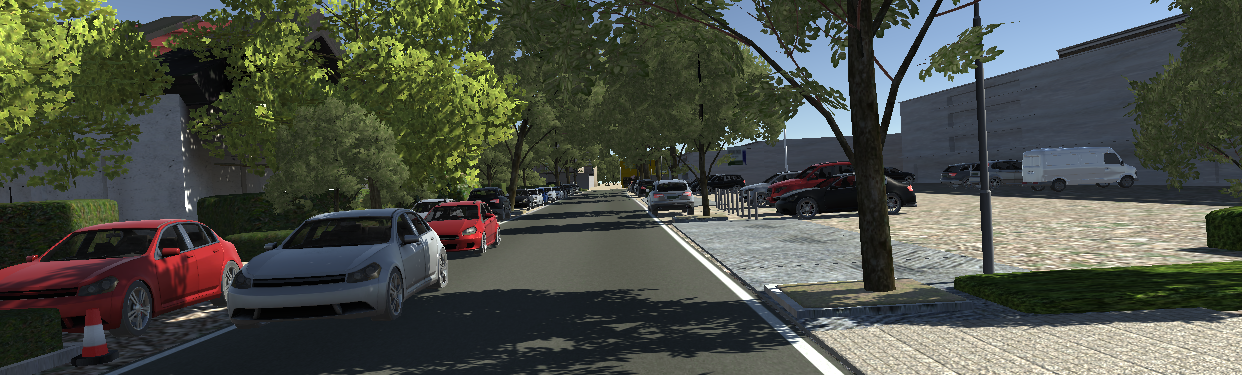}};
\supprowlabel{r1}{Input}
\node[piccell,below=0pt of r1] (r2) {\includegraphics[width=\suppimgw]{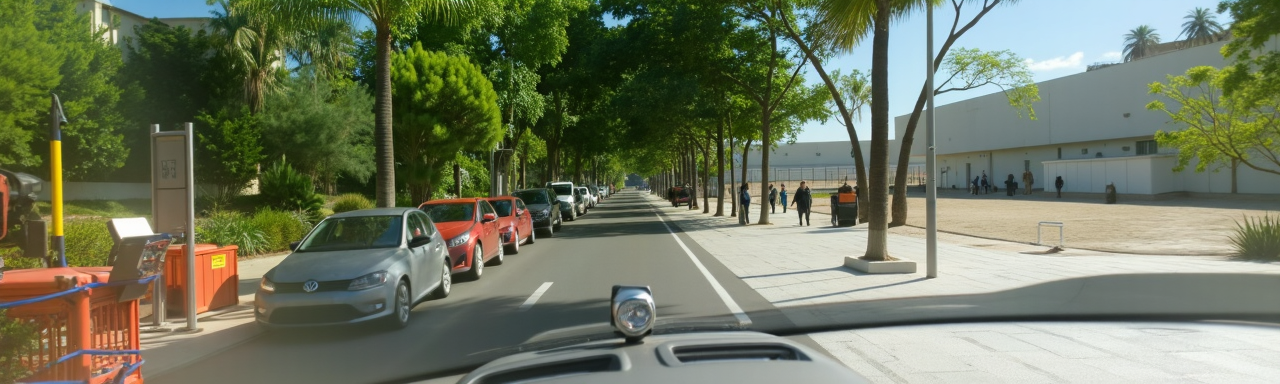}};
\supprowlabel{r2}{NeuralRemaster}
\node[piccell,below=0pt of r2] (r3) {\includegraphics[width=\suppimgw]{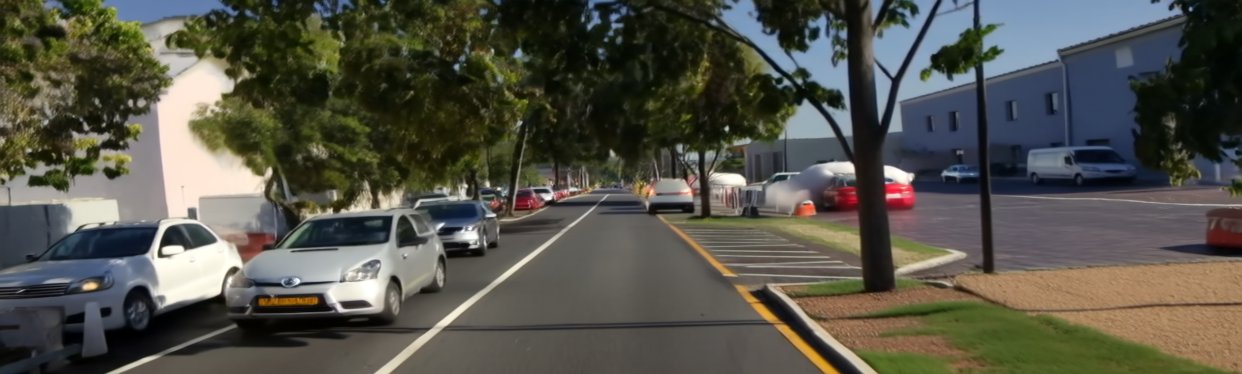}};
\supprowlabel{r3}{Cosmos Tr.~2.5}
\node[piccell,below=0pt of r3] (r4) {\includegraphics[width=\suppimgw]{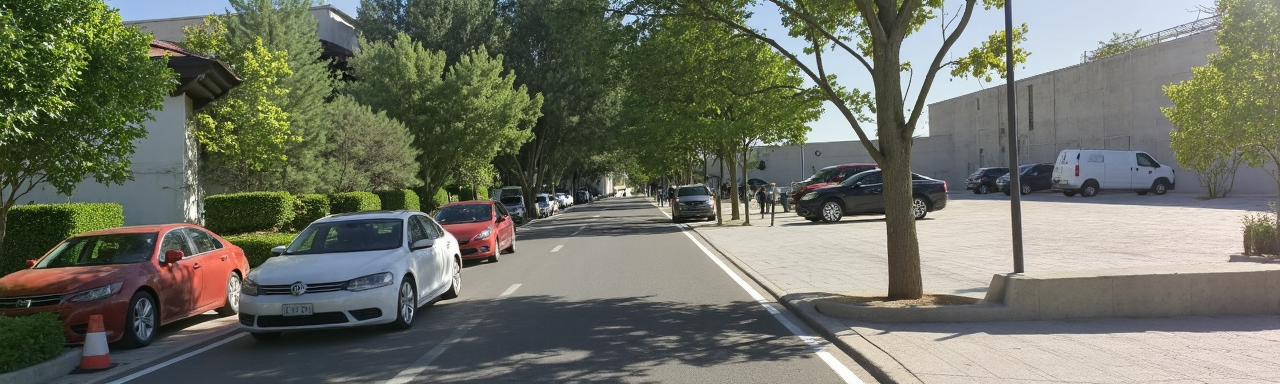}};
\supprowlabel{r4}{\textbf{$\psiPD$ (Ours)}}
\end{tikzpicture}
\caption{\textbf{vKITTI $\to$ KITTI qualitative comparison, Scene~3.}}
\label{fig:vkitti_s1}
\end{figure*}

\begin{figure*}[p]
\centering
\begin{tikzpicture}[piccell/.style={inner sep=0pt,outer sep=0pt,anchor=north west}]
\node[inner sep=0pt,outer sep=0pt] at (0,0) {};
\node[piccell] (r1) at (0,0) {\includegraphics[width=\suppimgw]{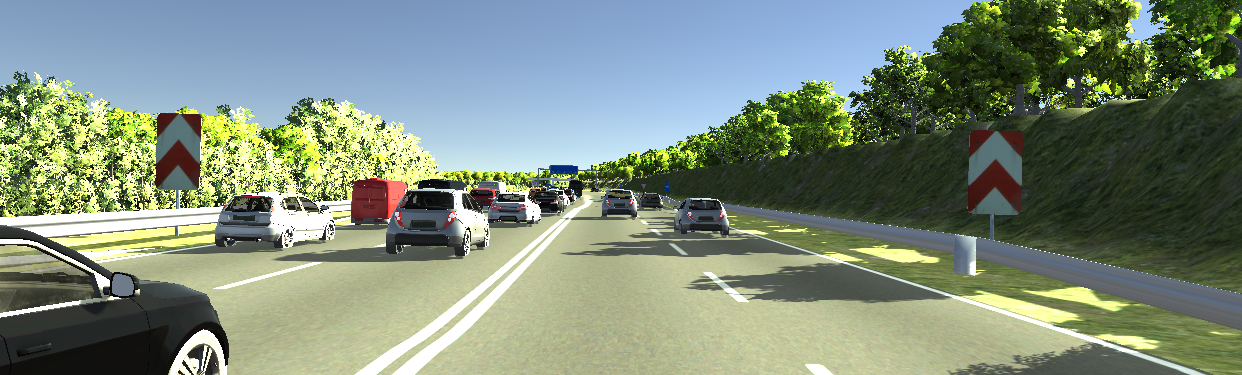}};
\supprowlabel{r1}{Input}
\node[piccell,below=0pt of r1] (r2) {\includegraphics[width=\suppimgw]{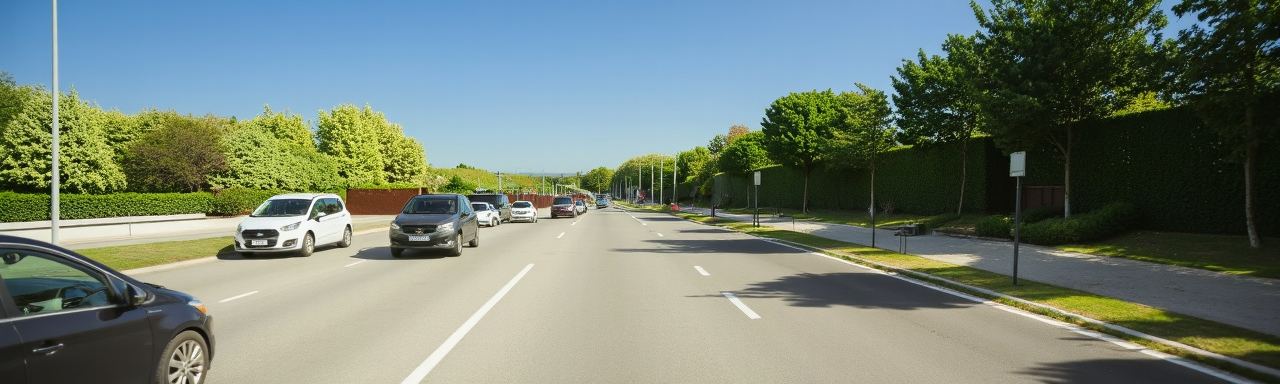}};
\supprowlabel{r2}{NeuralRemaster}
\node[piccell,below=0pt of r2] (r3) {\includegraphics[width=\suppimgw]{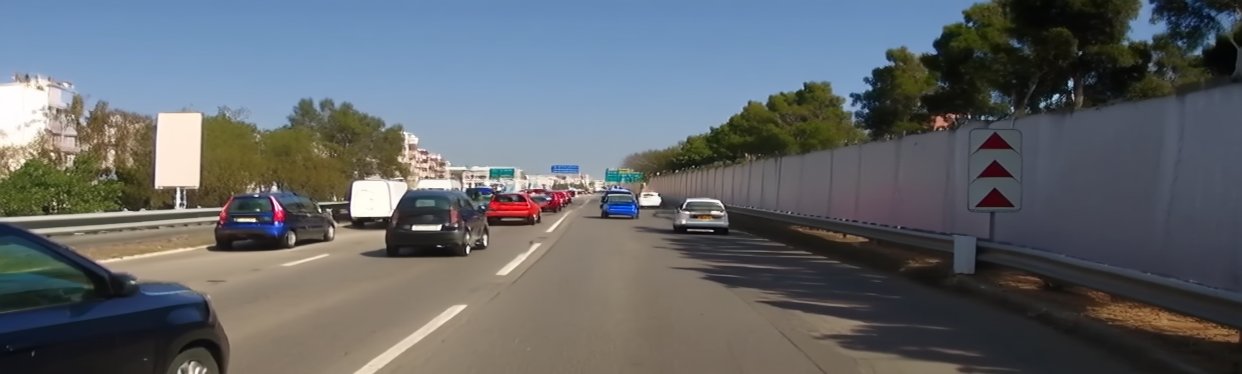}};
\supprowlabel{r3}{Cosmos Tr.~2.5}
\node[piccell,below=0pt of r3] (r4) {\includegraphics[width=\suppimgw]{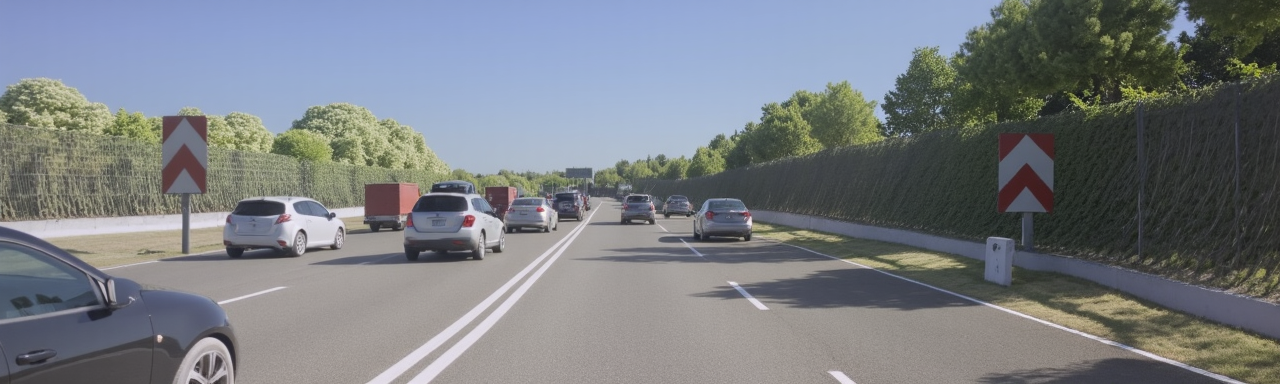}};
\supprowlabel{r4}{\textbf{$\psiPD$ (Ours)}}
\end{tikzpicture}
\caption{\textbf{vKITTI $\to$ KITTI qualitative comparison, Scene~4.}}
\label{fig:vkitti_s2}
\end{figure*}

\begin{figure*}[p]
\centering
\begin{tikzpicture}[piccell/.style={inner sep=0pt,outer sep=0pt,anchor=north west}]
\node[inner sep=0pt,outer sep=0pt] at (0,0) {};
\node[piccell] (r1) at (0,0) {\includegraphics[width=\suppimgw]{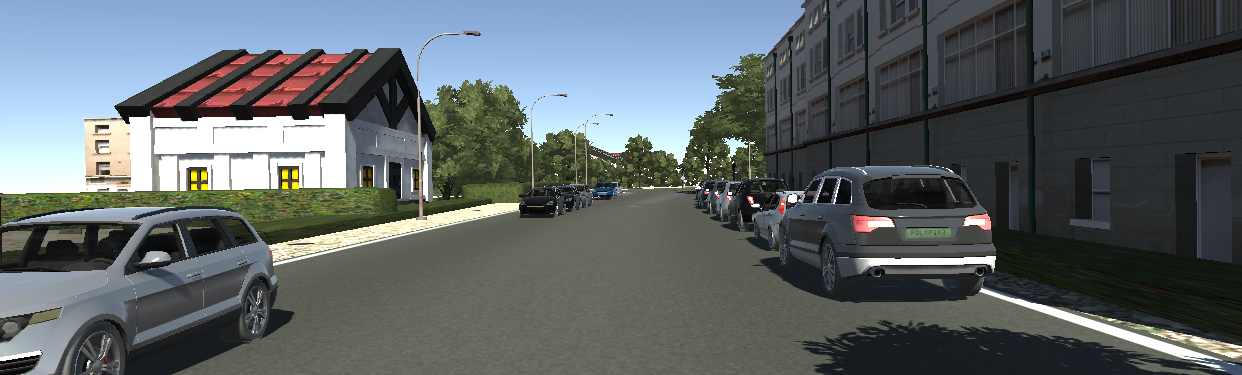}};
\supprowlabel{r1}{Input}
\node[piccell,below=0pt of r1] (r2) {\includegraphics[width=\suppimgw]{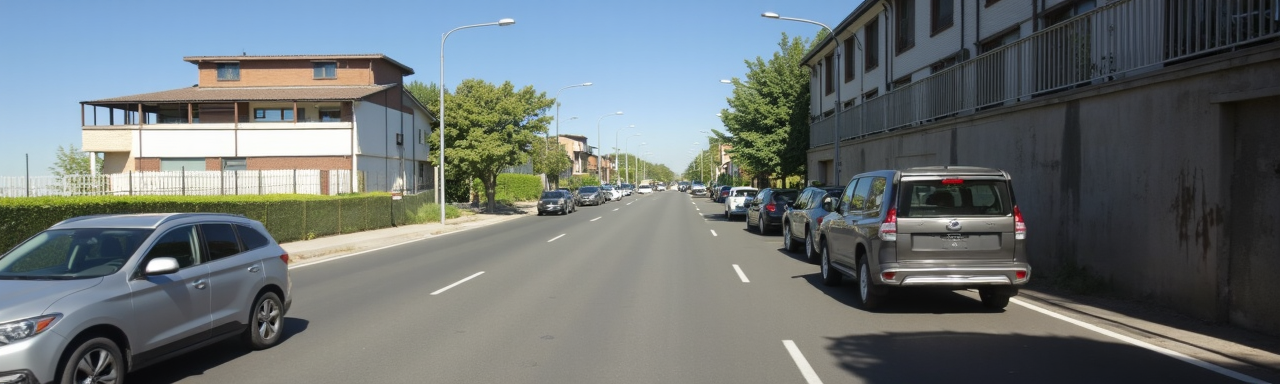}};
\supprowlabel{r2}{NeuralRemaster}
\node[piccell,below=0pt of r2] (r3) {\includegraphics[width=\suppimgw]{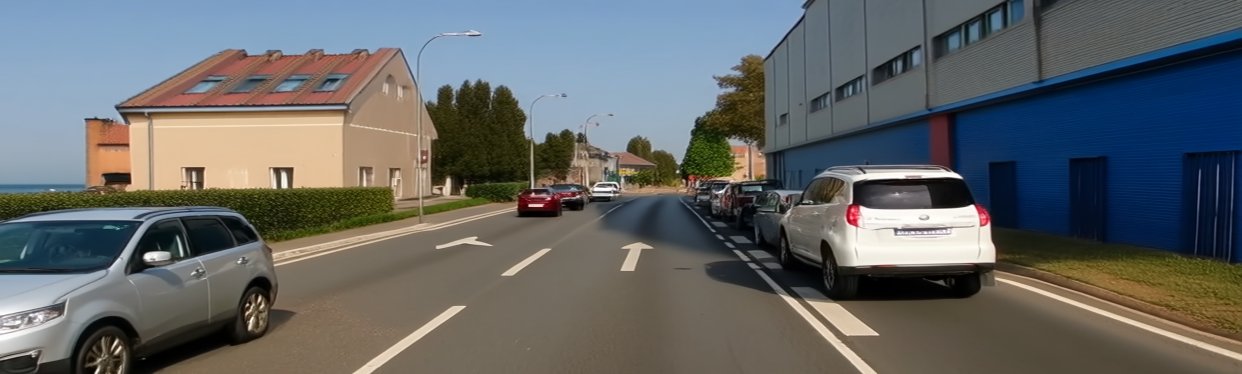}};
\supprowlabel{r3}{Cosmos Tr.~2.5}
\node[piccell,below=0pt of r3] (r4) {\includegraphics[width=\suppimgw]{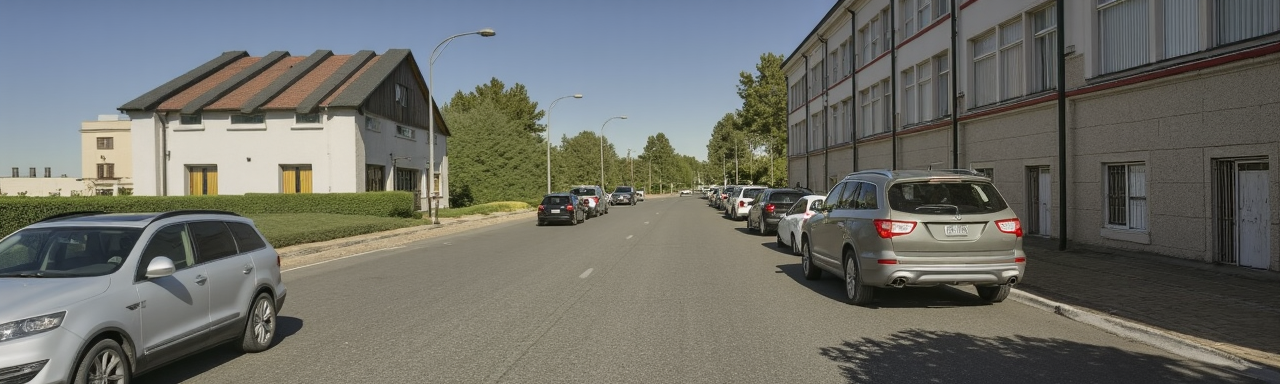}};
\supprowlabel{r4}{\textbf{$\psiPD$ (Ours)}}
\end{tikzpicture}
\caption{\textbf{vKITTI $\to$ KITTI qualitative comparison, Scene~5.}}
\label{fig:vkitti_s3}
\end{figure*}

\clearpage
\begin{figure*}[t]
    \centering
    \begin{tikzpicture}[
      piccell/.style={inner sep=0pt, outer sep=0pt, anchor=north west},
    ]
    \def\imgw{0.182\linewidth}
    \def\labelgap{0.5em}
    \def\labw{2.0cm}
    \def\framegap{0pt}

    \node[piccell] (r0f1) at (0,0) {\includegraphics[width=\imgw]{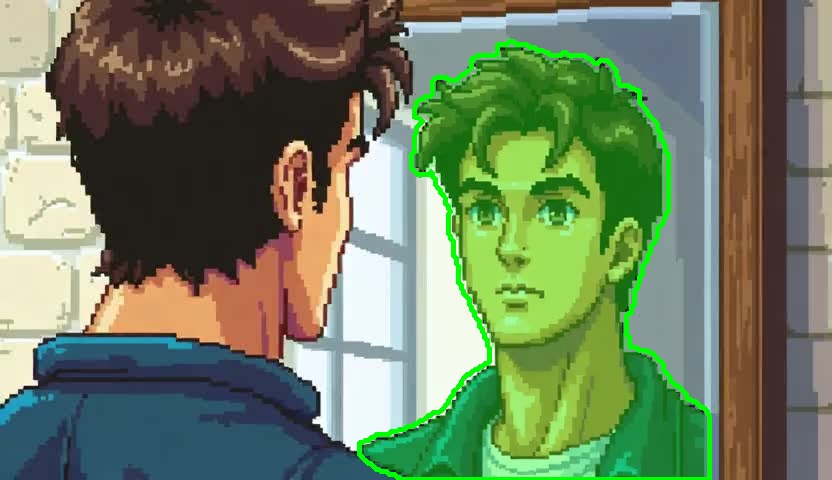}};
    \node[piccell, right=\framegap of r0f1] (r0f2) {\includegraphics[width=\imgw]{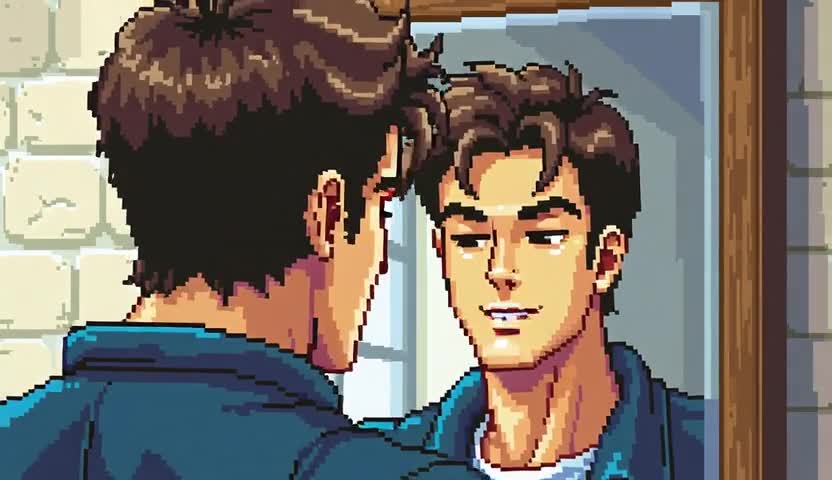}};
    \node[piccell, right=\framegap of r0f2] (r0f3) {\includegraphics[width=\imgw]{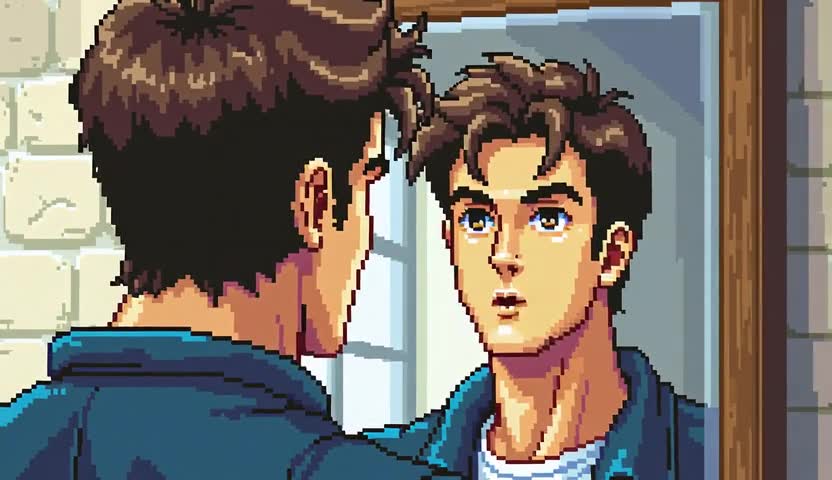}};
    \node[piccell, right=\framegap of r0f3] (r0f4) {\includegraphics[width=\imgw]{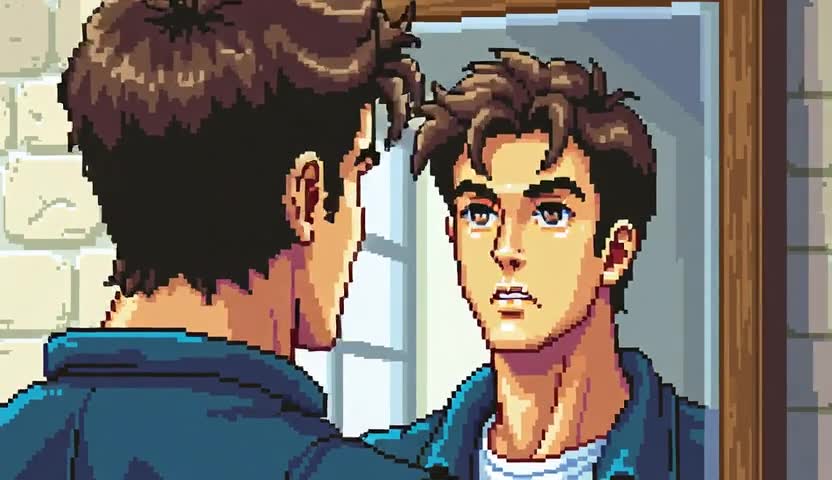}};
    \node[piccell, right=\framegap of r0f4] (r0f5) {\includegraphics[width=\imgw]{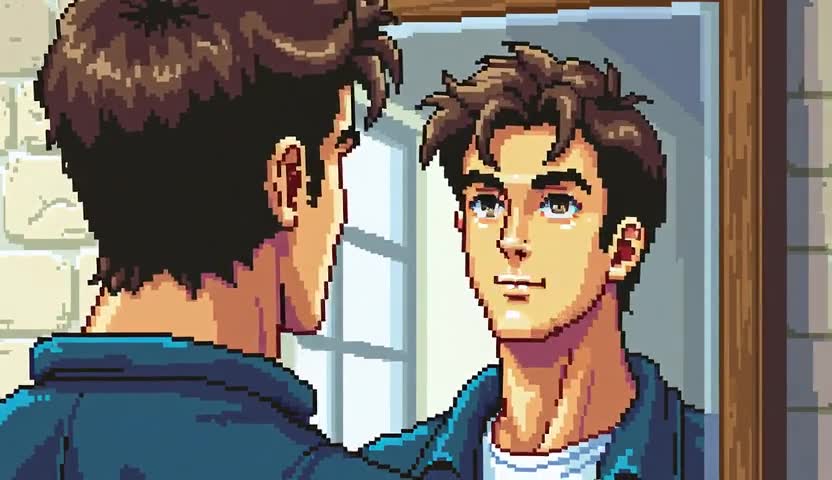}};
    \node[anchor=east, inner sep=0pt, outer sep=0pt]
      at ($(r0f1.west)+(-\labelgap,0)$)
      {\rotatebox{90}{\makebox[\labw][c]{\fontsize{6}{6}\selectfont Input}}};

    \node[piccell, below=0pt of r0f1] (r1f1) {\includegraphics[width=\imgw]{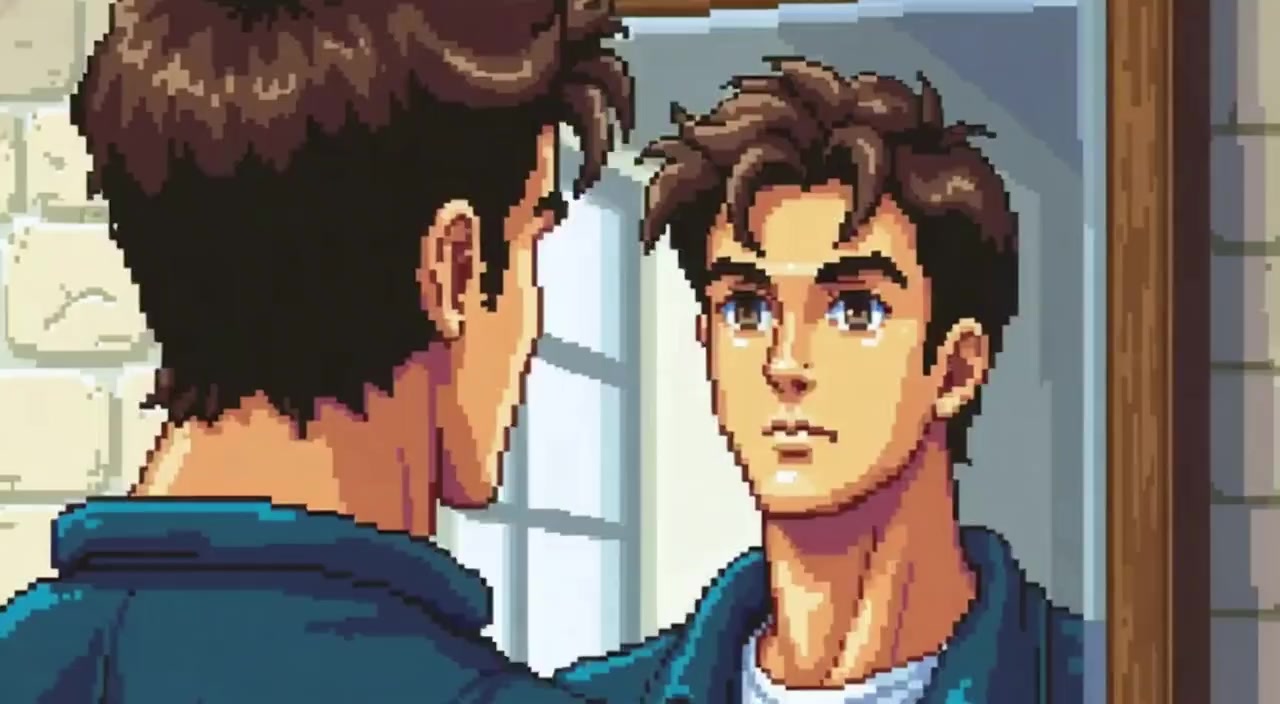}};
    \node[piccell, right=\framegap of r1f1] (r1f2) {\includegraphics[width=\imgw]{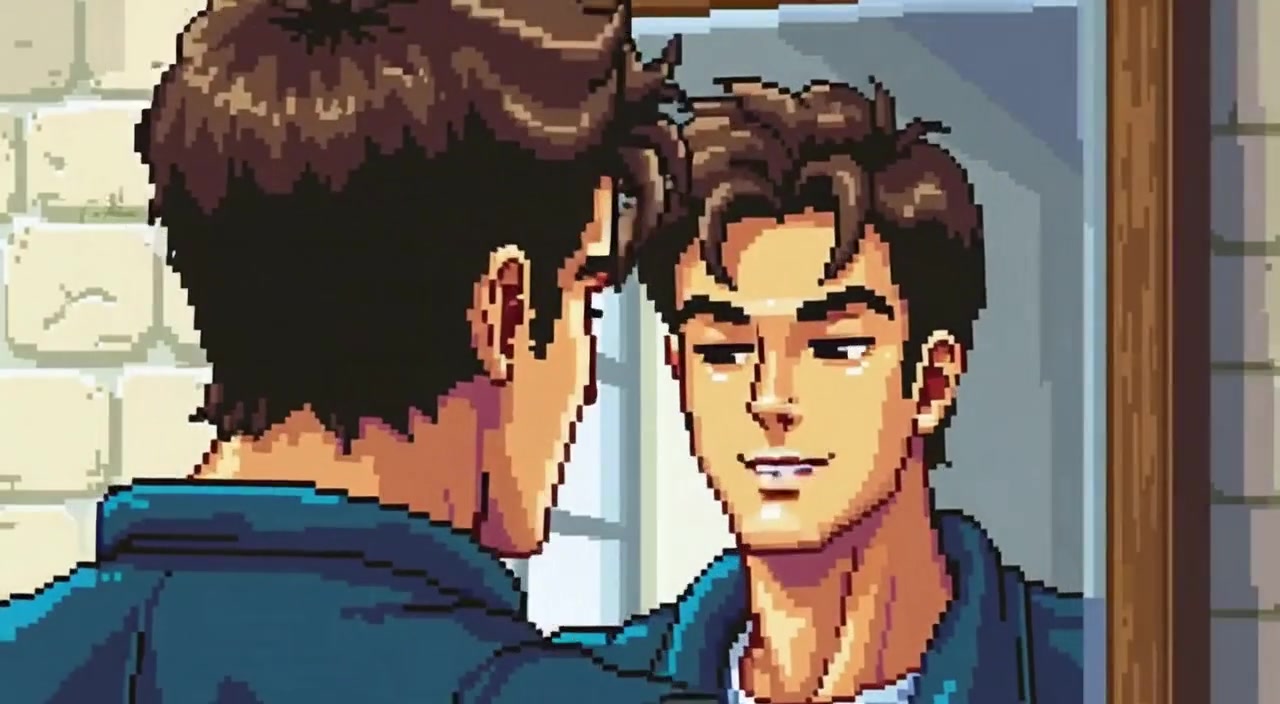}};
    \node[piccell, right=\framegap of r1f2] (r1f3) {\includegraphics[width=\imgw]{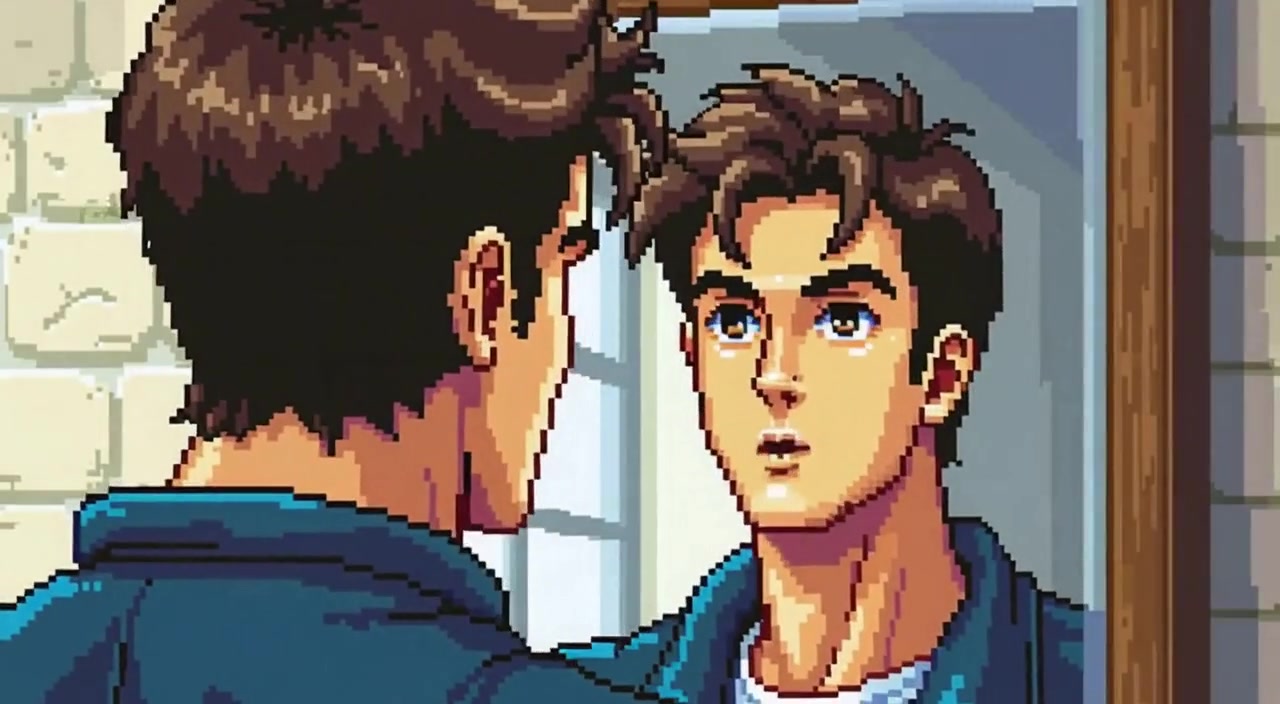}};
    \node[piccell, right=\framegap of r1f3] (r1f4) {\includegraphics[width=\imgw]{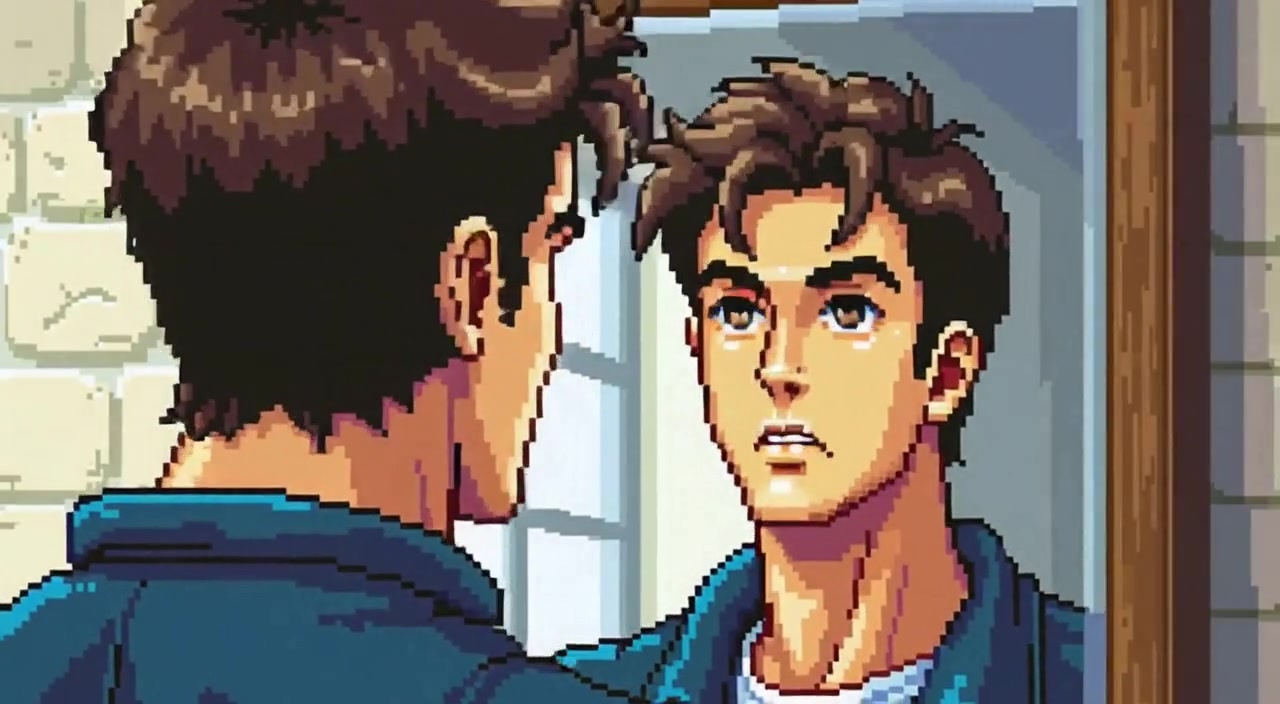}};
    \node[piccell, right=\framegap of r1f4] (r1f5) {\includegraphics[width=\imgw]{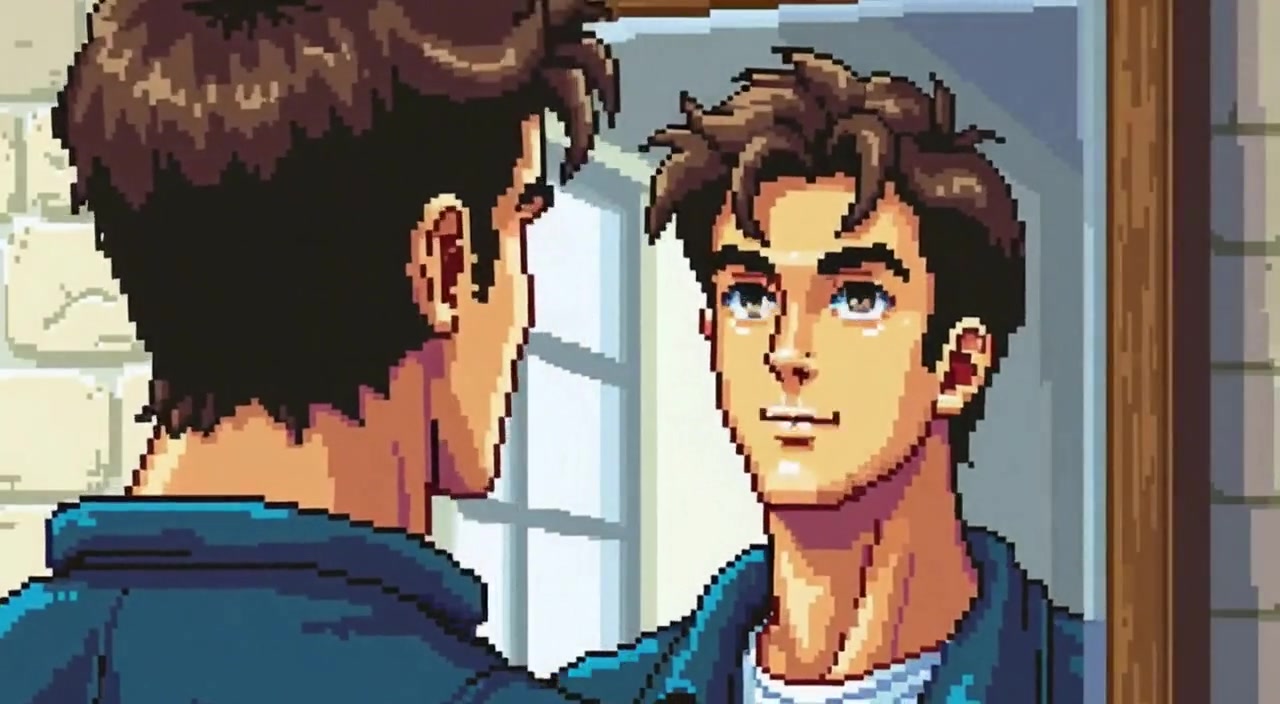}};
    \node[anchor=east, inner sep=0pt, outer sep=0pt]
      at ($(r1f1.west)+(-\labelgap,0)$)
      {\rotatebox{90}{\makebox[\labw][c]{\fontsize{6}{6}\selectfont Ditto (local)}}};

    \node[piccell, below=0pt of r1f1] (r2f1) {\includegraphics[width=\imgw]{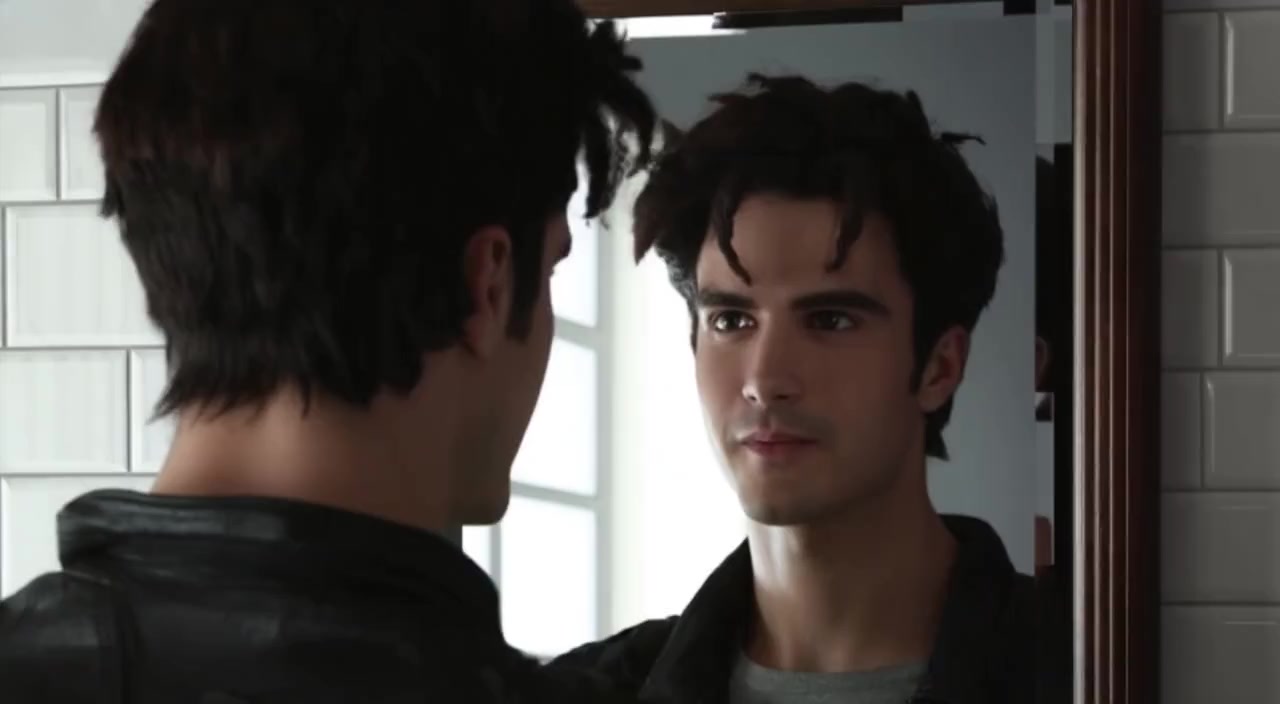}};
    \node[piccell, right=\framegap of r2f1] (r2f2) {\includegraphics[width=\imgw]{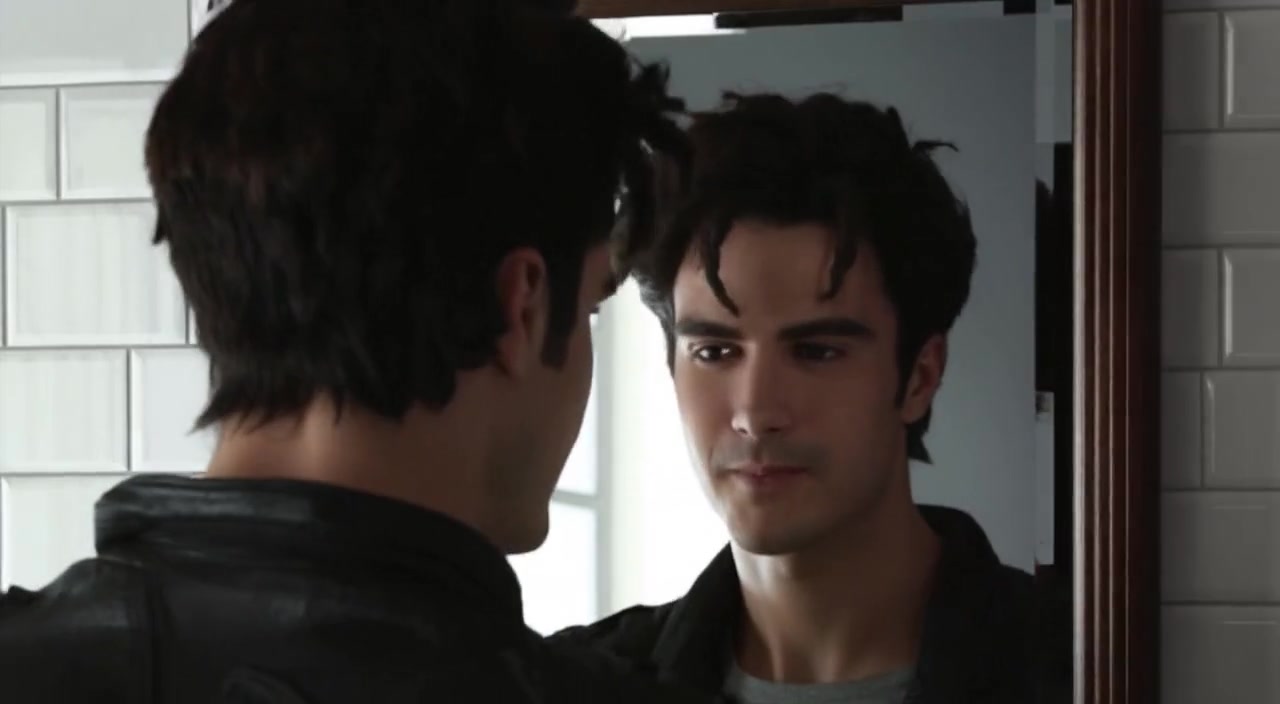}};
    \node[piccell, right=\framegap of r2f2] (r2f3) {\includegraphics[width=\imgw]{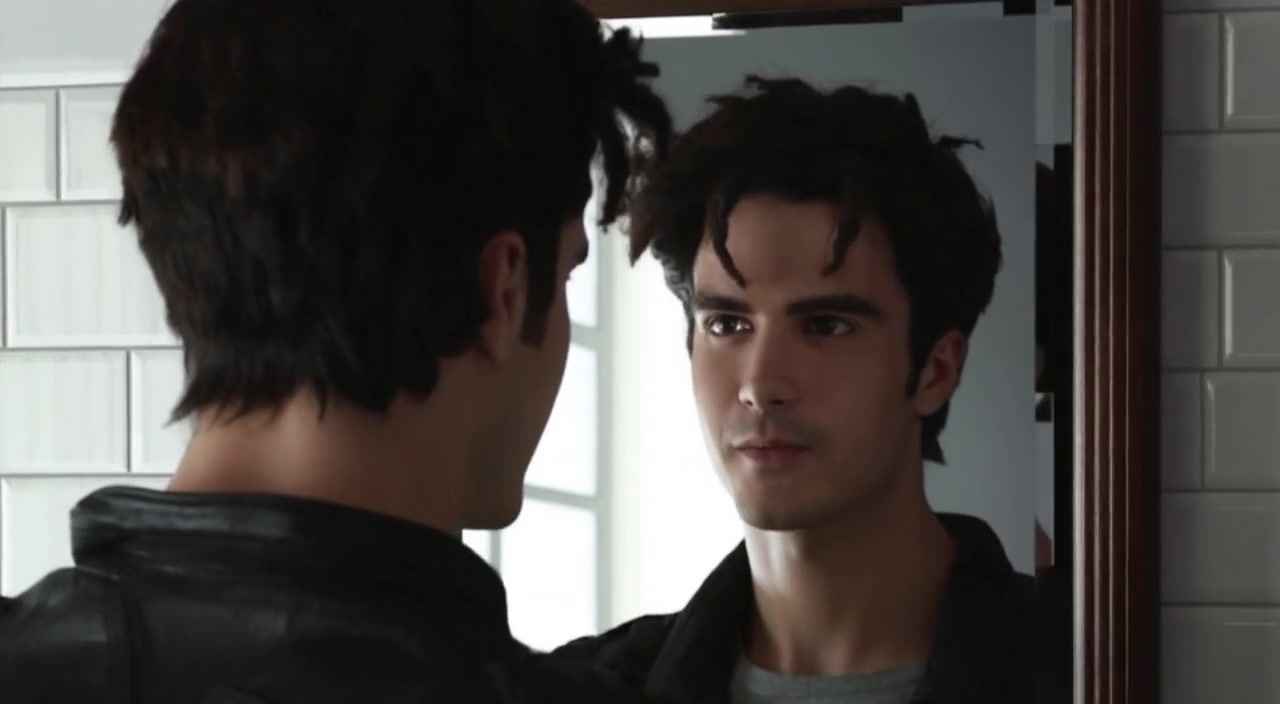}};
    \node[piccell, right=\framegap of r2f3] (r2f4) {\includegraphics[width=\imgw]{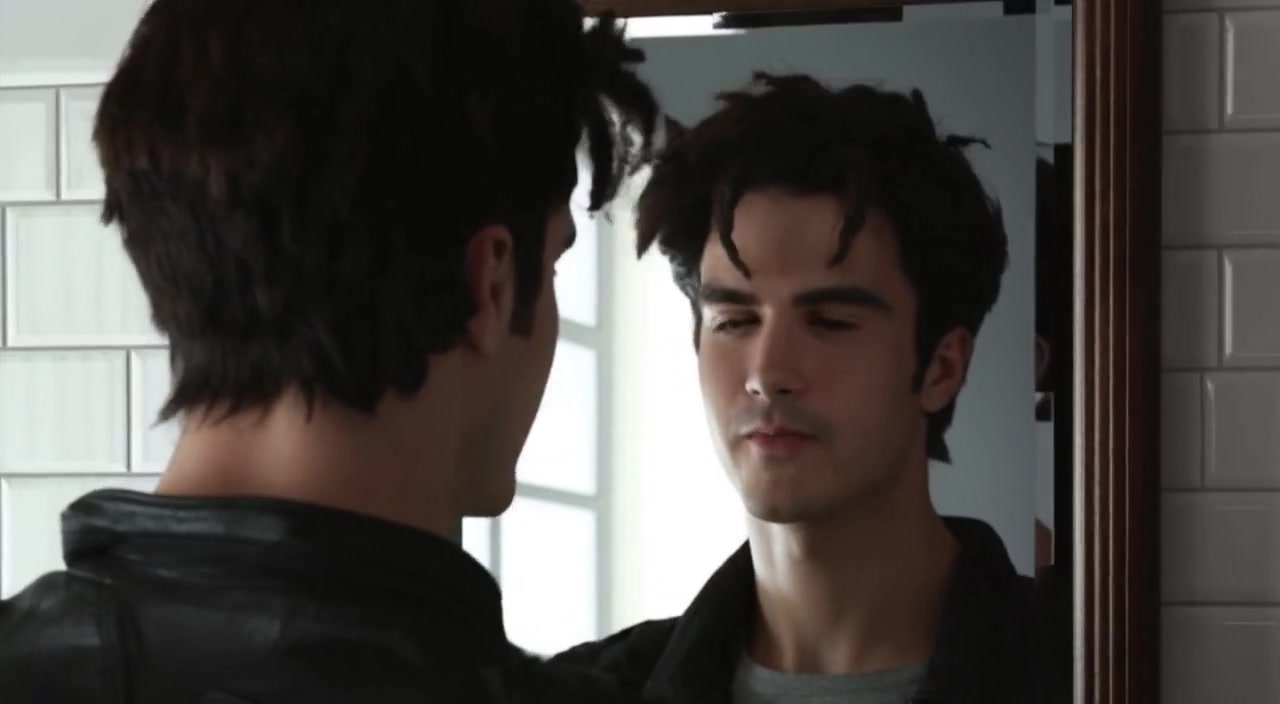}};
    \node[piccell, right=\framegap of r2f4] (r2f5) {\includegraphics[width=\imgw]{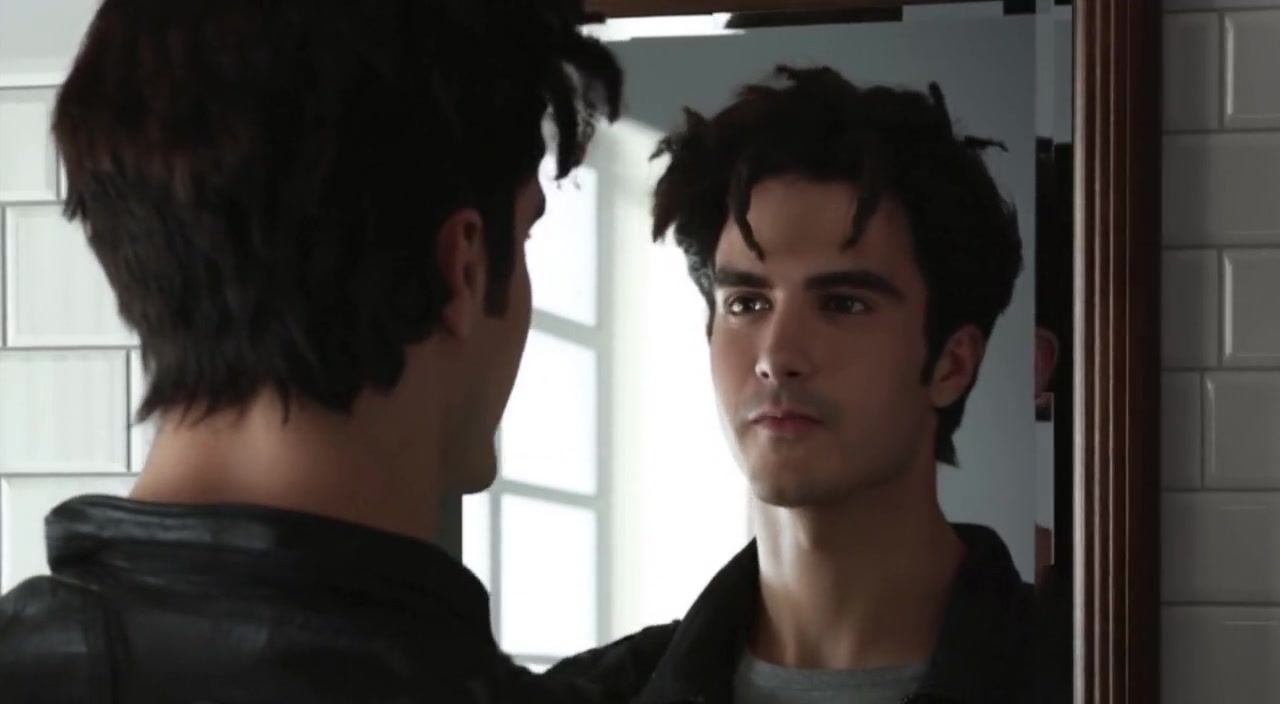}};
    \node[anchor=east, inner sep=0pt, outer sep=0pt]
      at ($(r2f1.west)+(-\labelgap,0)$)
      {\rotatebox{90}{\makebox[\labw][c]{\fontsize{6}{6}\selectfont Ditto (sim2real)}}};

    \node[piccell, below=0pt of r2f1] (r3f1) {\includegraphics[width=\imgw]{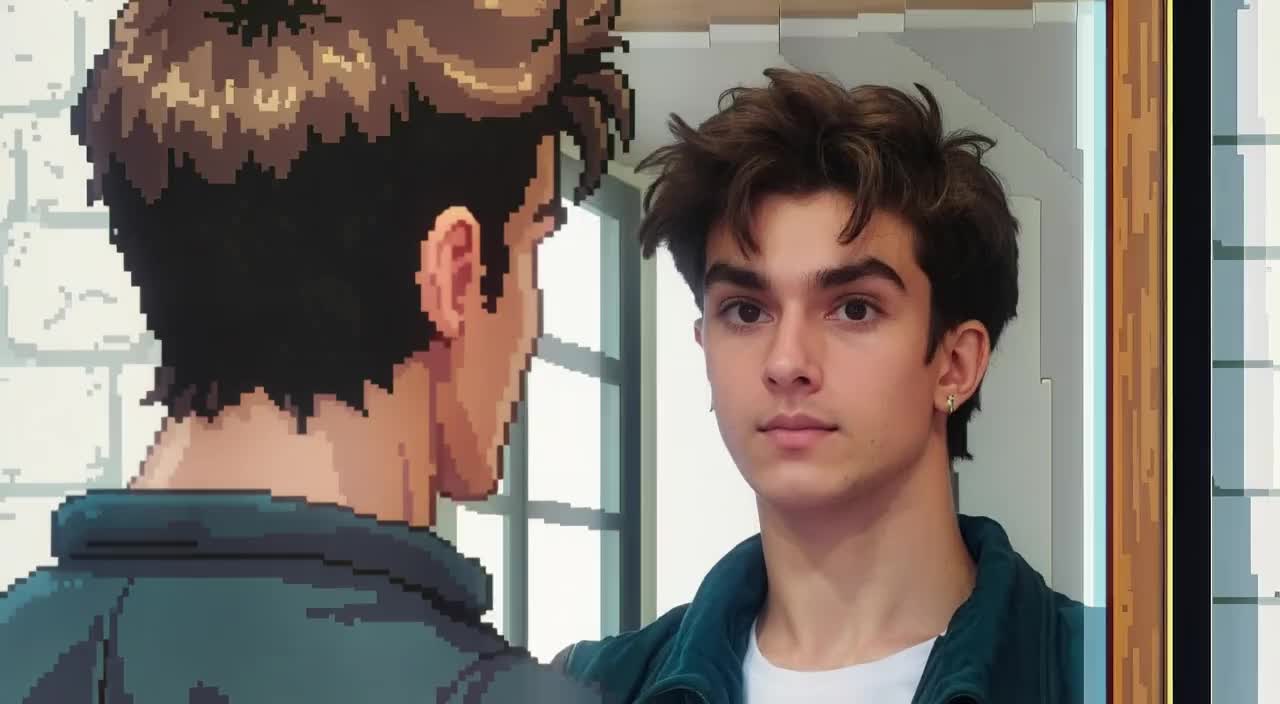}};
    \node[piccell, right=\framegap of r3f1] (r3f2) {\includegraphics[width=\imgw]{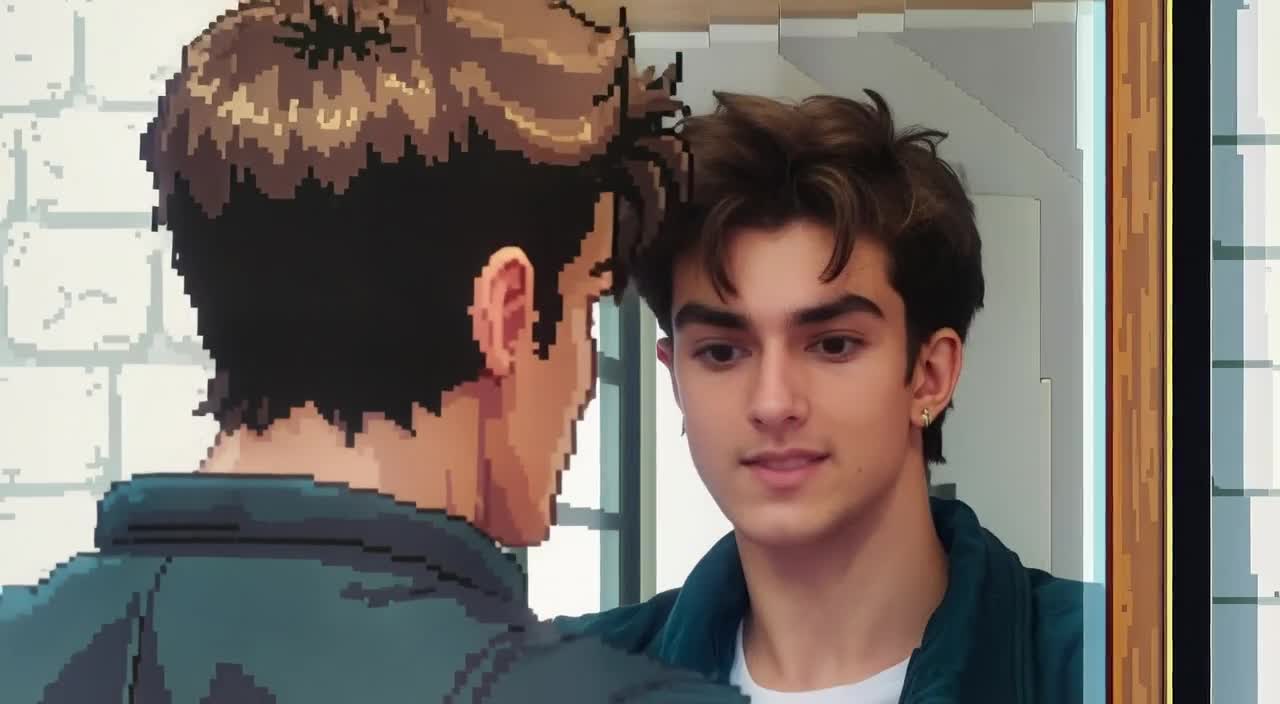}};
    \node[piccell, right=\framegap of r3f2] (r3f3) {\includegraphics[width=\imgw]{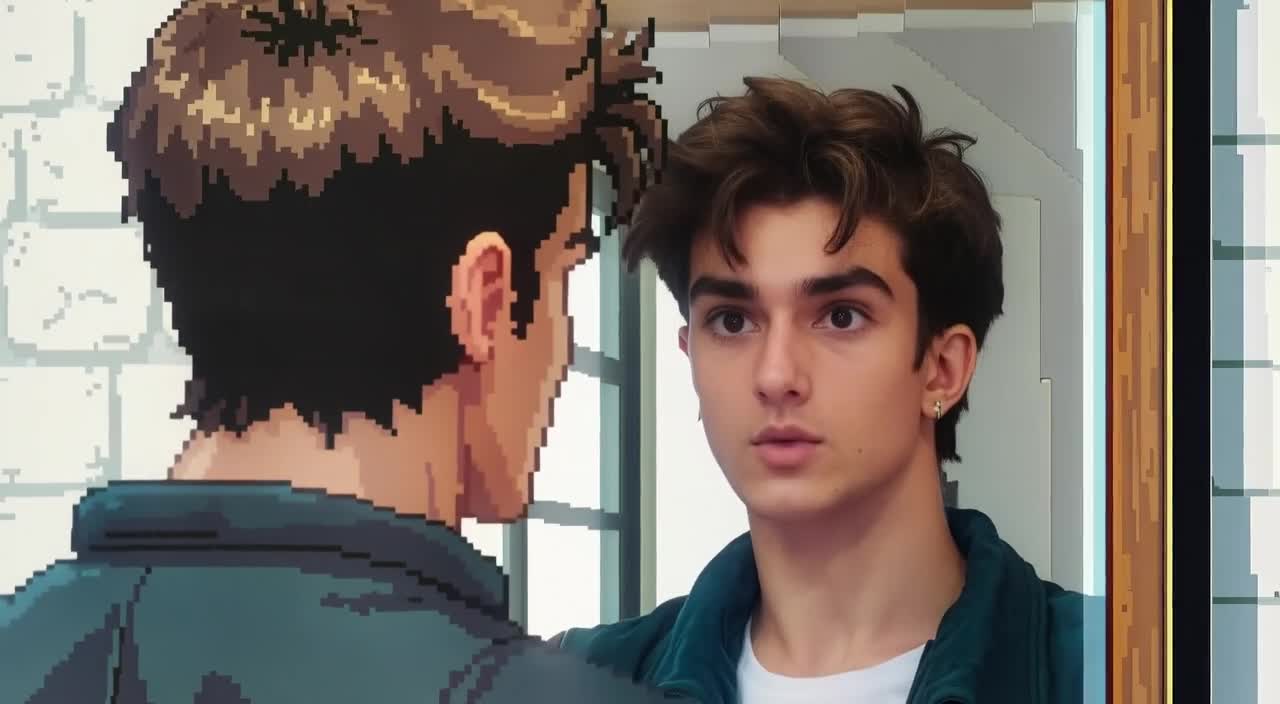}};
    \node[piccell, right=\framegap of r3f3] (r3f4) {\includegraphics[width=\imgw]{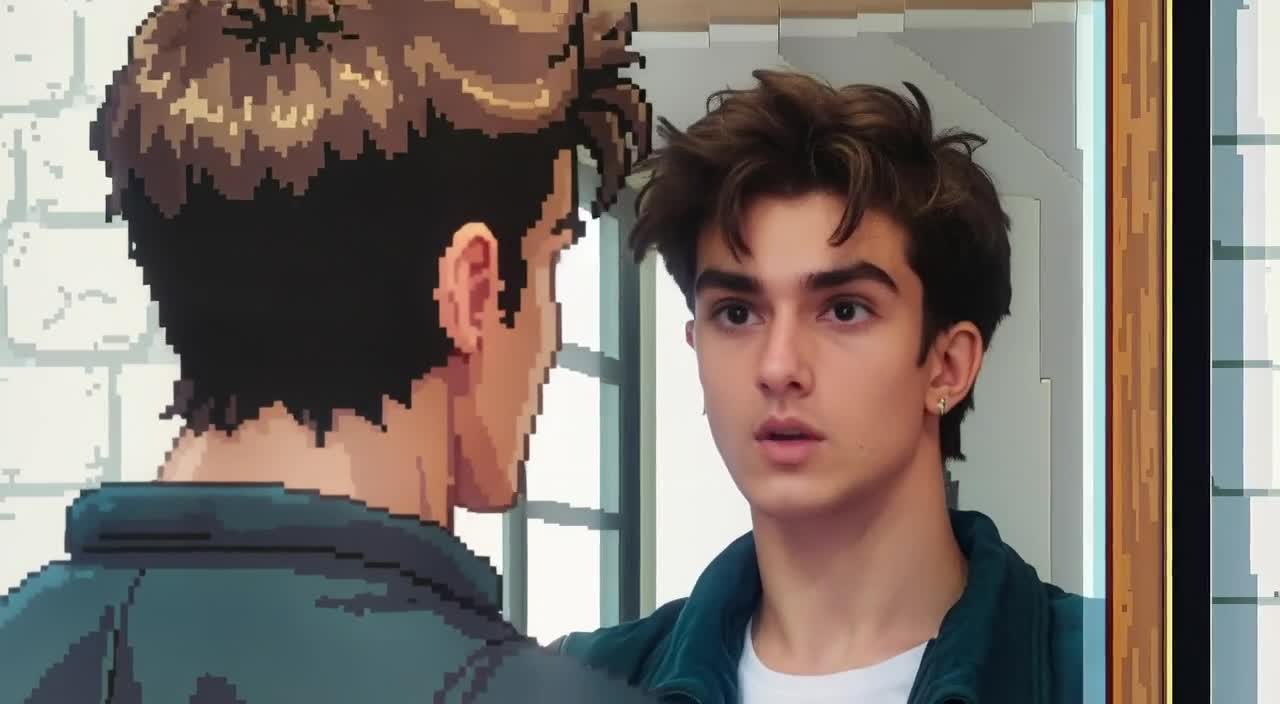}};
    \node[piccell, right=\framegap of r3f4] (r3f5) {\includegraphics[width=\imgw]{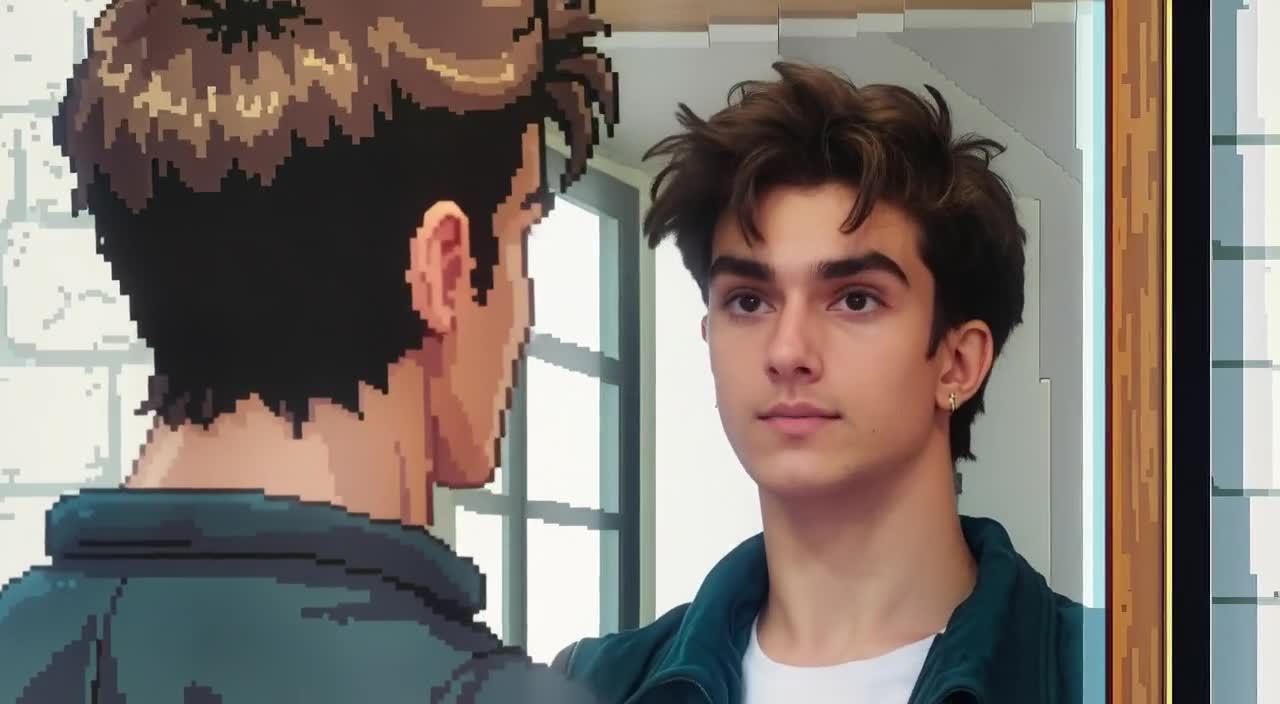}};
    \node[anchor=east, inner sep=0pt, outer sep=0pt]
      at ($(r3f1.west)+(-\labelgap,0)$)
      {\rotatebox{90}{\makebox[\labw][c]{\fontsize{6}{6}\selectfont $\psiPD$ (Ours)}}};

    \node[piccell, below=0pt of r3f1] (r3bf1) {\includegraphics[width=\imgw]{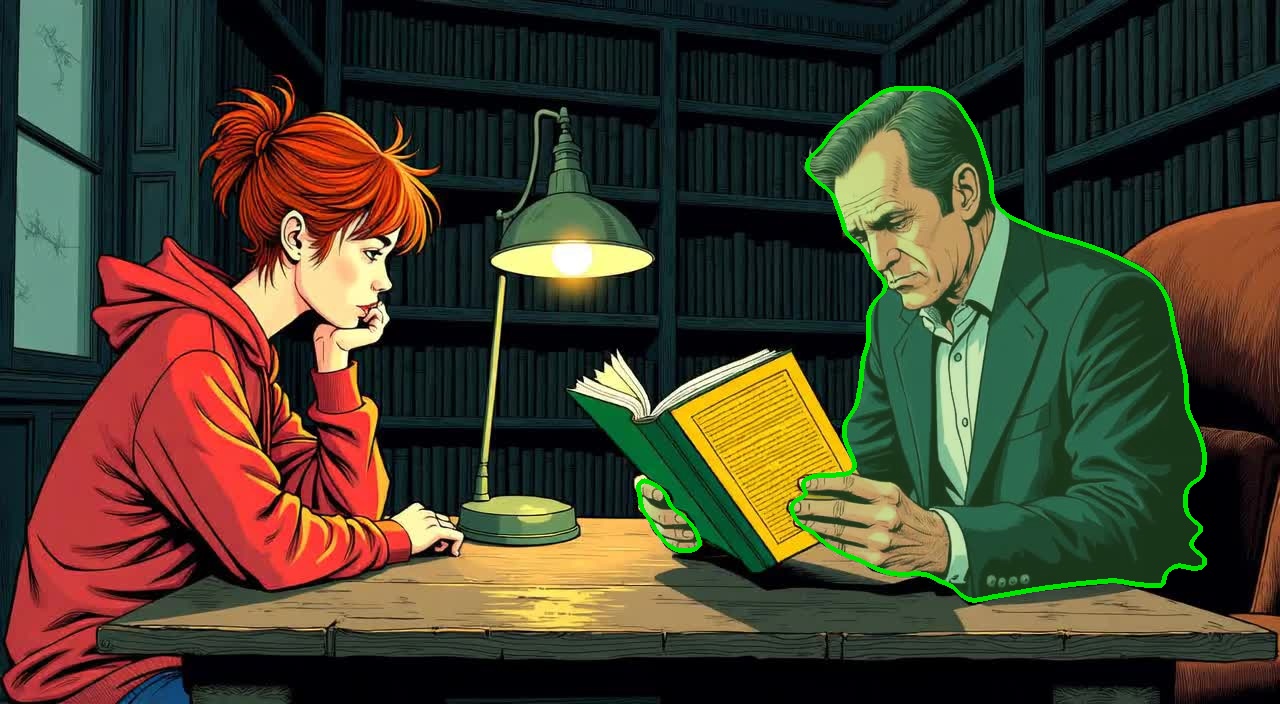}};
    \node[piccell, right=\framegap of r3bf1] (r3bf2) {\includegraphics[width=\imgw]{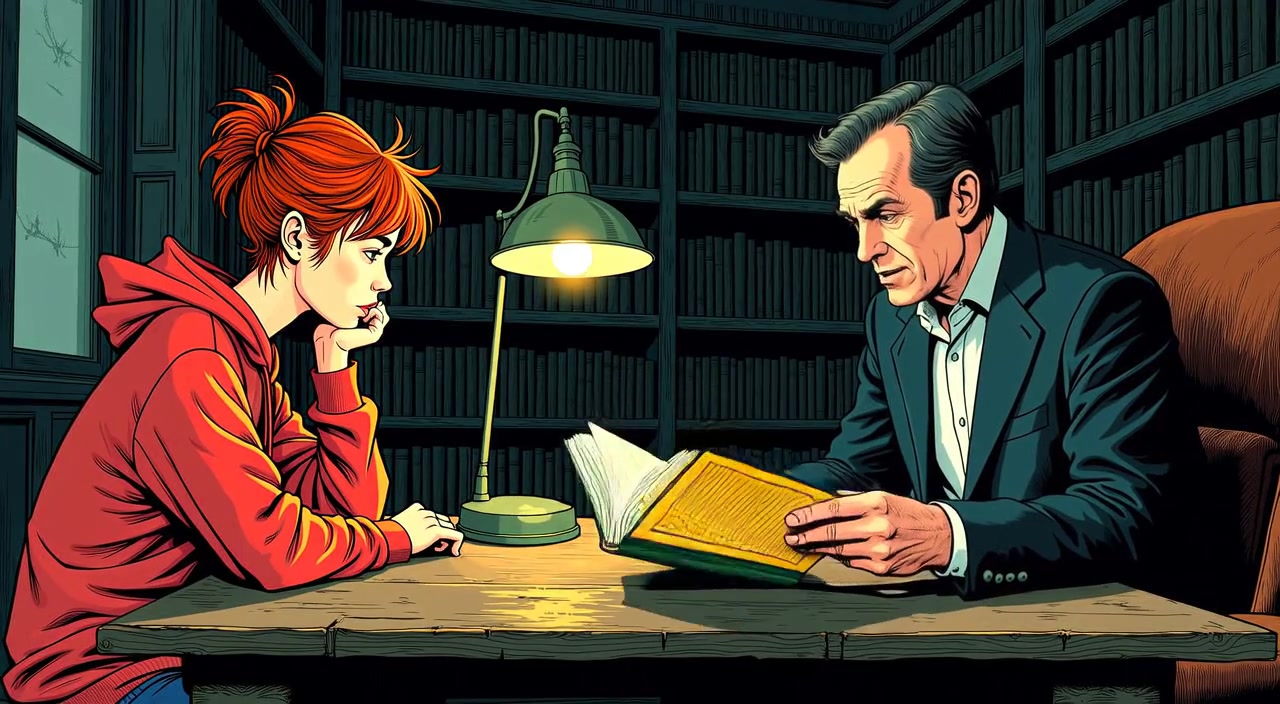}};
    \node[piccell, right=\framegap of r3bf2] (r3bf3) {\includegraphics[width=\imgw]{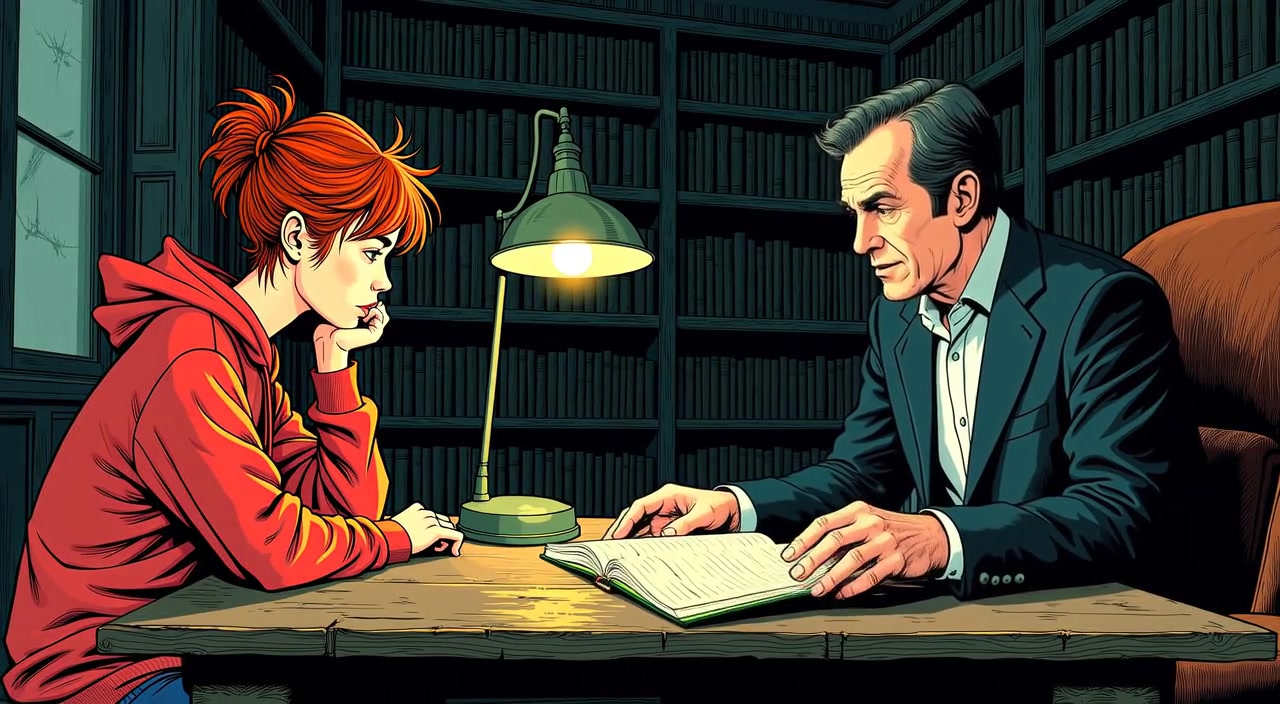}};
    \node[piccell, right=\framegap of r3bf3] (r3bf4) {\includegraphics[width=\imgw]{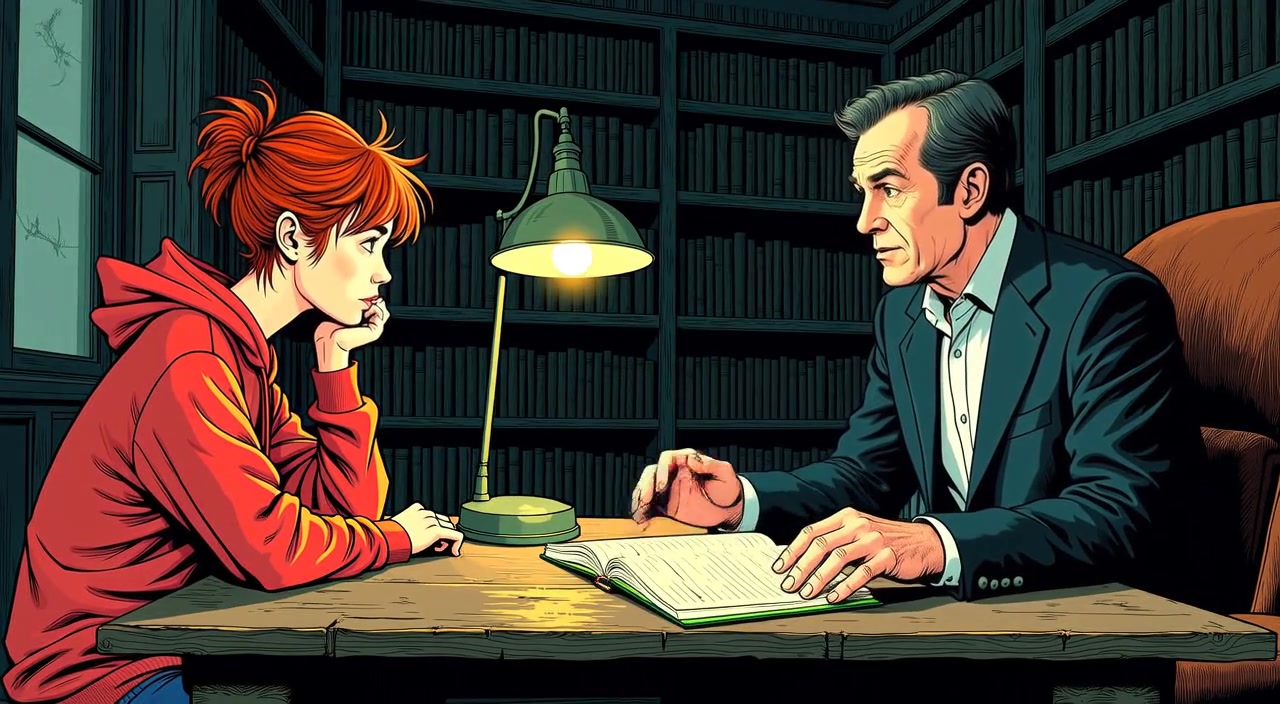}};
    \node[piccell, right=\framegap of r3bf4] (r3bf5) {\includegraphics[width=\imgw]{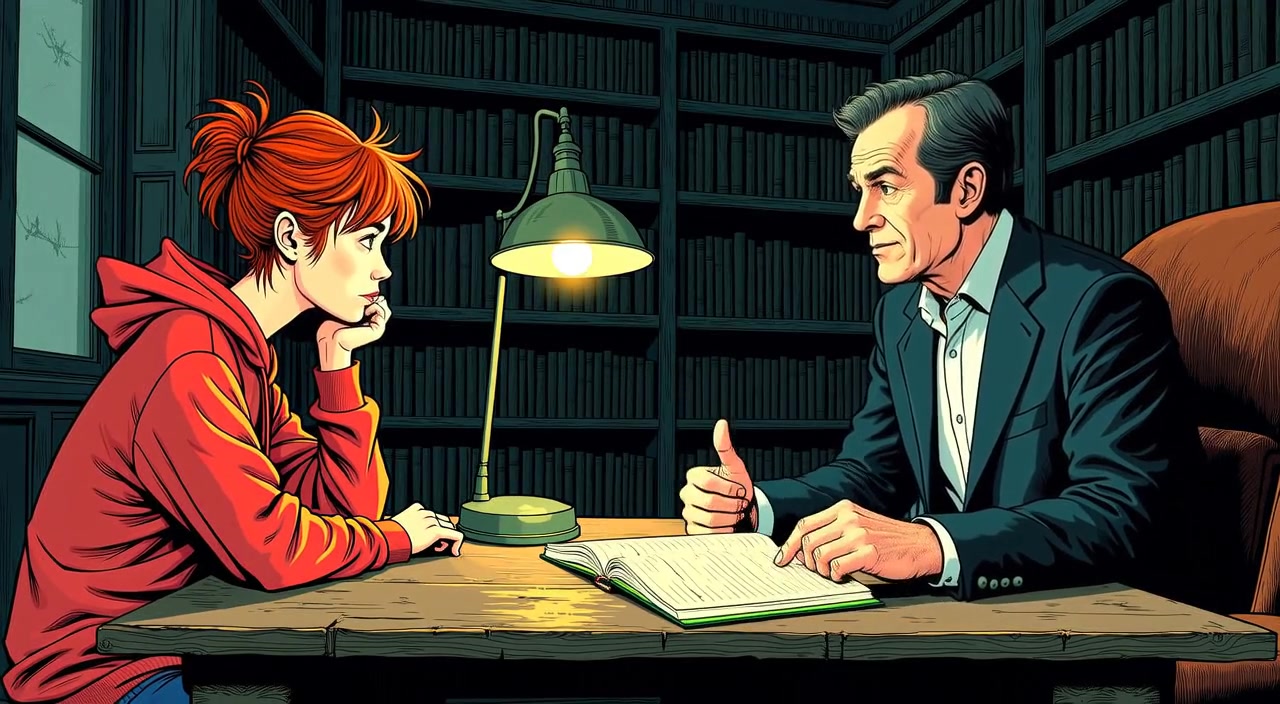}};
    \node[anchor=east, inner sep=0pt, outer sep=0pt]
      at ($(r3bf1.west)+(-\labelgap,0)$)
      {\rotatebox{90}{\makebox[\labw][c]{\fontsize{6}{6}\selectfont Input}}};

    \node[piccell, below=0pt of r3bf1] (r4f1) {\includegraphics[width=\imgw]{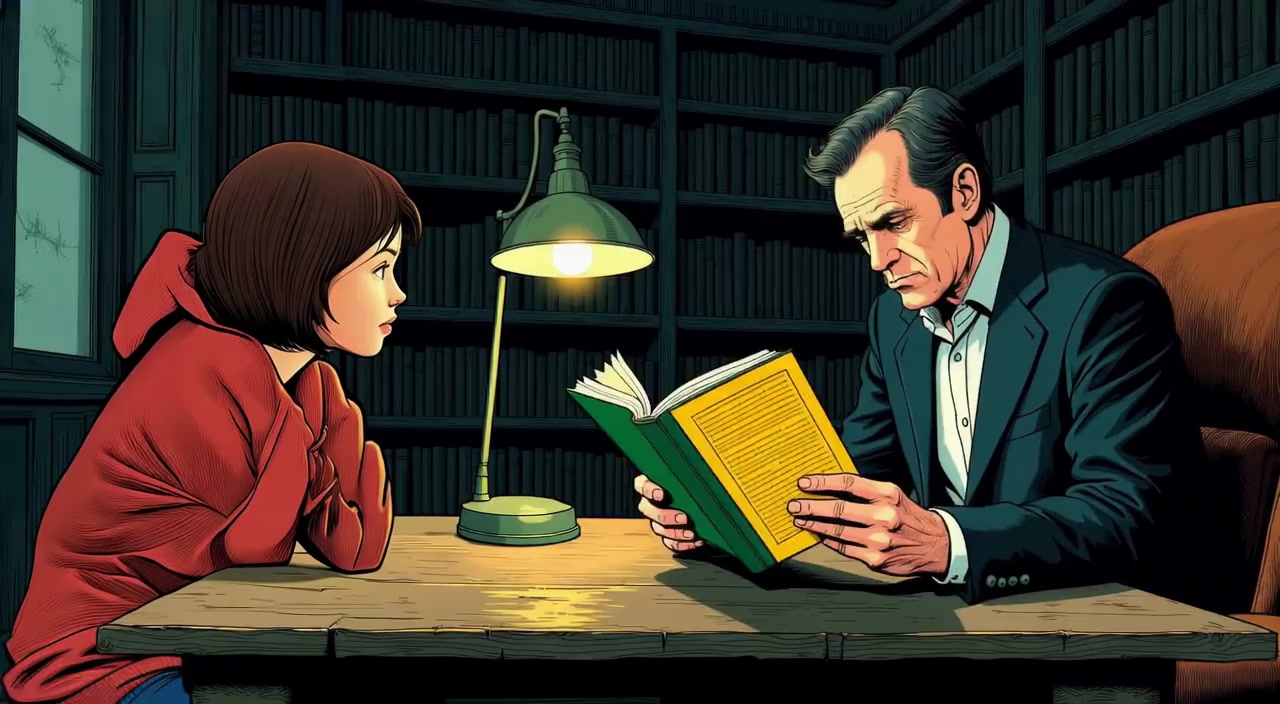}};
    \node[piccell, right=\framegap of r4f1] (r4f2) {\includegraphics[width=\imgw]{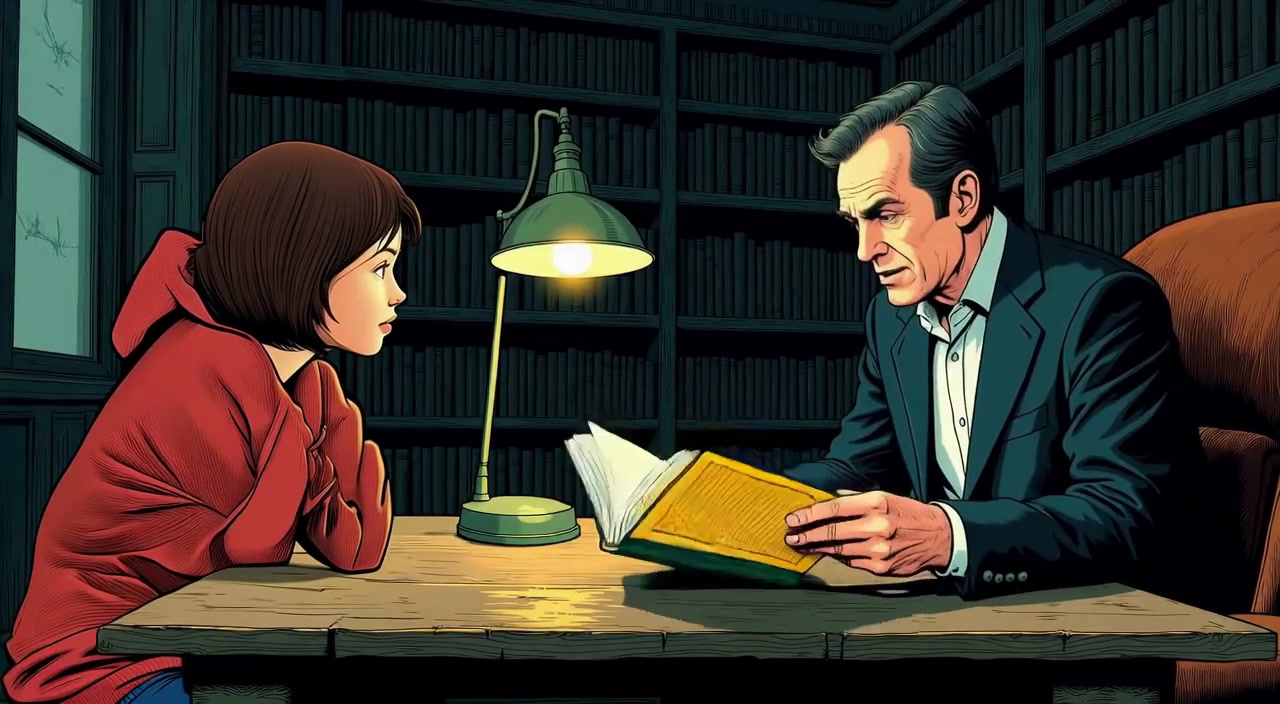}};
    \node[piccell, right=\framegap of r4f2] (r4f3) {\includegraphics[width=\imgw]{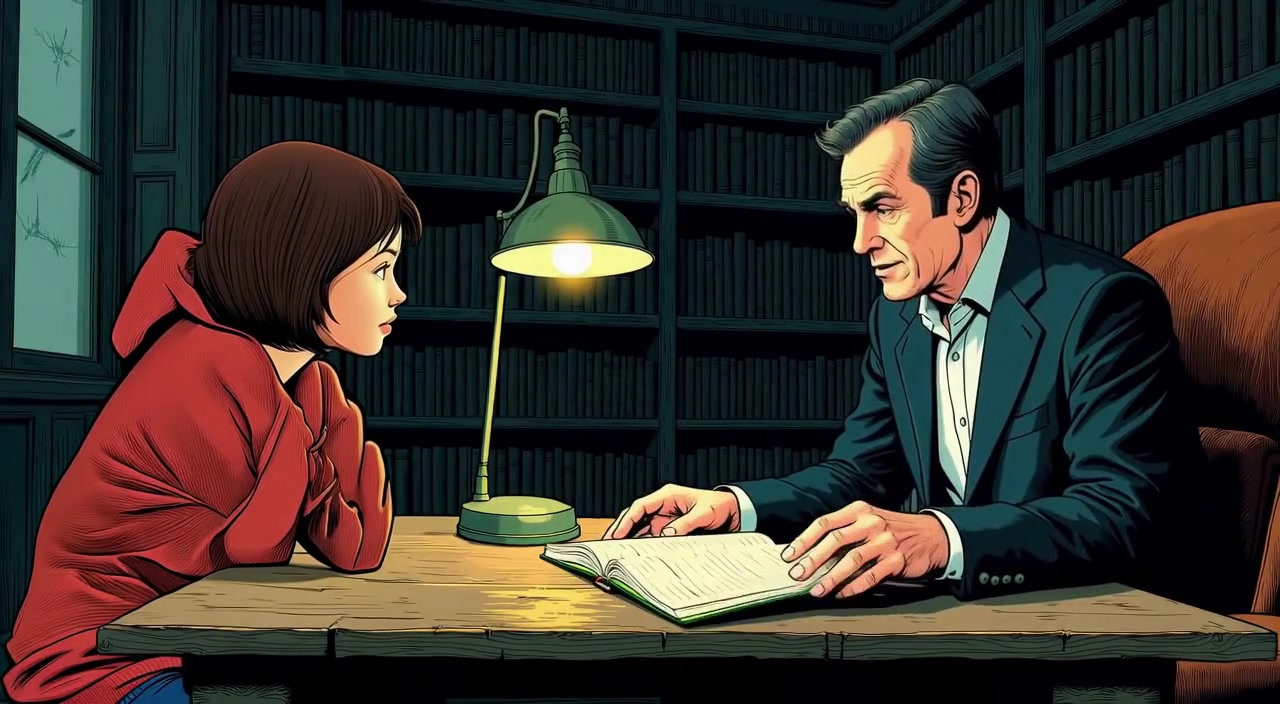}};
    \node[piccell, right=\framegap of r4f3] (r4f4) {\includegraphics[width=\imgw]{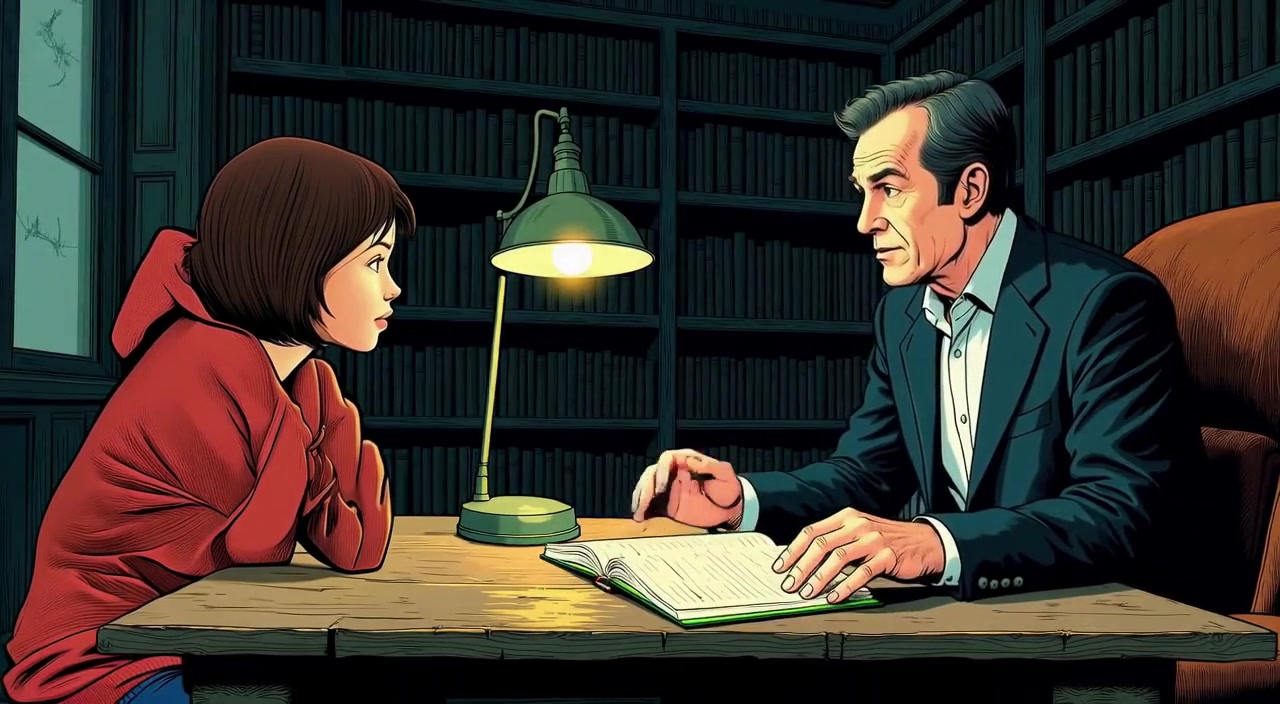}};
    \node[piccell, right=\framegap of r4f4] (r4f5) {\includegraphics[width=\imgw]{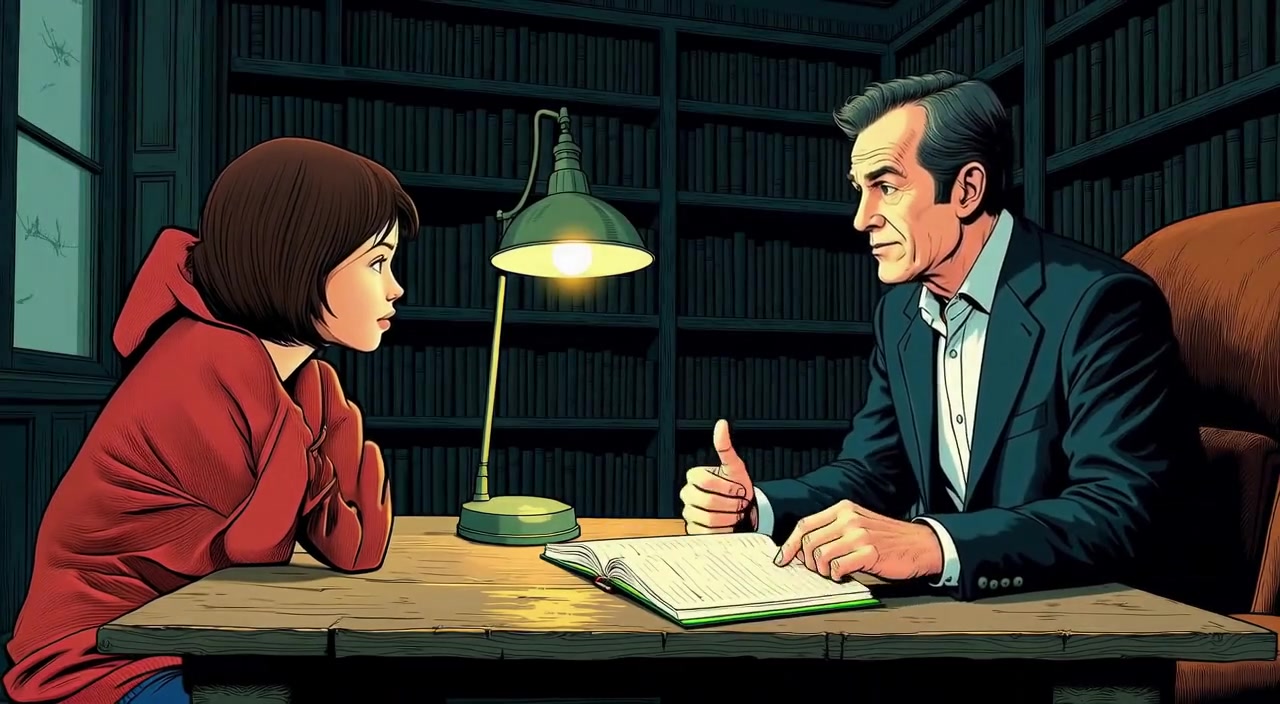}};
    \node[anchor=east, inner sep=0pt, outer sep=0pt]
      at ($(r4f1.west)+(-\labelgap,0)$)
      {\rotatebox{90}{\makebox[\labw][c]{\fontsize{6}{6}\selectfont Ditto (local)}}};

    \node[piccell, below=0pt of r4f1] (r5f1) {\includegraphics[width=\imgw]{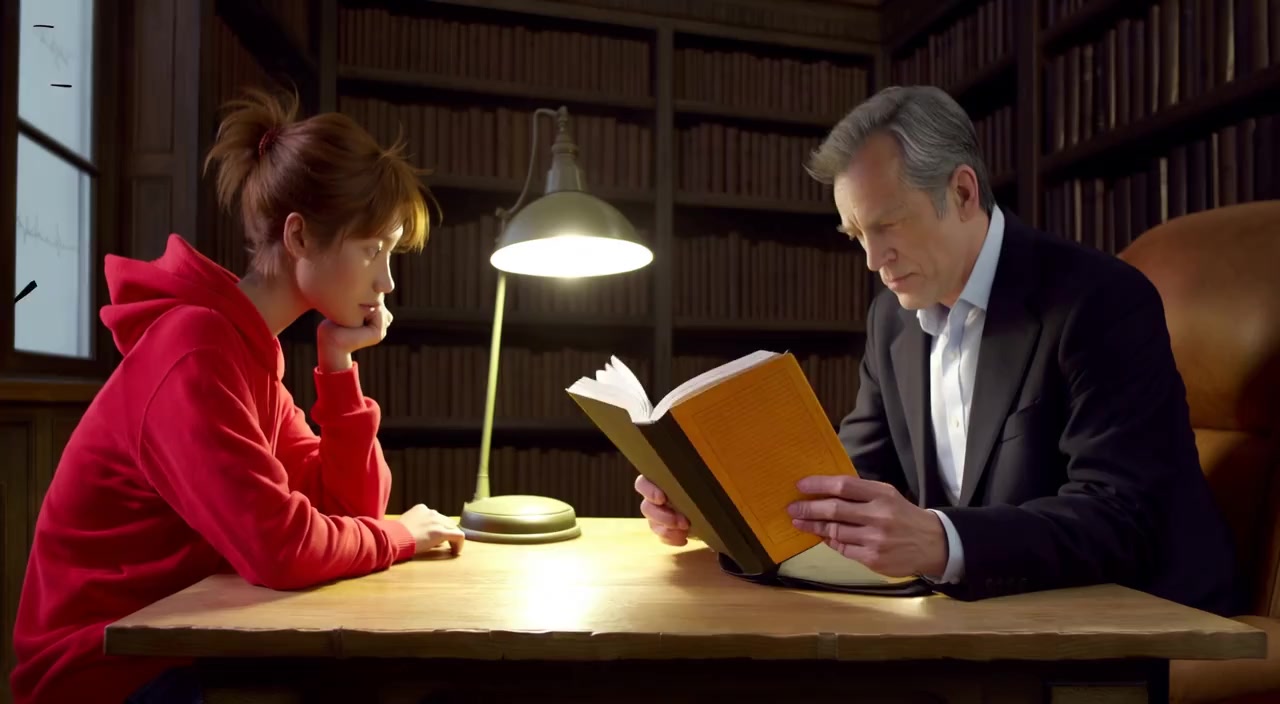}};
    \node[piccell, right=\framegap of r5f1] (r5f2) {\includegraphics[width=\imgw]{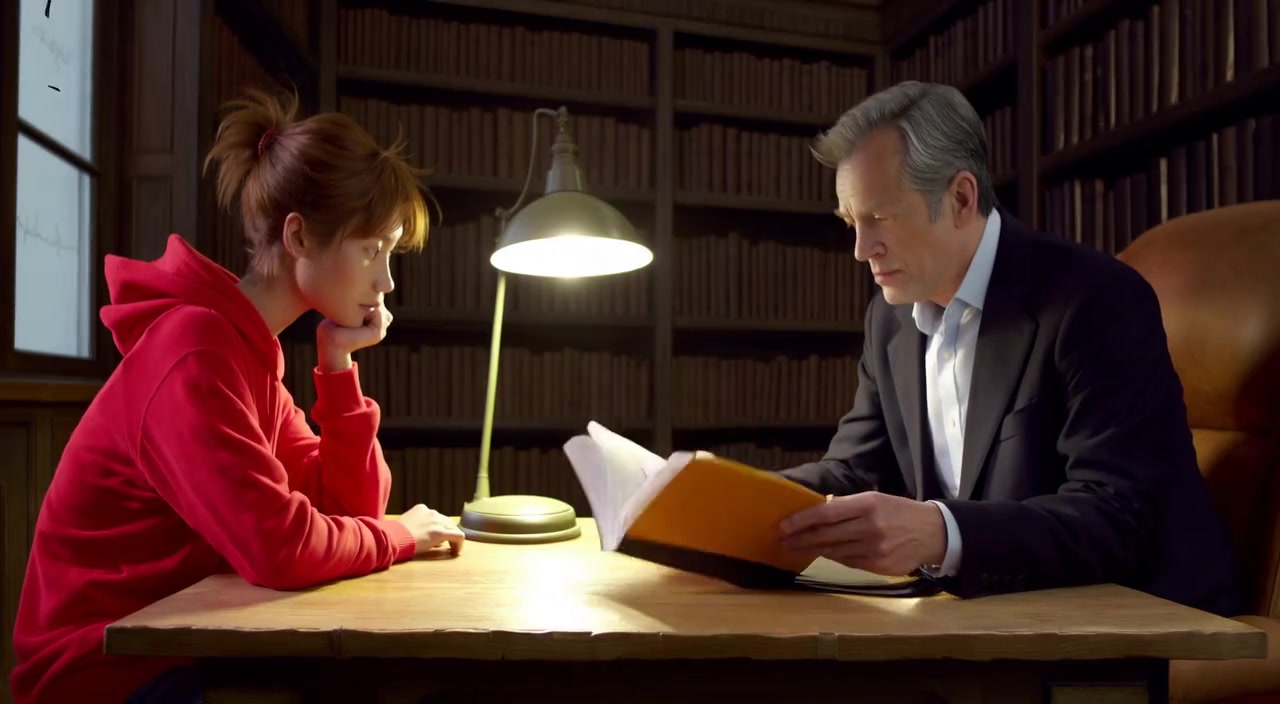}};
    \node[piccell, right=\framegap of r5f2] (r5f3) {\includegraphics[width=\imgw]{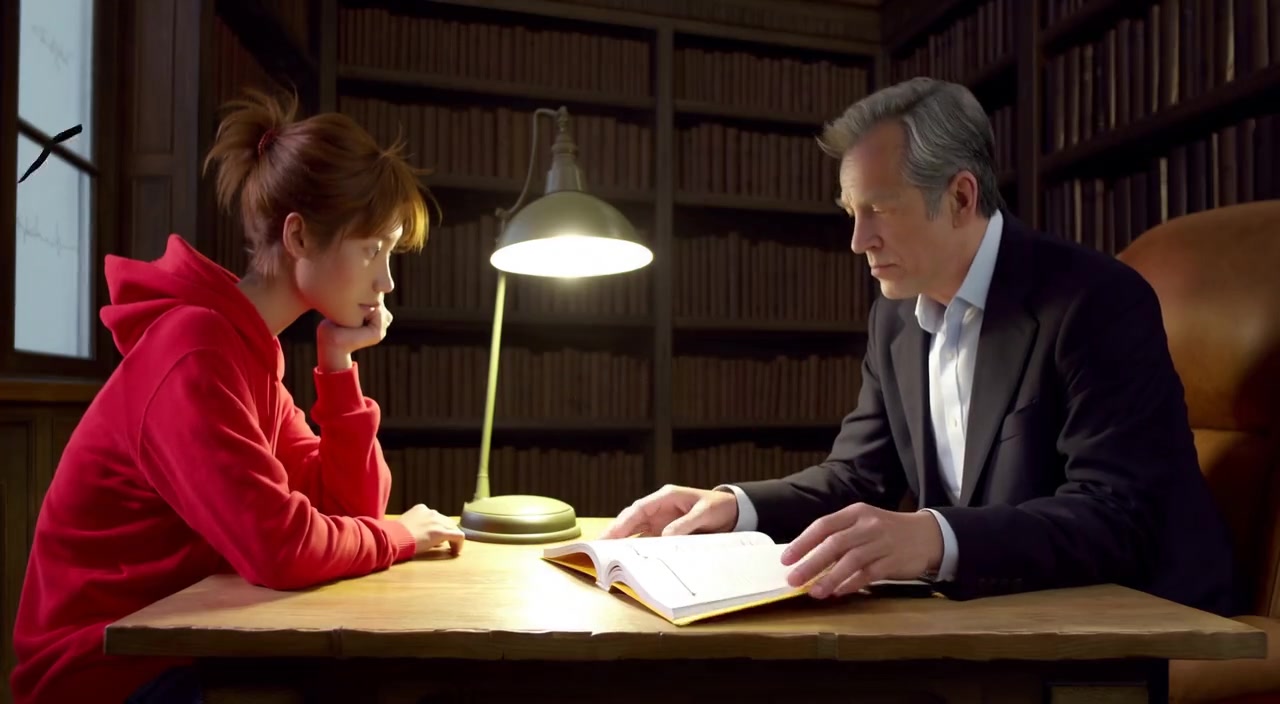}};
    \node[piccell, right=\framegap of r5f3] (r5f4) {\includegraphics[width=\imgw]{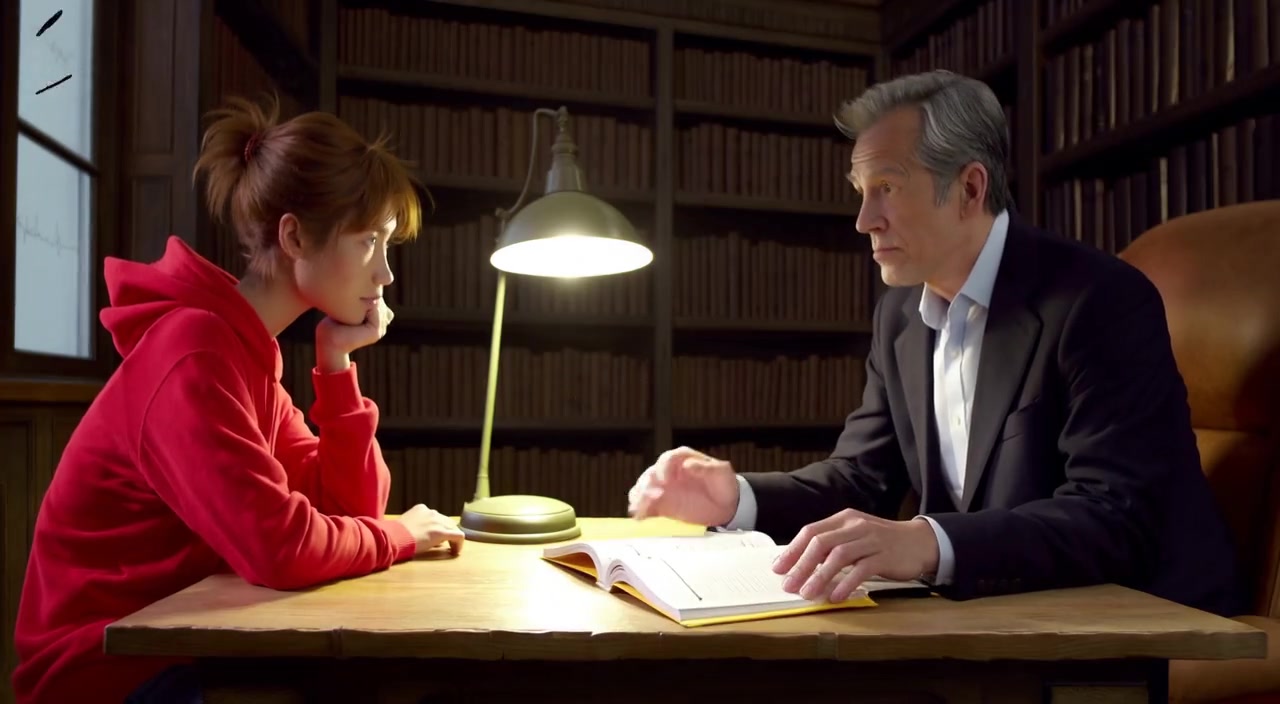}};
    \node[piccell, right=\framegap of r5f4] (r5f5) {\includegraphics[width=\imgw]{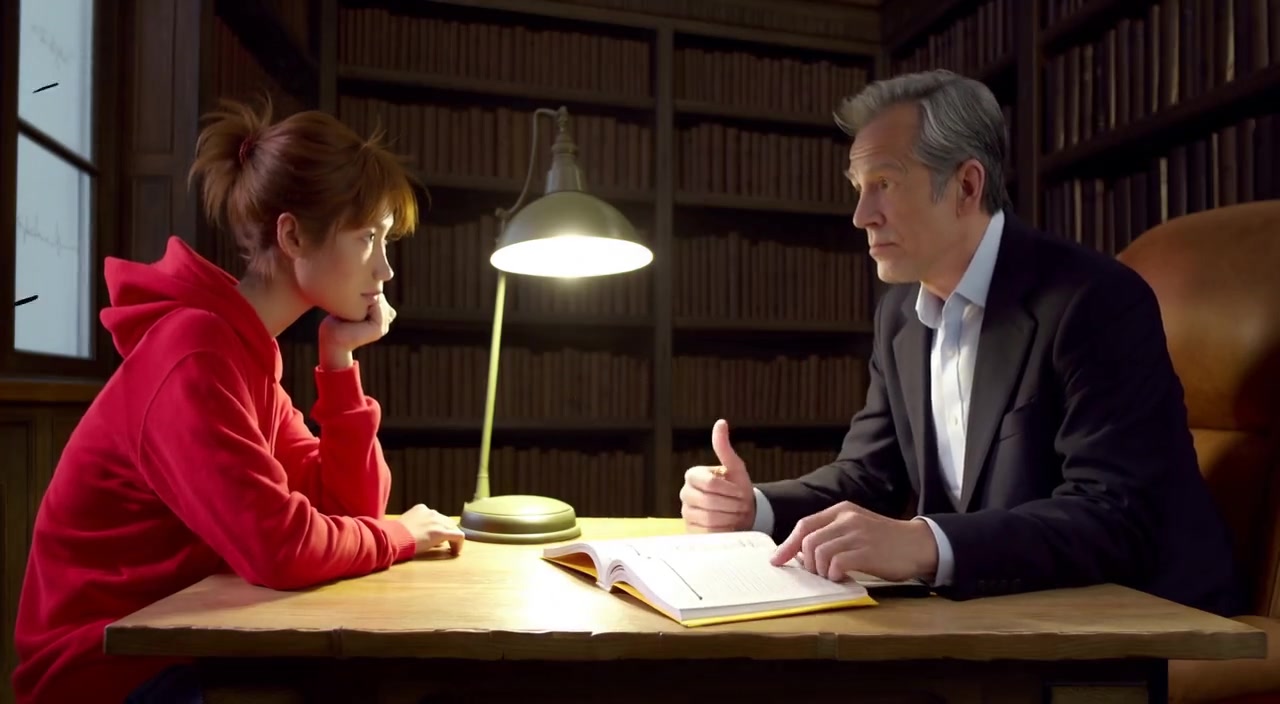}};
    \node[anchor=east, inner sep=0pt, outer sep=0pt]
      at ($(r5f1.west)+(-\labelgap,0)$)
      {\rotatebox{90}{\makebox[\labw][c]{\fontsize{6}{6}\selectfont Ditto (sim2real)}}};

    \node[piccell, below=0pt of r5f1] (r6f1) {\includegraphics[width=\imgw]{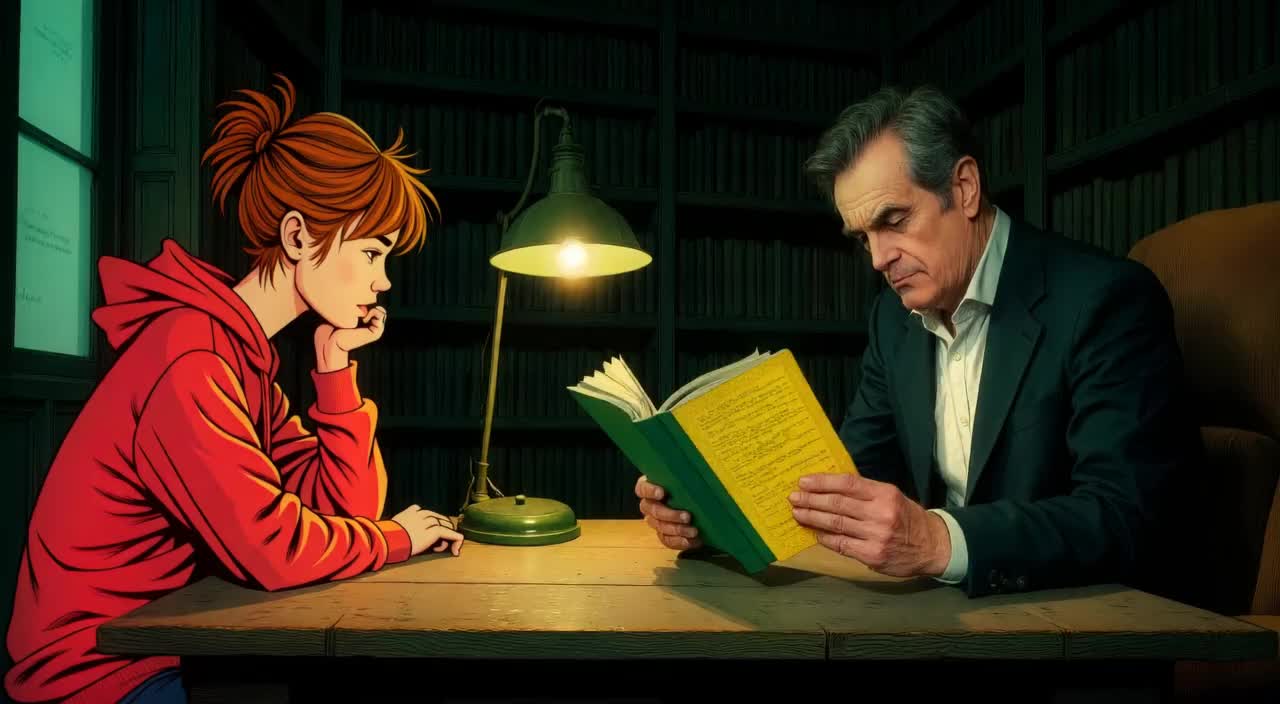}};
    \node[piccell, right=\framegap of r6f1] (r6f2) {\includegraphics[width=\imgw]{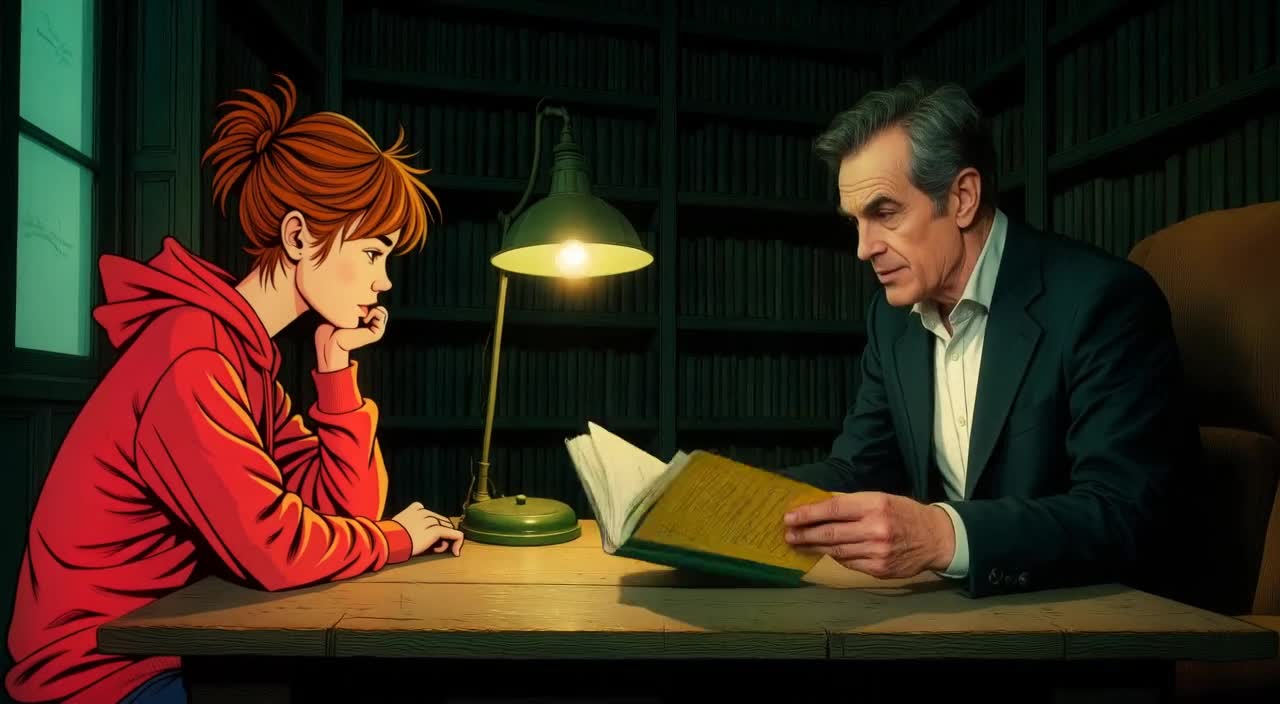}};
    \node[piccell, right=\framegap of r6f2] (r6f3) {\includegraphics[width=\imgw]{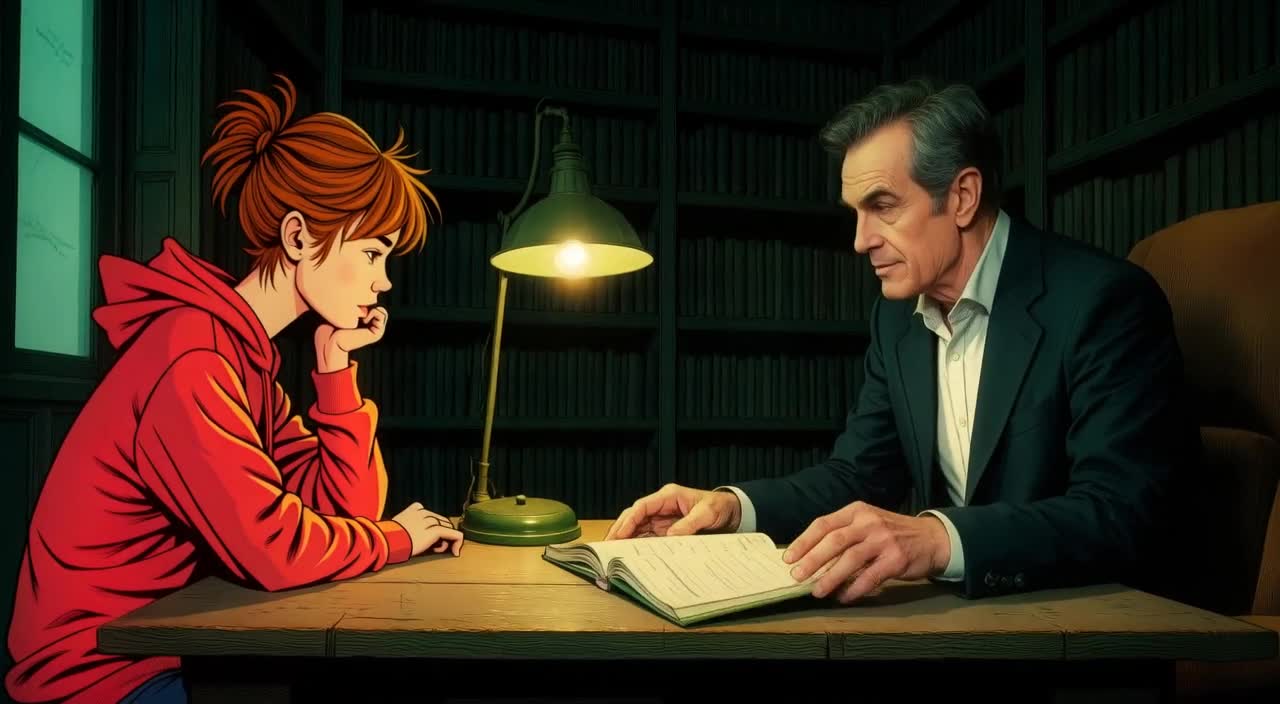}};
    \node[piccell, right=\framegap of r6f3] (r6f4) {\includegraphics[width=\imgw]{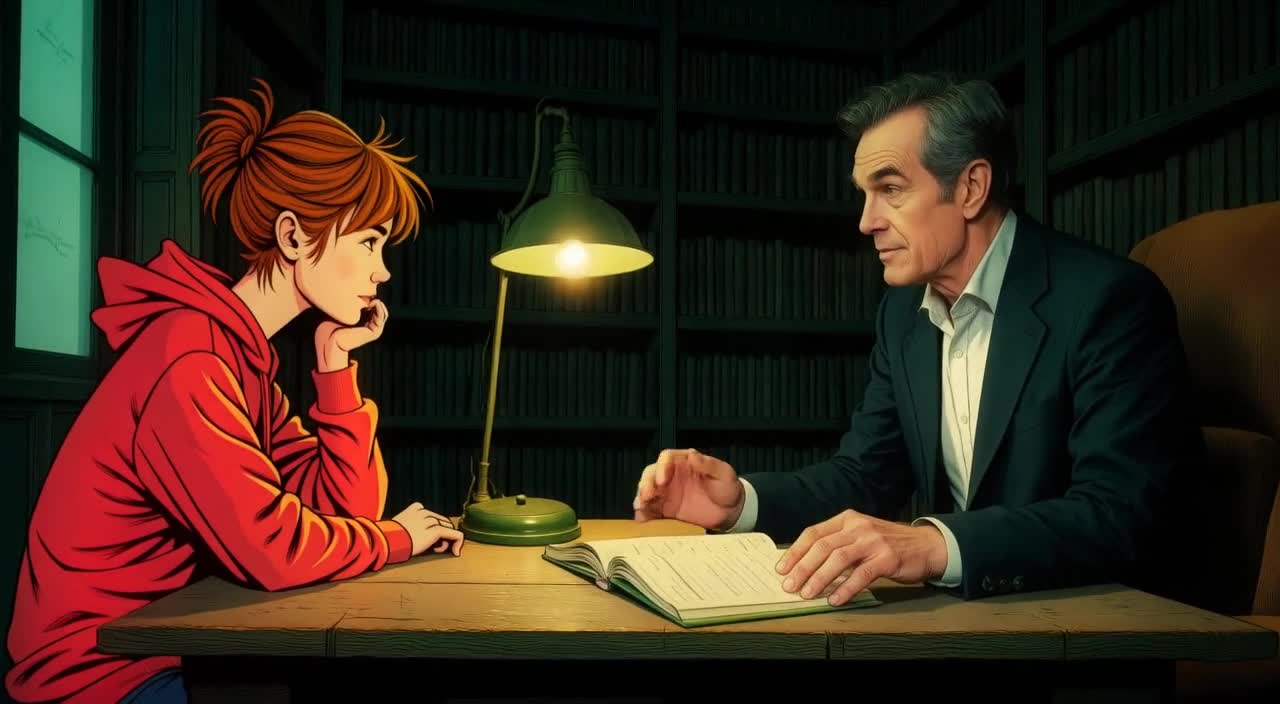}};
    \node[piccell, right=\framegap of r6f4] (r6f5) {\includegraphics[width=\imgw]{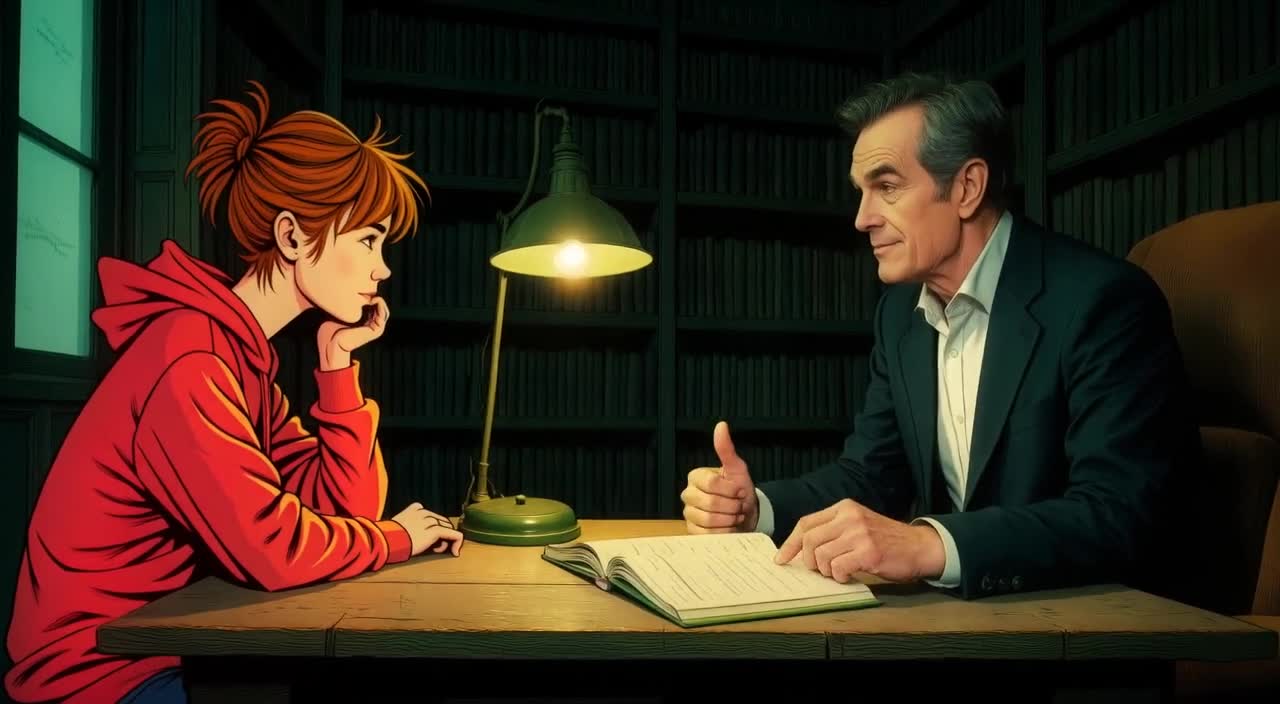}};
    \node[anchor=east, inner sep=0pt, outer sep=0pt]
      at ($(r6f1.west)+(-\labelgap,0)$)
      {\rotatebox{90}{\makebox[\labw][c]{\fontsize{6}{6}\selectfont $\psiPD$ (Ours)}}};

    \node[piccell, below=0pt of r6f1] (r6bf1) {\includegraphics[width=\imgw]{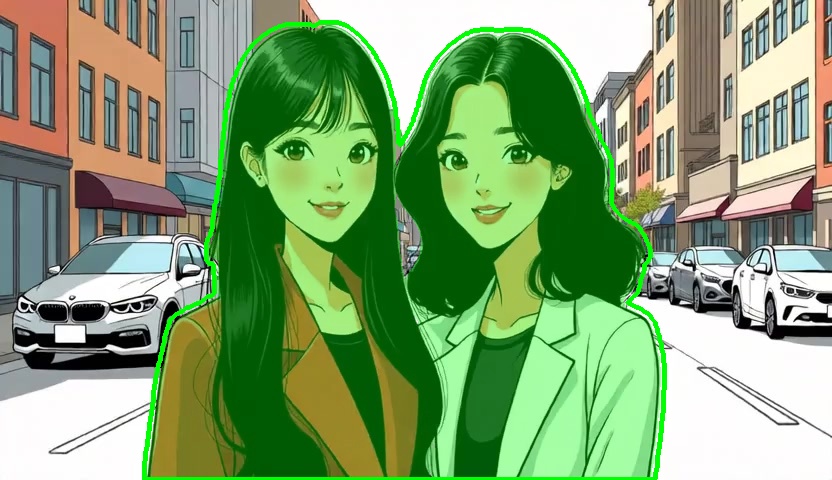}};
    \node[piccell, right=\framegap of r6bf1] (r6bf2) {\includegraphics[width=\imgw]{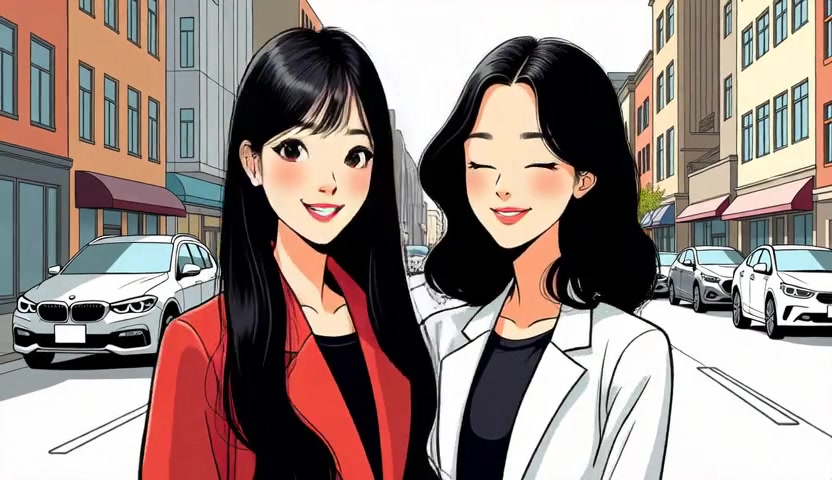}};
    \node[piccell, right=\framegap of r6bf2] (r6bf3) {\includegraphics[width=\imgw]{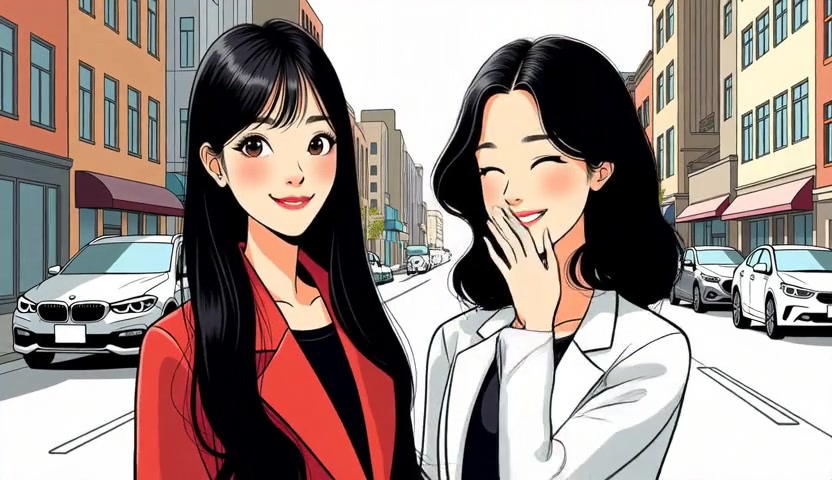}};
    \node[piccell, right=\framegap of r6bf3] (r6bf4) {\includegraphics[width=\imgw]{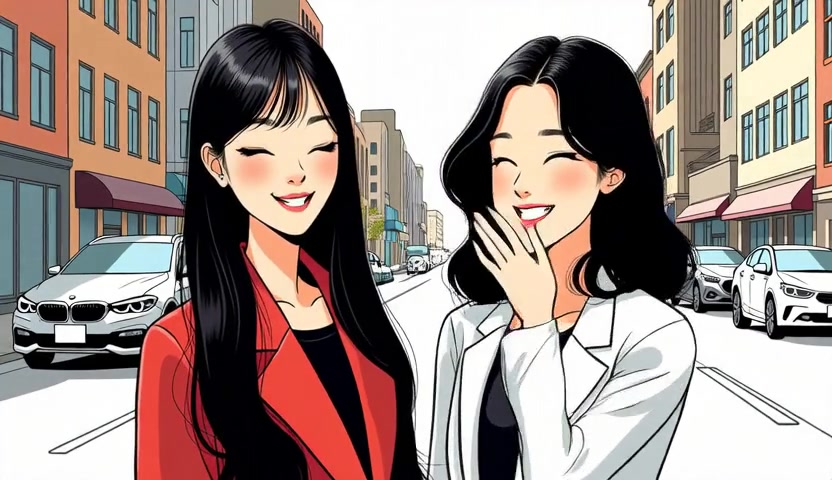}};
    \node[piccell, right=\framegap of r6bf4] (r6bf5) {\includegraphics[width=\imgw]{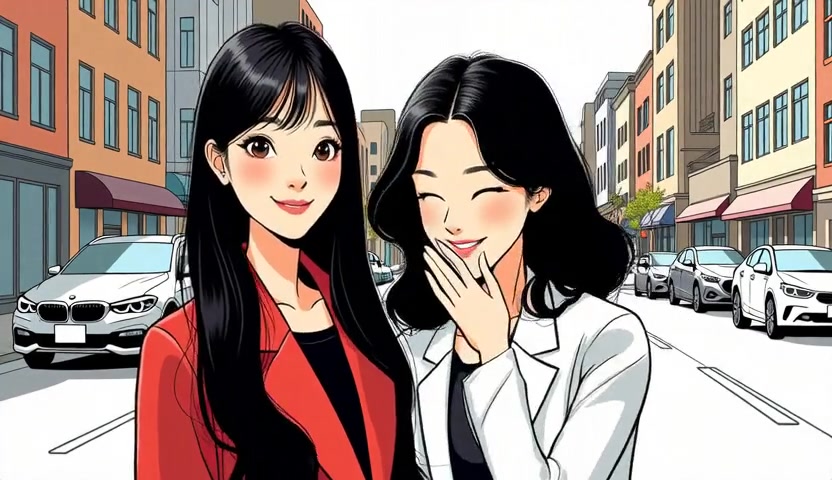}};
    \node[anchor=east, inner sep=0pt, outer sep=0pt]
      at ($(r6bf1.west)+(-\labelgap,0)$)
      {\rotatebox{90}{\makebox[\labw][c]{\fontsize{6}{6}\selectfont Input}}};

    \node[piccell, below=0pt of r6bf1] (r7f1) {\includegraphics[width=\imgw]{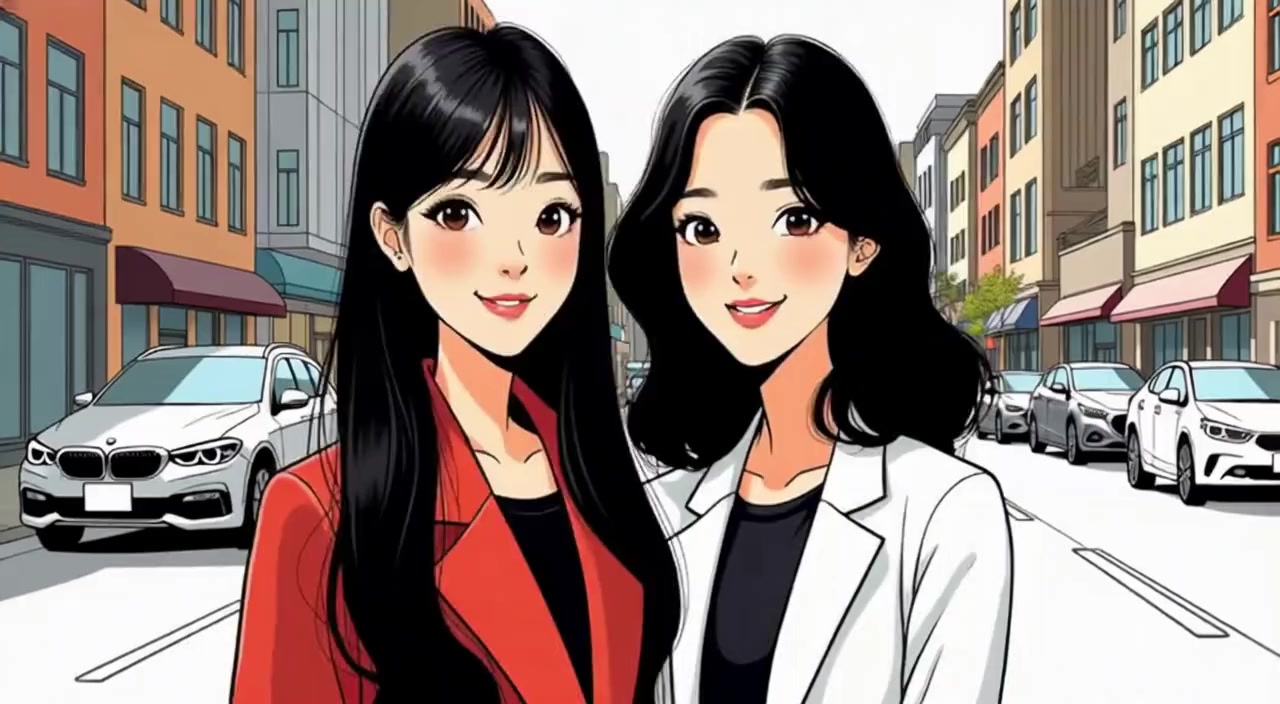}};
    \node[piccell, right=\framegap of r7f1] (r7f2) {\includegraphics[width=\imgw]{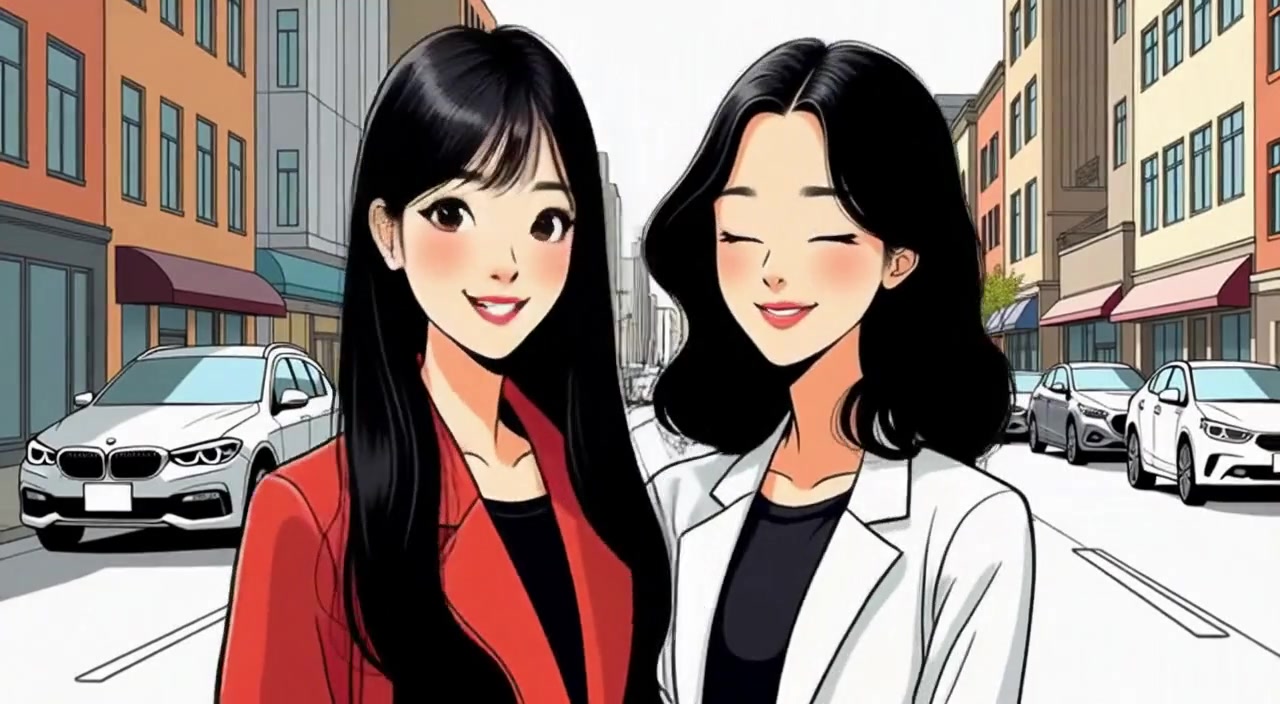}};
    \node[piccell, right=\framegap of r7f2] (r7f3) {\includegraphics[width=\imgw]{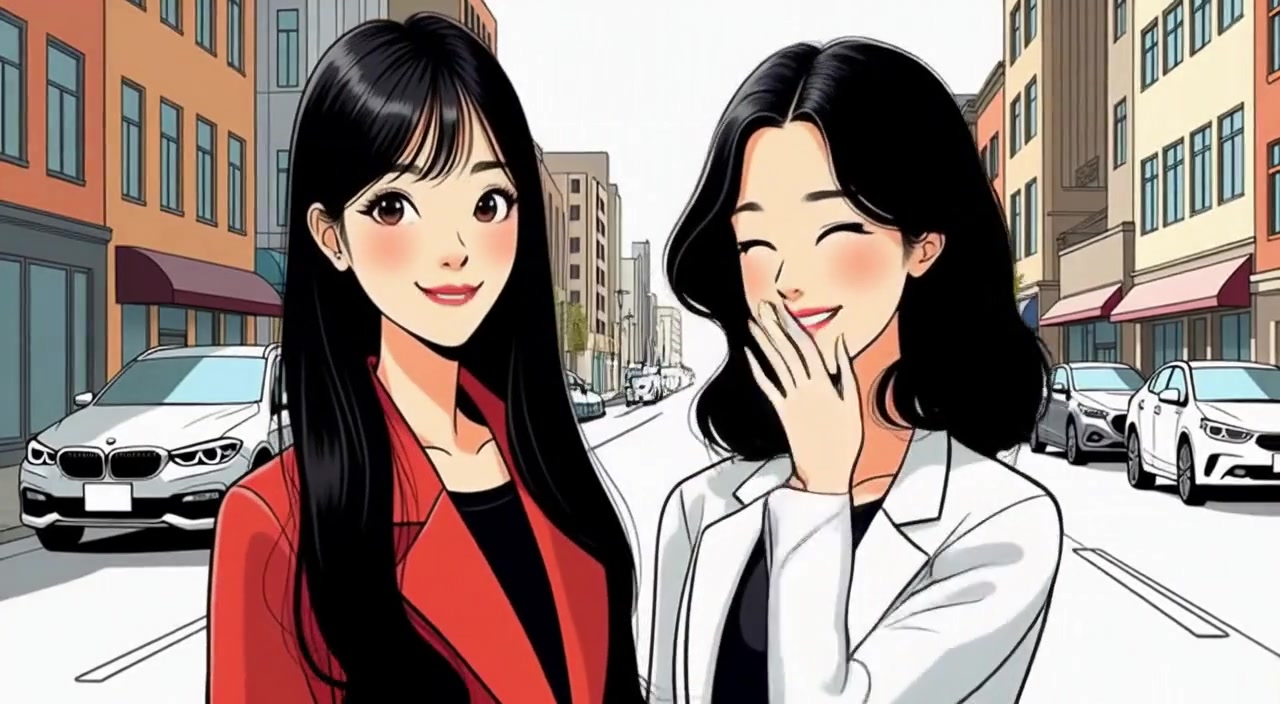}};
    \node[piccell, right=\framegap of r7f3] (r7f4) {\includegraphics[width=\imgw]{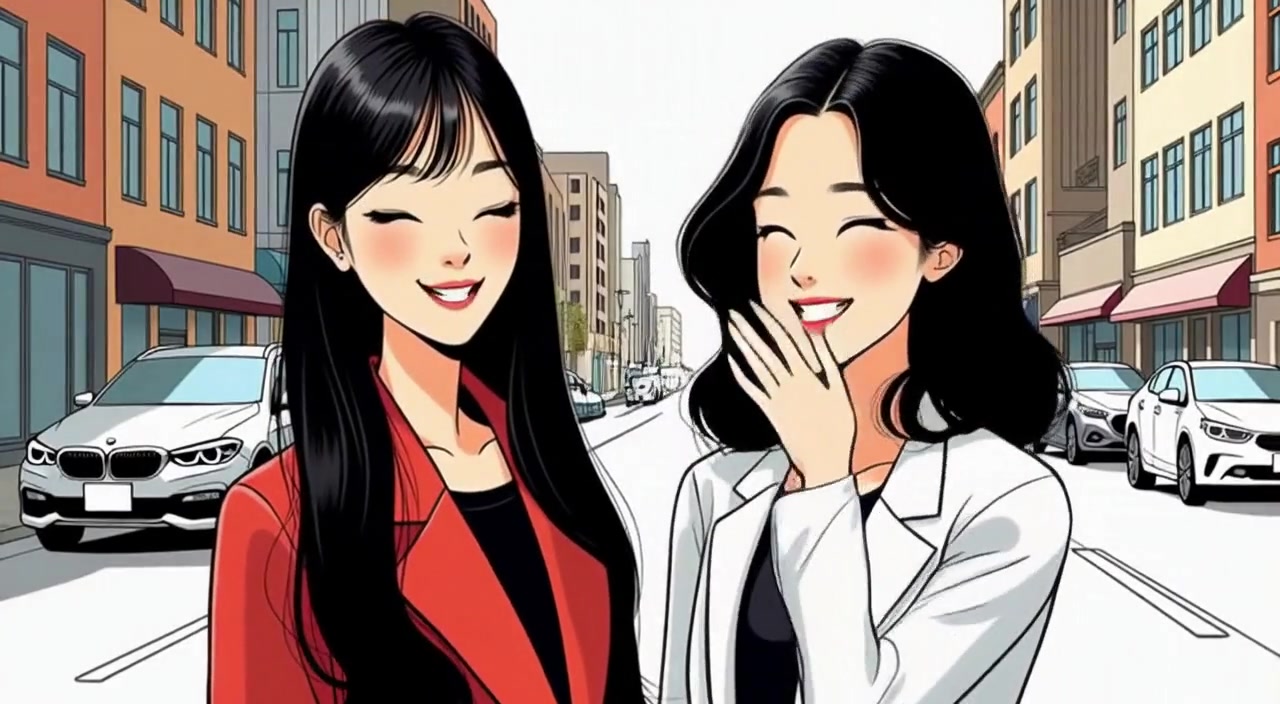}};
    \node[piccell, right=\framegap of r7f4] (r7f5) {\includegraphics[width=\imgw]{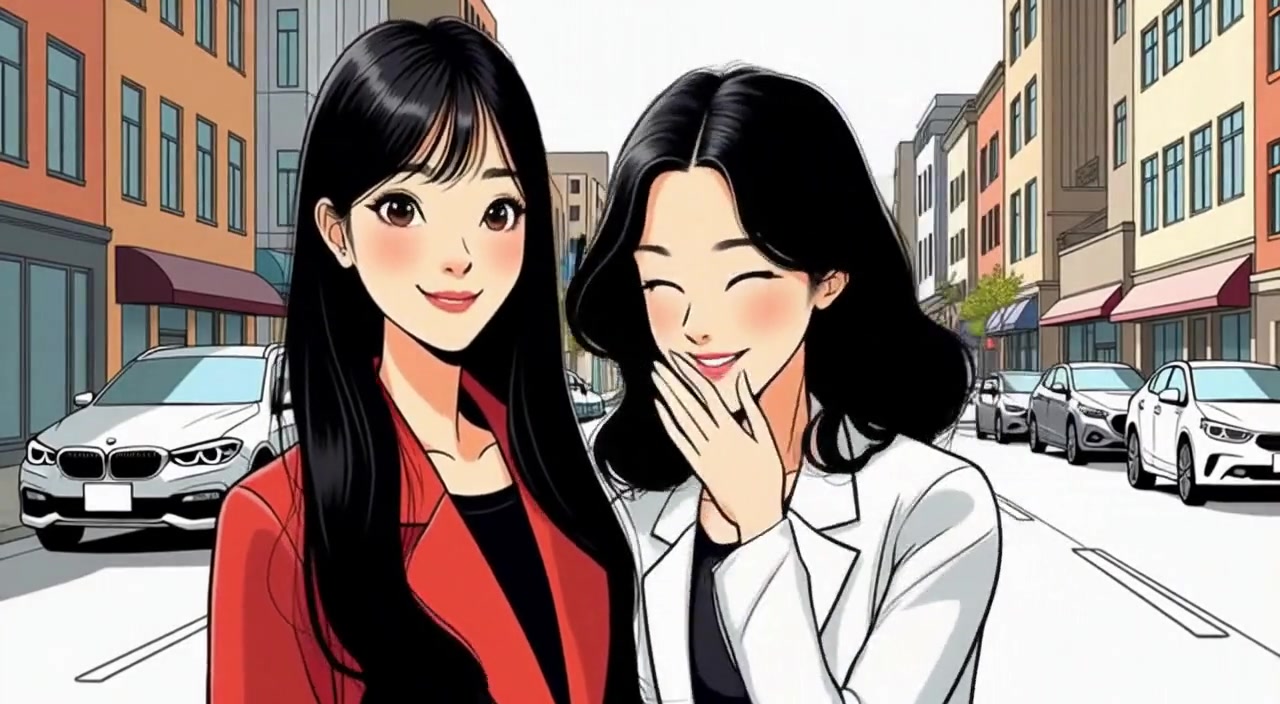}};
    \node[anchor=east, inner sep=0pt, outer sep=0pt]
      at ($(r7f1.west)+(-\labelgap,0)$)
      {\rotatebox{90}{\makebox[\labw][c]{\fontsize{6}{6}\selectfont Ditto (local)}}};

    \node[piccell, below=0pt of r7f1] (r8f1) {\includegraphics[width=\imgw]{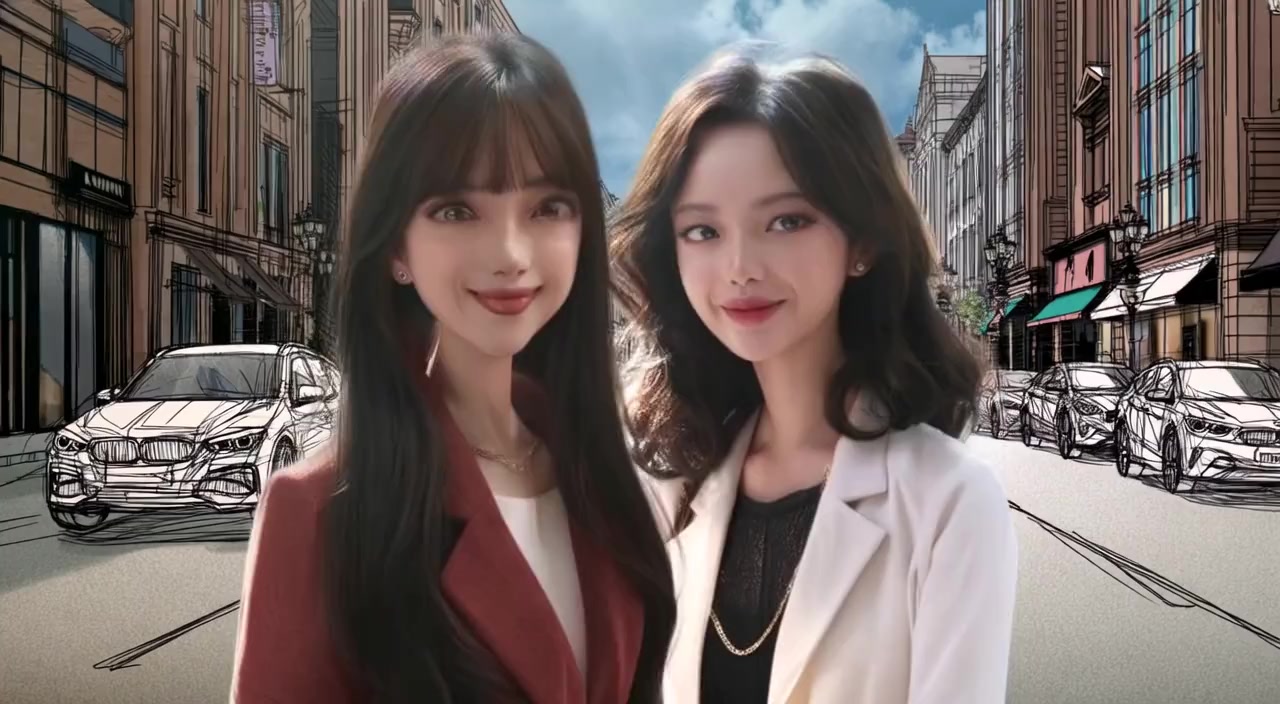}};
    \node[piccell, right=\framegap of r8f1] (r8f2) {\includegraphics[width=\imgw]{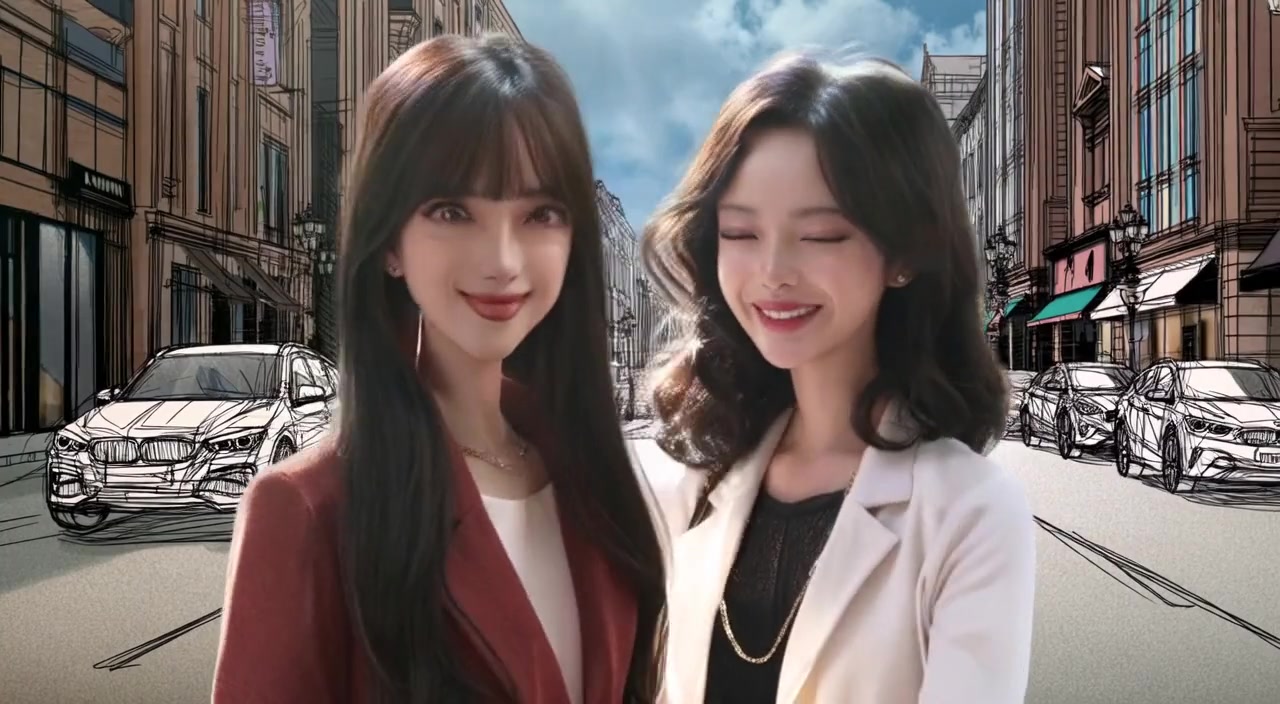}};
    \node[piccell, right=\framegap of r8f2] (r8f3) {\includegraphics[width=\imgw]{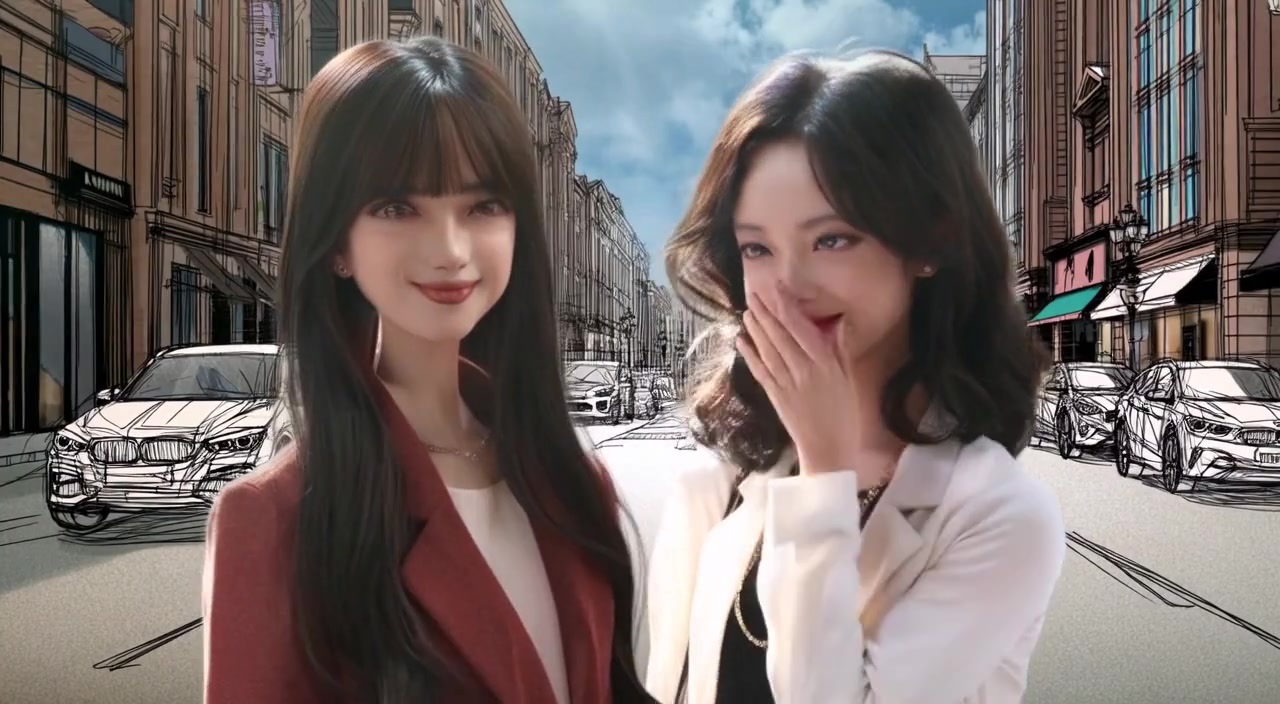}};
    \node[piccell, right=\framegap of r8f3] (r8f4) {\includegraphics[width=\imgw]{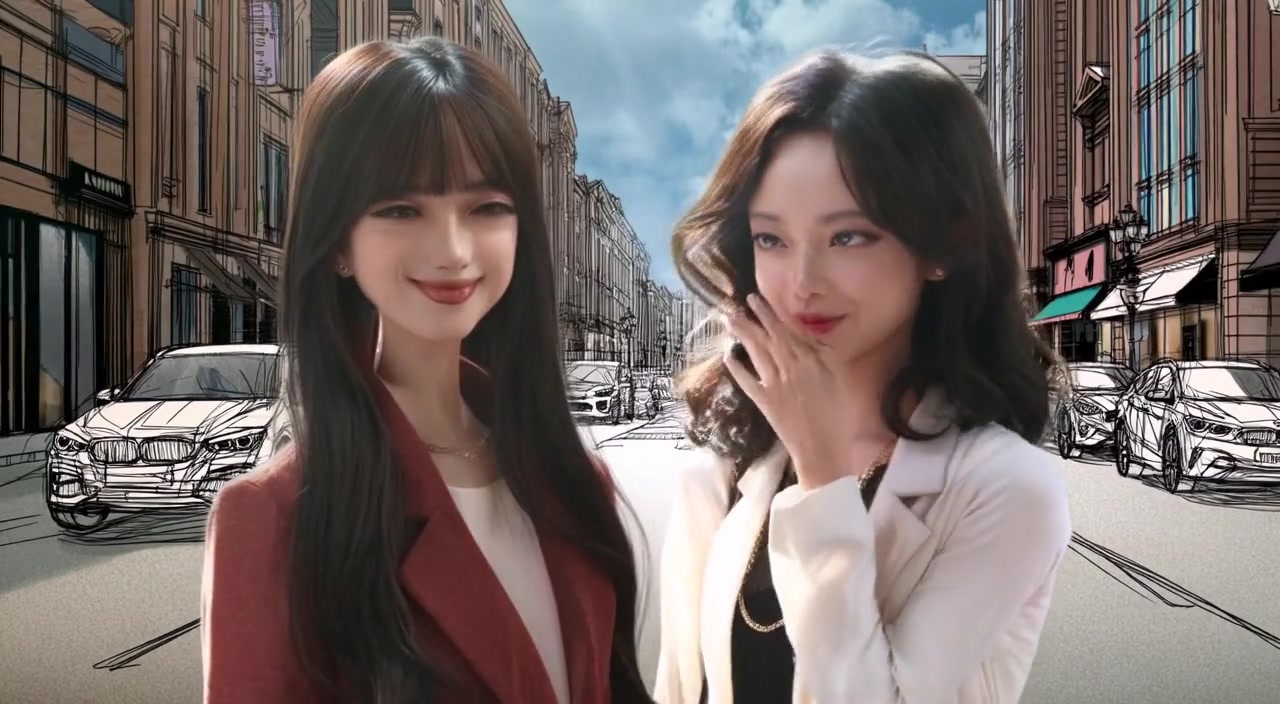}};
    \node[piccell, right=\framegap of r8f4] (r8f5) {\includegraphics[width=\imgw]{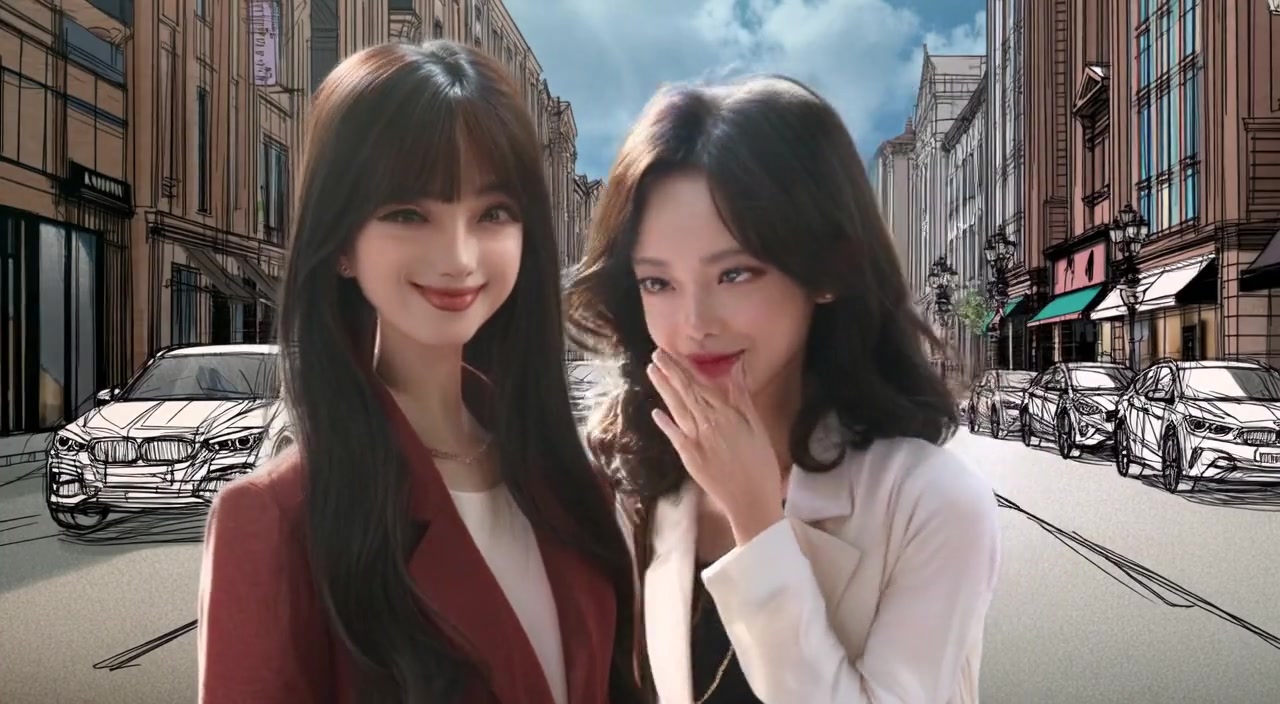}};
    \node[anchor=east, inner sep=0pt, outer sep=0pt]
      at ($(r8f1.west)+(-\labelgap,0)$)
      {\rotatebox{90}{\makebox[\labw][c]{\fontsize{6}{6}\selectfont Ditto (sim2real)}}};

    \node[piccell, below=0pt of r8f1] (r9f1) {\includegraphics[width=\imgw]{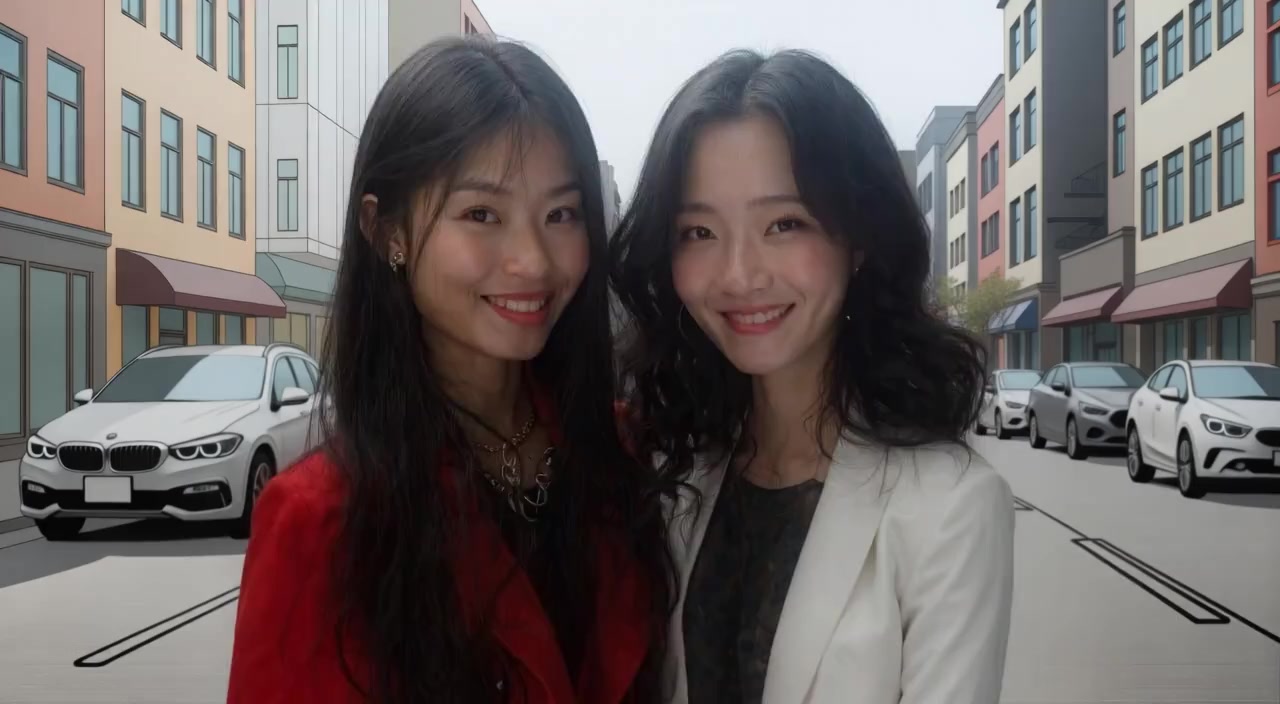}};
    \node[piccell, right=\framegap of r9f1] (r9f2) {\includegraphics[width=\imgw]{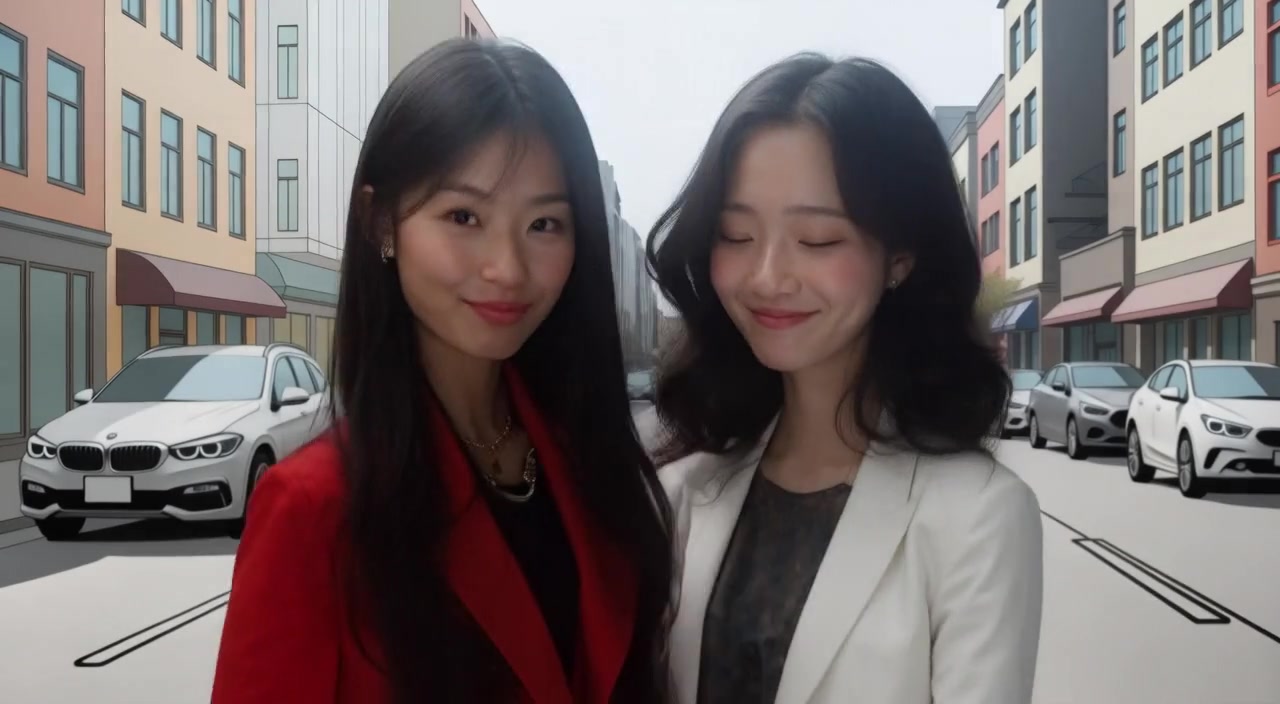}};
    \node[piccell, right=\framegap of r9f2] (r9f3) {\includegraphics[width=\imgw]{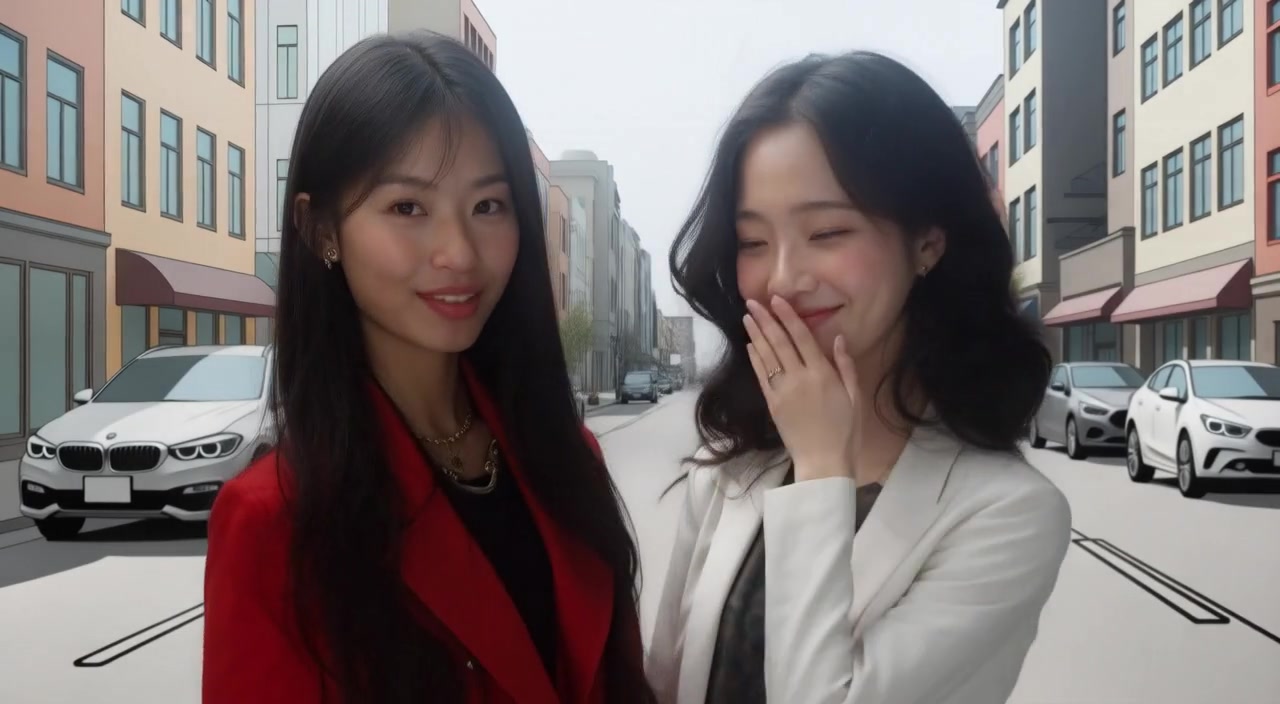}};
    \node[piccell, right=\framegap of r9f3] (r9f4) {\includegraphics[width=\imgw]{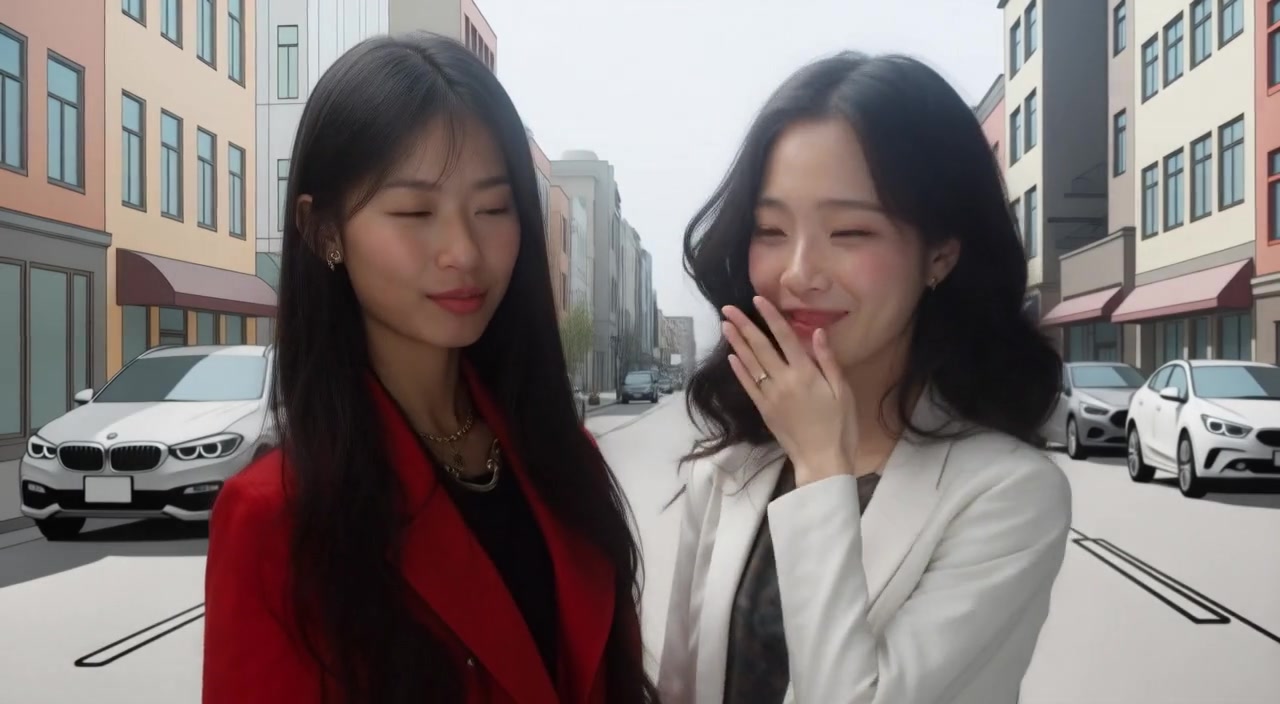}};
    \node[piccell, right=\framegap of r9f4] (r9f5) {\includegraphics[width=\imgw]{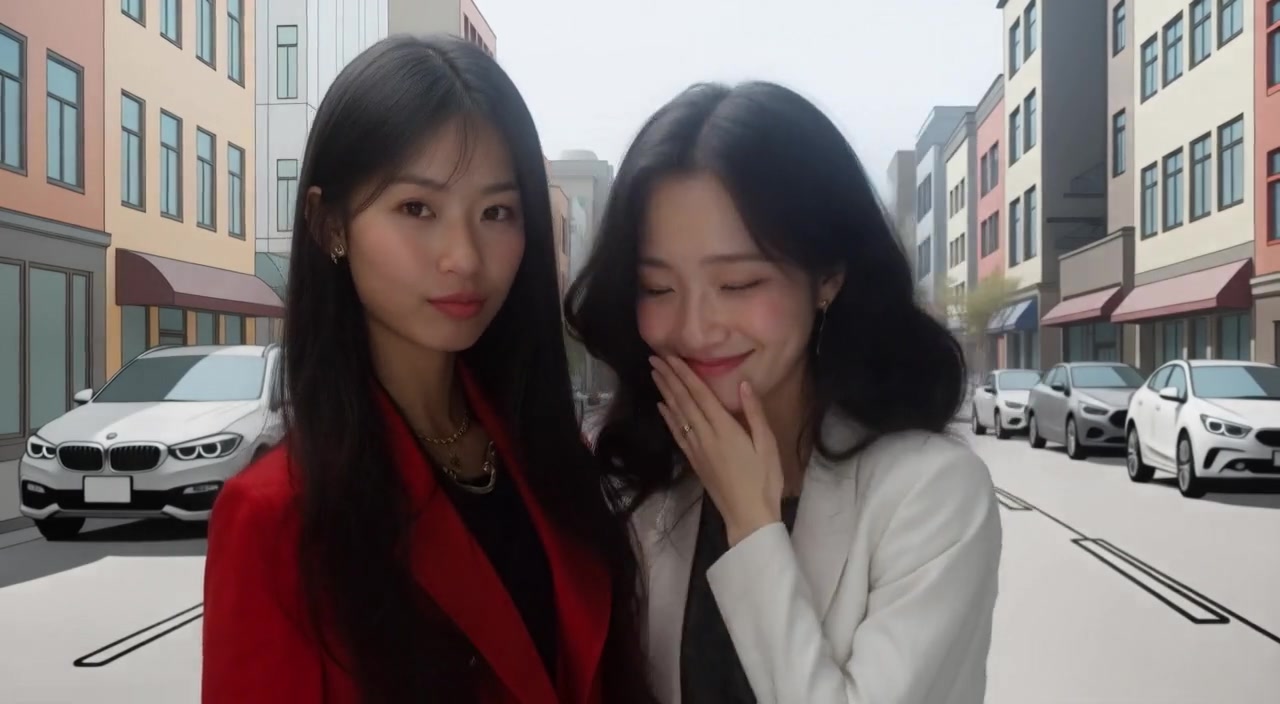}};
    \node[anchor=east, inner sep=0pt, outer sep=0pt]
      at ($(r9f1.west)+(-\labelgap,0)$)
      {\rotatebox{90}{\makebox[\labw][c]{\fontsize{6}{6}\selectfont $\psiPD$ (Ours)}}};
    \end{tikzpicture}
    \caption{\textbf{Zero-shot instance-level video translation.} Five consecutive frames from three scenes. In all cases, the target instance is highlighted in green in the first input frame.}
    \label{fig:application_full}
\end{figure*}

\clearpage

\section{Zero-Shot Instance-Level Translation}
\label{supp:instance}

\cref{fig:application_full} extends the instance-level comparison of Fig.~7 in the main paper to three scenes, showing five consecutive frames per scene for both Ditto checkpoints and $\psiPD$.
In each scene, both Ditto checkpoints receive an instruction to make only the target region photorealistic: the character's mirror reflection in Scene~1 (comic-style mirror), the man on the right in Scene~2 (animated library), and the two foreground characters in Scene~3 (anime-style illustration), with the sketched background to remain untranslated.
In all three scenes, Ditto (local) leaves the target essentially unchanged, while Ditto (sim2real) applies a global style transfer to the entire frame; neither checkpoint confines the edit to the target region.
In contrast, $\psiPD$ translates only the target region, controlled by a per-pixel cutoff map, with negligible inference overhead and no architectural modification.

\section{Relation to NeuralRemaster}
\label{supp:neuralremaster}

As the closest prior work, NeuralRemaster shares the high-level goal of phase-preserving diffusion but differs from $\psiPD$ along three concrete axes.
\emph{Basis:} NeuralRemaster uses globally-supported Fourier coefficients, where any spectral modification leaks into every spatial location; $\psiPD$ uses localized DT-$\mathbb{C}$WPT packets that confine phase injection to compact spatial supports, substantially reducing ringing and boundary leakage.
Crucially, because NeuralRemaster's Fourier phase preservation encodes low-pass filtering artifacts (ringing and boundary leakage) directly into the preserved phase, the powerful diffusion prior is forced to hallucinate realistic but incorrect structures to reconcile these artifacts, breaking geometric consistency with the original simulation.
In contrast, $\psiPD$ avoids injecting false structures due to the compact spatial support of its wavelet packets.
\emph{Cutoff:} NeuralRemaster applies a single global scalar radius $r$, forcing one realism--consistency trade-off across the entire image; $\psiPD$ accepts a spatially adaptive cutoff map $\mathbf{F}$, which makes instance-level translation a zero-shot consequence of the design.
\emph{Low-band treatment:} NeuralRemaster preserves the source low-frequency content, locking the output to the synthetic illumination prior; $\psiPD$'s Low-Frequency Randomization frees the model to synthesize in-distribution real-world appearance.

\section{Limitations}
\label{supp:limitations}

Because $\psiPD$ preserves the phase structure of the source, the achievable realism is bounded by the geometric fidelity of the simulator itself.
Appearance deficiencies (such as flat textures or uniform surface colors) can be corrected by the diffusion model.
However, when simulator objects have incorrect proportions or stylized shapes that deviate from real-world counterparts, phase preservation anchors those geometric errors into the output: the result is a photorealistic rendering of an unrealistic shape, which can appear more uncanny than the original.
In such settings, improving the simulator's geometric accuracy is a prerequisite for further realism gains.

\fi

\end{document}